%% file: main.tex
\documentclass[lettersize,journal]{IEEEtran}
\usepackage{amsmath,amsfonts}
\usepackage{algorithmic}
\usepackage{algorithm}
\usepackage{array}
\usepackage[caption=false,font=normalsize,labelfont=sf,textfont=sf]{subfig}
\usepackage{textcomp}
\usepackage{stfloats}
\usepackage{url}
\usepackage{verbatim}
\usepackage{graphicx}
\usepackage{xcolor}
\usepackage[normalem]{ulem}
\usepackage[switch]{lineno}

\usepackage{cite}
\usepackage{tikz}
\usepackage{makecell}
\usepackage{float}
\usepackage{placeins}
\input{utils/include}

\input{utils/general_utils}

\input{utils/math_utils}

\graphicspath{{content/}}
\graphicspath{{content/figures/}}

\begin{document}

\title{Foundation and Multimodal Large Language Models for Face Presentation and Morph Attack Detection}

\author{
Hatef Otroshi Shahreza,
Asif Hussain Khan,
Peter Lorenz,
Alain Komaty,
S\'{e}bastien Marcel
\thanks{All authors are affiliated with Idiap Research Institute, Switzerland.
S\'{e}bastien Marcel is also affiliated with Université de Lausanne, Switzerland.}
\thanks{Hatef Otroshi Shahreza, Asif Hussain Khan, and Peter Lorenz contributed equally to this work.}
\thanks{Correspondence: Hatef Otroshi Shahreza (hatef.otroshi@idiap.ch).}
\thanks{This work was funded by the European Union projects CarMen
(Grant Agreement No. 101168325) and ORION (Grant Agreement No. 101225611).}
}



\maketitle

\input{content/00_abstract}

\input{content/01_introduction}

\input{content/02_related_work}

\input{content/03_FMs_4_PAD_MAD}

\input{content/04_experiments}

\input{content/05_discussions}

\input{content/06_conclusions}

\section*{Acknowledgments}
The authors would like to thank Dr. Anjith George for fruitful discussions and  his valuable comments.
\bibliographystyle{IEEEtran}
\bibliography{refs}


\vfill\input{content/supplementary}

\end{document}

%% file: utils/include.tex
\usepackage{microtype}
\usepackage{graphicx}
\usepackage{booktabs} 

\usepackage{amsmath}
\usepackage{amssymb}
\usepackage{mathtools}
\usepackage{amsthm}
\usepackage{MnSymbol}
\usepackage{wasysym}

\usepackage{wasysym}

\usepackage[breaklinks,colorlinks]{hyperref}

\usepackage[capitalize,noabbrev]{cleveref}

\usepackage[textsize=tiny]{todonotes}

\usepackage{pifont}

\usepackage{blindtext}
\usepackage{lipsum}
\usepackage{threeparttable}

\usepackage{times}
\usepackage{multirow}
\usepackage{url}

\usepackage{enumitem}
\usepackage{xspace}
\usepackage{xfrac}
\usepackage{comment}
\usepackage{tikz}
\usetikzlibrary{arrows.meta, positioning, fit, backgrounds, calc}
\usepackage{etoolbox}

\usepackage{placeins}

\definecolor{PL_color}{rgb}{0.858, 0.188, 0.478}

\usepackage{pifont}
\usepackage{adjustbox}
\usepackage{threeparttable}
\usepackage{multirow}

\DeclareMathAlphabet\mathbfcal{OMS}{cmsy}{b}{n}

\DeclarePairedDelimiterX{\inp}[2]{\langle}{\rangle}{#1, #2}

\usepackage{acro}

\usepackage{color, colortbl}
\definecolor{Gray}{gray}{0.93}
\definecolor{Orange}{rgb}{1,0.5,0}
\definecolor{DGray}{gray}{0.83}
\definecolor{LightCyan}{rgb}{0.88,1,1}

\RequirePackage{color}\definecolor{RED}{rgb}{1,0,0}\definecolor{BLUE}{rgb}{0,0,1} 

\usepackage{newfloat}
\DeclareFloatingEnvironment[name={Supplementary Figure}]{suppfigure}
\usepackage{sidecap}

\usepackage{xcolor}
\usepackage{arydshln}

\newcommand{\iconfrozen}{\raisebox{-0.12ex}{\includegraphics[height=1.25em]{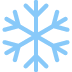}}}
\newcommand{\icontrainable}{\raisebox{-0.12ex}{\includegraphics[height=1.25em]{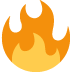}}}

%% file: utils/general_utils.tex
\usepackage{pifont}

\newcommand{\bestcell}[1]{\textbf{#1}}

\usepackage{color, colortbl}
\definecolor{Gray}{gray}{0.93}
\definecolor{Orange}{rgb}{1,0.5,0}
\definecolor{DGray}{gray}{0.83}
\definecolor{LightCyan}{rgb}{0.88,1,1}
\definecolor{lightbrown}{rgb}{0.71, 0.4, 0.11}

\usepackage[most]{tcolorbox}

%% file: utils/math_utils.tex
\usepackage{amsmath,amsfonts,bm}

\def\eqref#1{(\ref{#1})}

\def\1{\bm{1}}

\def\ra{{\textnormal{a}}}
\def\rb{{\textnormal{b}}}
\def\rc{{\textnormal{c}}}
\def\rd{{\textnormal{d}}}
\def\re{{\textnormal{e}}}

\DeclareMathAlphabet{\mathsfit}{\encodingdefault}{\sfdefault}{m}{sl}
\SetMathAlphabet{\mathsfit}{bold}{\encodingdefault}{\sfdefault}{bx}{n}



%% file: content/00_abstract.tex
\begin{abstract}
    Face recognition systems are increasingly deployed in security-critical applications, yet they remain vulnerable to presentation and morph attacks. Presentation attack detection (PAD) and morphing attack detection (MAD) are therefore essential components of trustworthy face biometrics. Despite advancements in PAD and MAD methods, existing detectors suffer from limited generalization and degrade in cross-dataset evaluation. In this paper, we systematically investigate whether general-purpose foundation models (FMs) and multimodal large language models (MLLMs) encode PAD-relevant and MAD-relevant information, and how such models can best be deployed for both tasks. We study five approaches with increasing access to the internal information of the model: (i) zero-shot prompting of off-the-shelf MLLMs; (ii) training a shallow model on the next-token logit probabilities at the output of the MLLM; (iii) parameter-efficient fine-tuning on task-specific question–answer data, yielding two specialized MLLMs, called PADLLM and MADLLM, which additionally provide textual reasoning for their decisions; (iv) linear probing of frozen vision encoders; and (v) fine-tuning of vision encoders of FMs and MLLMs. We benchmark 16 open-weight MLLMs and 30 vision encoder backbones on four PAD datasets (MSU-MFSD, CASIA-FASD, Replay-Attack, and OULU-NPU) and four MAD datasets (FFHQ, FRGC, FRLL, and FERET). Our experiments show that FMs and MLLMs can achieve significant performance for PAD and MAD. In addition, the fine-tuned models achieve state-of-the-art detection performance in cross-dataset evaluation, indicating that general-purpose pretrained representations carry substantial attack-relevant information. Source code of all our experiments will be publicly released.

\end{abstract}
        
\begin{IEEEkeywords}
Foundation Model, Multimodal Large Language Model (MLLM), Vision-Encoder, Face Recognition, Presentation Attack Detection (PAD), Morph Attack Detection (MAD).
\end{IEEEkeywords}


%% file: content/01_introduction.tex
\section{Introduction}

Face recognition systems are increasingly used in security-critical applications such as border control, digital identity verification, and physical access control.
Despite their high recognition accuracy, these systems remain vulnerable to biometric attacks, which undermine their reliability. 
Two threats dominate this landscape: 1) presentation attacks, which appears in the acquisition stage by presenting artificial samples (such as printed photographs, replayed videos, or physical masks, etc.), to impersonate a genuine user~\cite{galbally2014biometric,marcel2019handbook}; 2) morphing attacks, which exploit the enrollment stage by combining the facial characteristics of multiple identities into a single image so that a matcher can accept as either individual~\cite{schwaiger2022man, ferrara2014magic,venkatesh2021face,shahreza2025generation}.

\begin{figure*}[htbp]
    \centering
    \input{content/figures/pipeline_figure}
    \caption{
    MLLM-based presentation and morph attack detection pipeline and key contributions (as referred to \cref{sec:method} A-E).
    }
    \label{fig:generic_pipeline}
\end{figure*}
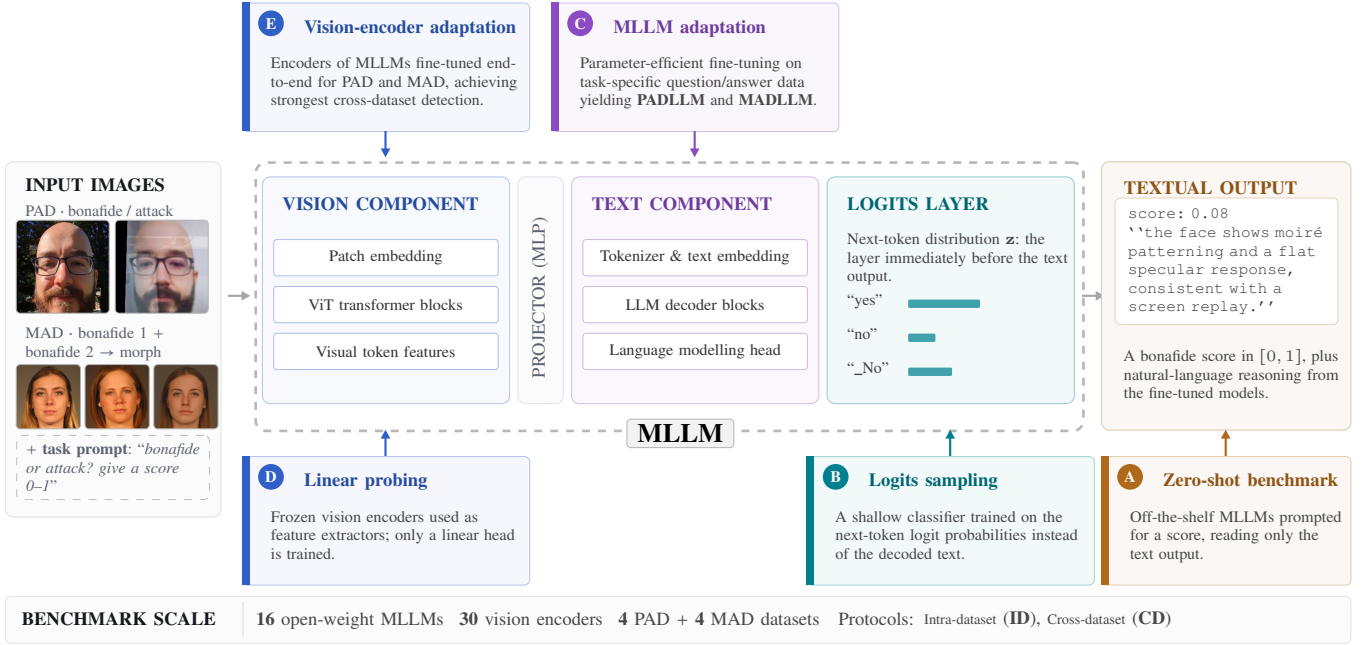

The corresponding defense mechanisms, \textit{i.e.}, presentation attack detection (PAD) and morph attack detection (MAD), have therefore become essential components for trustworthy face biometrics.

Both attack families have evolved substantially over recent years.
Presentation attacks have evolved from relatively simple printed photographs and screen replays to high-quality displays, 3D masks, and other physical presentation attack instruments capable of reproducing increasingly realistic facial appearance ~\cite{chingovska2012effectiveness,erdogmus2013spoofing, heusch2020deep}. 
Correspondingly, PAD methods have evolved from handcrafted texture and motion cues toward deep neural networks designed to capture richer spatial, temporal, and physiological evidence ~\cite{kollreider2007real,atoum2017face,maatta2011face,liu2018learning}.

Morphing also began with landmark-based image warping~\cite{ferrara2014magic} and has since moved to generative adversarial networks~\cite{venkatesh2020can, sarkar2022gan, zhang2021mipgan} and diffusion models~\cite{damer2023mordiff}, each generation leaving fewer visible artifacts than the last. 
This arms-race dynamic means PAD and MAD systems validated on one generation of attacks routinely underperform on the next, motivating detectors that generalize from broad visual and semantic cues rather than artifacts specific to a single attack generator.
Along the same vein, deep neural networks trained for PAD can achieve strong performance when trained and tested on the same dataset, but degrade under cross-dataset shifts in sensors, illumination, and attack types~\cite{DBLP:conf/cvpr/ShaoLLY19,DBLP:conf/aaai/ShaoLY20,DBLP:conf/cvpr/JiaZSC20,DBLP:journals/tbbis/WangWDG22,DBLP:journals/tcsv/YanZH22,fang2024face}.
Despite differences in presentation attacks and morph attacks, PAD and MAD can be similarly formulated as a binary classification problem (\textit{i.e.,} to distinguish bonafide vs attack). 

Recently, foundation models (FMs) and multimodal large language models (MLLMs) have been shown to achieve strong performance in different applications~\cite{naveed2025comprehensive,raiaan2024review,shahreza2025foundation}.
Vision FMs and MLLMs pretrained on large-scale image or image–text data offer an alternative by providing generic representations that may already encode cues relevant for PAD~\cite{srivatsan2023flip,lin2025instructflip,ozgur2025foundpad,feng2026benchmarking,wang2025fsfm} or MAD~\cite{zhang2025chatgpt,maricexploring,patwardhan2024empowering}.
Latest works explored the use of FMs for PAD through multimodal architectures (e.g., FLIP~\cite{srivatsan2023flip}, I-FAS~\cite{zhang2025interpretable}, InstructFLIP~\cite{lin2025instructflip}), fine-tuning (e.g., CLIP~\cite{radford2021learning} in FoundPAD~\cite{ozgur2025foundpad}, and on DINOv2 with Registers~\cite{feng2026benchmarking}), and self-supervised pre-training without labels, i.e., FSFM~\cite{wang2025fsfm} for PAD. Similar approaches have also been explored for MAD, such as multimodal architecture studies \cite{colbois2024evaluating}, fine-tuning \cite{caldeira2025madation}, etc. Each method uses different approaches to deploy FMs and MLLMs for PAD or MAD tasks.

In this paper, we systematically explore whether general-purpose foundation models and multimodal large language models contain sufficient PAD-relevant and MAD-relevant representations, and what the best approach is to deploy FMs or MLLMs for PAD and MAD. To this end, we consider five different approaches with increasing access to a model's internal information (\Cref{fig:generic_pipeline}). In the first approach, we focus on zero-shot evaluation of 16 open-weight general-purpose MLLMs as off-the-shelf models and assess their performance for PAD and MAD tasks. We provide each model with an input image and a textual prompt asking it to provide a score indicating whether the image is bonafide or an attack. In the second approach, we focus on the next-token probability at the output of MLLMs and train a shallow model to predict PAD or MAD scores based on token logit probabilities. In our third approach, we adapt general-purpose MLLMs for PAD and MAD with parameter-efficient fine-tuning. We generate separate question-answer datasets for PAD and MAD tasks, and then fine-tune an MLLM for each task. The resulting models (PADLLM and MADLLM) improve performance compared to pretrained baselines. In the next approaches, we focus on vision-encoders in MLLMs and FMs and use them as feature extractors. We first evaluate the performance of 30 different vision-encoders when used as frozen feature extractors with a linear-probing protocol (our fourth approach). Finally, as our fifth approach, we fine-tune and adapt the vision-encoders and train them for PAD and MAD. We evaluate the models on various datasets, including MSU-MFSD, CASIA-FASD, Replay-Attack, and OULU-NPU presentation-attack datasets for PAD, and the FFHQ, FRGC, FRLL, and FERET datasets for MAD. In summary, our contributions are as follows:
\begin{itemize}
    \item We present a comprehensive study of using foundation models and multimodal large language models for presentation and morph attack detection.
    
    \item We present two new specialized MLLMs, called PADLLM and MADLLM. The trained models achieve state-of-the-art performance and provide reasoning for attack detection.

    \item We improve the performance of MLLMs based on next-token probability.

    \item We fine-tune vision-encoders of FMs and MLLMs. The resulting models achieve strong detection performance.
    
    \item We benchmark 16 open-weight MLLMs and 30 frozen vision-encoder backbones, and explore which pretraining can better represent PAD-related and MAD-relevant information.

\end{itemize}

%% file: content/figures/pipeline_figure.tex
\definecolor{pipink}{HTML}{15181D}
\definecolor{pipmuted}{HTML}{5B6270}
\definecolor{piphair}{HTML}{DFE3EA}
\definecolor{colvis}{HTML}{2F5FC7}   
\definecolor{coltxt}{HTML}{8A4BC4}   
\definecolor{collog}{HTML}{00808F}   
\definecolor{colout}{HTML}{B0691A}   
\newcommand{\figfam}{\rmfamily}
\newcommand{\figS}{\scriptsize}
\newcommand{\figT}{\fontsize{6}{7.0}\selectfont}

\newcommand{\faceslot}[5]{%
  \coordinate (fsA) at (#1,#2);
  \coordinate (fsB) at ([shift={(#3,-#4)}]fsA);
  \begin{scope}
    \clip[rounded corners=1.5pt] (fsA) rectangle (fsB);
    \node[inner sep=0pt, outer sep=0pt] at ($(fsA)!0.5!(fsB)$)
      {\includegraphics[width=#3cm]{#5}};
  \end{scope}
  \draw[black!30, line width=0.3pt, rounded corners=1.5pt]
    (fsA) rectangle (fsB);}

\newcommand{\faceslotL}[5]{%
  \coordinate (fsA) at (#1,#2);
  \coordinate (fsB) at ([shift={(#3,-#4)}]fsA);
  \begin{scope}
    \clip[rounded corners=1.5pt] (fsA) rectangle (fsB);
    \node[inner sep=0pt, outer sep=0pt] at ($(fsA)!0.5!(fsB)$)
      {\includegraphics[height=#4cm]{#5}};
  \end{scope}
  \draw[black!30, line width=0.3pt, rounded corners=1.5pt]
    (fsA) rectangle (fsB);}

\newcommand{\layerbox}[5]{%
  \node[anchor=north west, draw=#4!40, fill=white, line width=0.35pt,
        rounded corners=1.2pt, minimum width=#3cm, minimum height=0.48cm,
        inner sep=1pt, text width={#3cm-0.24cm}, align=center,
        font=\figfam\figT, text=black!80] at (#1,#2) {#5};}

\newcommand{\callout}[7]{%
  \node[anchor=north west, draw=#4!35, fill=#4!5, line width=0.4pt,
        rounded corners=2pt, minimum width=#3cm, minimum height=1.70cm,
        inner sep=0pt] at (#1,#2) {};
  \fill[#4] (#1,#2) rectangle ++(0.1,-1.70);
  \node[circle, fill=#4, text=white, inner sep=0pt, minimum size=3.4mm,
        font=\figfam\bfseries\figT, anchor=north west]
        at ($(#1,#2)+(0.26,-0.15)$) {#5};
  \node[anchor=base west, font=\figfam\bfseries\figS, text=#4!80!black]
        at ($(#1,#2)+(0.70,-0.40)$) {#6};
  \node[anchor=north west, text width={#3cm-0.5cm}, align=left,
        font=\figfam\figT, text=black!72]
        at ($(#1,#2)+(0.26,-0.60)$) {#7};}

\newcommand{\upcue}[4]{\draw[-{Triangle[width=3.4pt,length=3.4pt]},
        line width=0.8pt, #4] (#1,#2) -- (#1,#3);}

\begin{tikzpicture}[x=1cm, y=1cm, font=\figfam]

\callout{3.135}{8.90}{3.80}{colvis}{E}{Vision-encoder adaptation}%
  {Encoders of MLLMs fine-tuned end-to-end for PAD and MAD, achieving strongest cross-dataset detection.}
\upcue{5.035}{7.20}{6.85}{colvis}

\callout{7.225}{8.90}{3.80}{coltxt}{C}{MLLM adaptation}%
  {Parameter-efficient fine-tuning on task-specific question/answer data yielding \textbf{PADLLM} and \textbf{MADLLM}.}
\upcue{9.125}{7.20}{6.85}{coltxt}

\node[anchor=north west, draw=piphair, fill=black!1.5, line width=0.4pt,
      rounded corners=3pt, minimum width=2.85cm, minimum height=4.70cm,
      inner sep=0pt] at (0.0,6.8) {};
\node[anchor=north west, font=\figfam\bfseries\figS, text=pipink]
      at (0.15,6.70) {INPUT IMAGES};

\node[anchor=north west, font=\figfam\figT, text=pipmuted]
      at (0.15,6.34) {PAD $\cdot$ bonafide / attack};
\faceslot{0.15}{6.02}{1.23}{1.23}{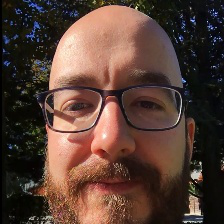}
\faceslot{1.46}{6.02}{1.23}{1.23}{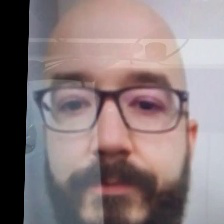}

\node[anchor=north west, text width=2.60cm, align=left,
      font=\figfam\figT, text=pipmuted]
      at (0.15,4.73) {MAD $\cdot$ bonafide 1 $+$ bonafide 2 $\rightarrow$ morph};
\faceslot{0.15}{4.11}{0.85}{0.85}{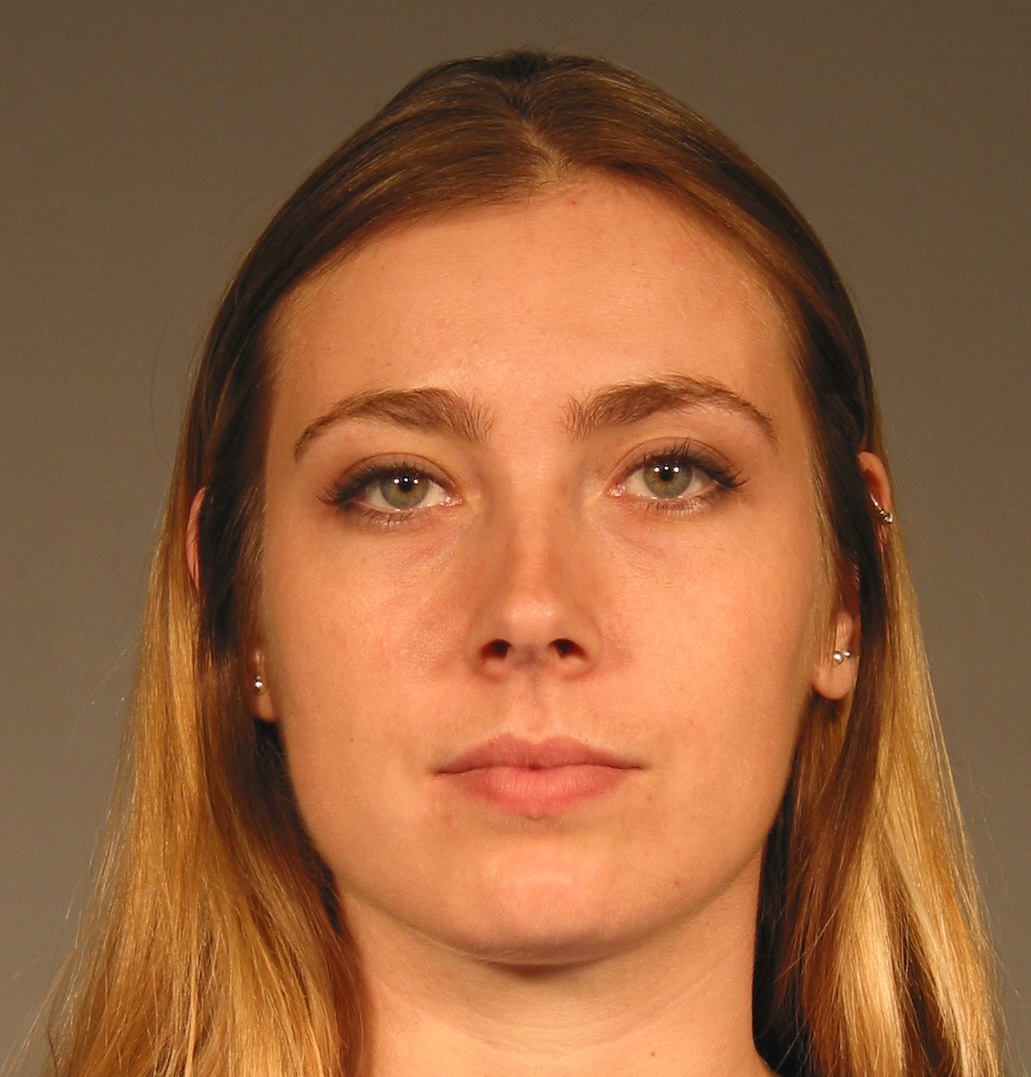}
\faceslot{1.06}{4.11}{0.85}{0.85}{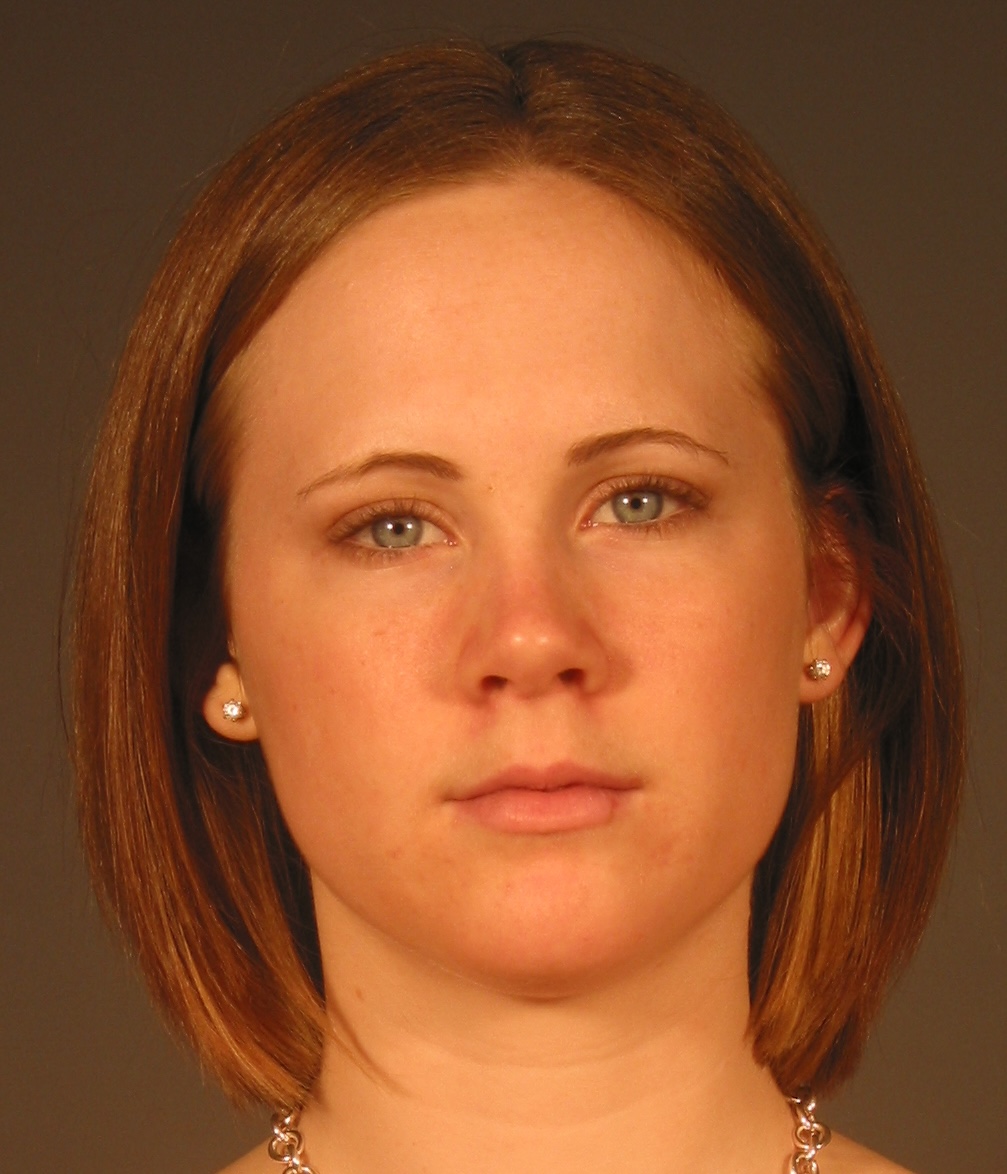}
\faceslot{1.97}{4.11}{0.85}{0.85}{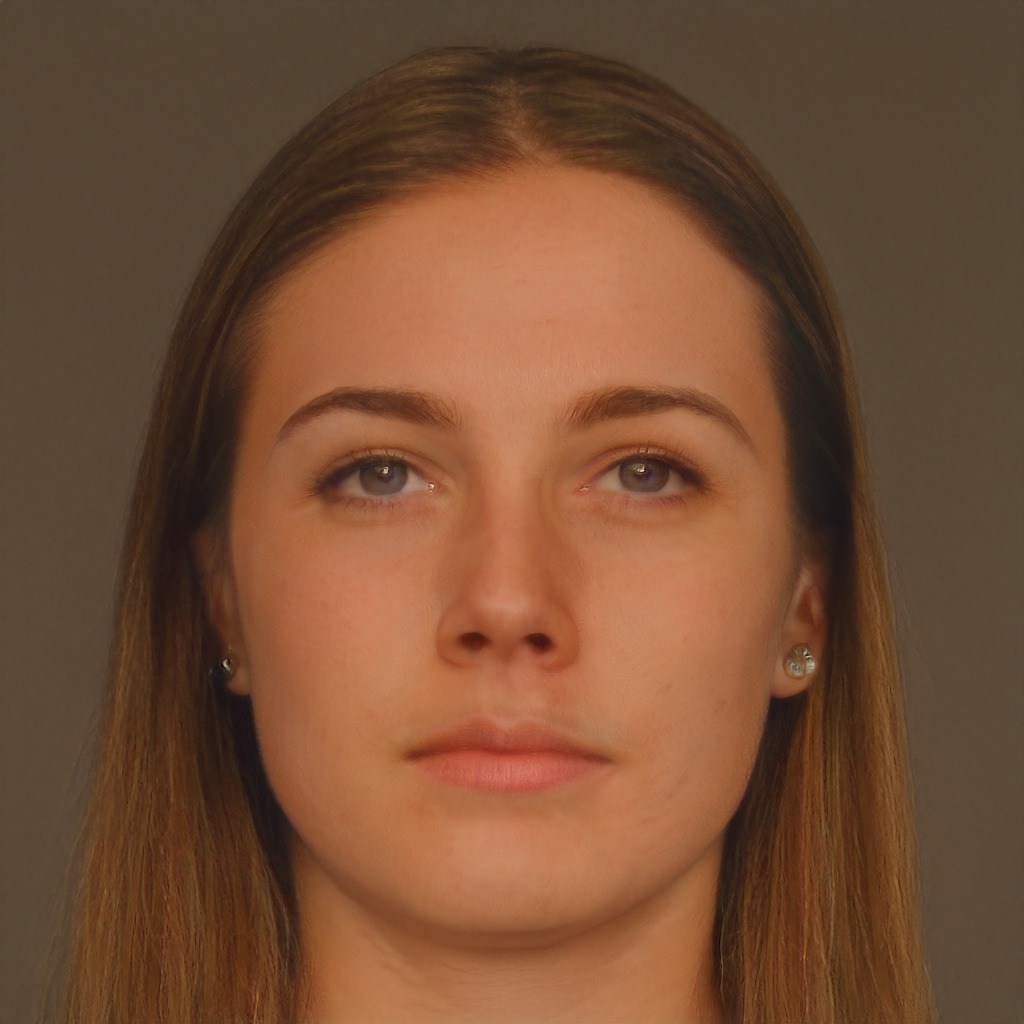}

\node[anchor=north west, draw=black!25, dashed, line width=0.35pt,
      rounded corners=2pt, minimum width=2.55cm, minimum height=0.55cm,
      inner sep=3pt, text width=2.30cm, align=left,
      font=\figfam\figT, text=pipmuted] at (0.15,3.18)
      {$+$ \textbf{task prompt}: ``\textit{bonafide or attack? give a score 0--1}''};

\draw[-{Triangle[width=3.4pt,length=3.4pt]}, line width=0.8pt, black!35]
      (2.95,5.02) -- (3.25,5.02);

\node[anchor=north west, draw=black!28, dashed, line width=1.00pt,
      rounded corners=4pt, minimum width=10.95cm, minimum height=3.55cm,
      inner sep=0pt] at (3.30,6.8) {};
\node[anchor=east, draw=black!28, fill=black!5, text=black, rounded corners=1.5pt,
      inner xsep=4pt, inner ysep=2pt, font=\figfam\bfseries\normalsize]
      at (9.65,3.20) {MLLM};

\node[anchor=north west, draw=colvis!35, fill=colvis!4, line width=0.4pt,
      rounded corners=3pt, minimum width=3.27cm, minimum height=3.00cm,
      inner sep=0pt] at (3.40,6.60) {};
\node[anchor=north west, font=\figfam\bfseries\figS, text=colvis!80!black]
      at (3.55,6.45) {VISION COMPONENT};
\layerbox{3.55}{5.77}{2.97}{colvis}{Patch embedding}
\layerbox{3.55}{5.15}{2.97}{colvis}{ViT transformer blocks}
\layerbox{3.55}{4.53}{2.97}{colvis}{Visual token features}

\node[anchor=north west, draw=piphair, fill=black!1.5, line width=0.4pt,
      rounded corners=3pt, minimum width=0.60cm, minimum height=3.00cm,
      inner sep=0pt] at (6.78,6.60) {};
\node[rotate=90, font=\figfam\figS, text=pipmuted] at (7.08,5.00)
      {PROJECTOR (MLP)};

\node[anchor=north west, draw=coltxt!35, fill=coltxt!4, line width=0.4pt,
      rounded corners=3pt, minimum width=3.27cm, minimum height=3.00cm,
      inner sep=0pt] at (7.49,6.60) {};
\node[anchor=north west, font=\figfam\bfseries\figS, text=coltxt!80!black]
      at (7.64,6.45) {TEXT COMPONENT};
\layerbox{7.64}{5.77}{2.97}{coltxt}{Tokenizer \& text embedding}
\layerbox{7.64}{5.15}{2.97}{coltxt}{LLM decoder blocks}
\layerbox{7.64}{4.53}{2.97}{coltxt}{Language modelling head}

\node[anchor=north west, draw=collog!40, fill=collog!5, line width=0.4pt,
      rounded corners=3pt, minimum width=3.27cm, minimum height=3.00cm,
      inner sep=0pt] at (10.87,6.60) {};
\node[anchor=north west, font=\figfam\bfseries\figS, text=collog!80!black]
      at (11.02,6.45) {LOGITS LAYER};
\node[anchor=north west, text width=2.97cm, align=left,
      font=\figfam\figT, text=black!72] at (11.02,5.97)
      {Next-token distribution $\mathbf{z}$: the layer immediately before the
       text output.};
\node[anchor=base west, font=\figfam\figT, text=black!75] at (11.02,4.90) {``yes''};
\fill[collog!75, rounded corners=0.5pt] (11.95,4.86) rectangle ++(0.95,0.11);
\node[anchor=base west, font=\figfam\figT, text=black!75] at (11.02,4.45) {``no''};
\fill[collog!75, rounded corners=0.5pt] (11.95,4.41) rectangle ++(0.36,0.11);
\node[anchor=base west, font=\figfam\figT, text=black!75] at (11.02,4.00) {``\_No''};
\fill[collog!75, rounded corners=0.5pt] (11.95,3.96) rectangle ++(0.58,0.11);

\draw[-{Triangle[width=3.4pt,length=3.4pt]}, line width=0.8pt, black!35]
      (14.25,5.02) -- (14.55,5.02);

\node[anchor=north west, draw=colout!35, fill=colout!5, line width=0.4pt,
      rounded corners=3pt, minimum width=3.3cm, minimum height=3.55cm,
      inner sep=0pt] at (14.50,6.8) {};
\node[anchor=north west, font=\figfam\bfseries\figS, text=colout!80!black]
      at (14.68,6.66) {TEXTUAL OUTPUT};
\node[anchor=north west, draw=colout!30, fill=white, line width=0.35pt,
      rounded corners=2pt, minimum width=2.95cm, inner sep=4pt,
      text width=2.60cm, align=left, font=\ttfamily\figT, text=black!80]
      at (14.68,6.32)
      {score: 0.08\\ ``the face shows moir\'e patterning and a flat specular
       response, consistent with a screen replay.''};
\node[anchor=north west, text width=2.95cm, align=left,
      font=\figfam\figT, text=black!72] at (14.68,4.44)
      {A bonafide score in $[0,1]$, plus natural-language reasoning from the fine-tuned models.};

\upcue{5.035}{2.90}{3.25}{colvis}
\callout{3.135}{2.90}{3.80}{colvis}{D}{Linear probing}%
  {Frozen vision encoders used as feature extractors; only a linear head is trained.}
\upcue{12.505}{2.90}{3.25}{collog}
\callout{10.605}{2.90}{3.80}{collog}{B}{Logits sampling}%
  {A shallow classifier trained on the next-token logit probabilities instead
   of the decoded text.}
\upcue{16.15}{2.90}{3.25}{colout}
\callout{14.50}{2.90}{3.30}{colout}{A}{Zero-shot benchmark}%
  {Off-the-shelf MLLMs prompted for a score, reading only the text output.}
\node[anchor=north west, draw=piphair, fill=black!1.5, line width=0.4pt,
      rounded corners=2pt, minimum width=17.8cm, minimum height=0.62cm,
      inner sep=0pt] at (0,1.05) {};
\node[text=white, inner sep=0pt, minimum size=3.4mm,
      font=\figfam\bfseries\figT] at (0.40,0.74) {~};
\node[anchor=base west, font=\figfam\bfseries\figS, text=pipink]
      at (0.1,0.67) {BENCHMARK SCALE};
\draw[piphair, line width=0.5pt] (3.15,0.51) -- (3.15,0.97);
\node[anchor=base west, font=\figfam\figS, text=black!72]
      (bmcounts) at (3.2,0.67)
      {\textbf{16} open-weight MLLMs \enspace \textbf{30} vision encoders
       \enspace \textbf{4} PAD $+$ \textbf{4} MAD datasets};
\node[anchor=base west, font=\figfam\figS, text=black!72, inner sep=0pt,
      xshift=0.12cm] (bmproto) at (bmcounts.east |- 0,0.67) {Protocols:};
\node[anchor=west, font=\figfam\figS, text=black!72, align=left,
      inner sep=0pt, xshift=0.12cm] (bmstack) at (bmproto.east |- 0,0.74)
      {\scalebox{0.78}{Intra-dataset} (\textbf{ID}),
       \scalebox{0.78}{Cross-dataset} (\textbf{CD})};
      
\end{tikzpicture}

%% file: content/02_related_work.tex
\section{Related Work}
\subsection{Foundation Models}
Foundation models (FMs) have emerged as an important component of modern machine learning systems.
The foundation model refers to models that are pretrained on large-scale data and can be adapted to a wide range of downstream tasks \cite{bommasani2021opportunities}. 
%
Contrastive image--text models include CLIP~\cite{radford2021learning}, SigLIP~\cite{siglip2023}, and EVA-CLIP~\cite{sun2023eva}.
Self-distillation and masked-token objectives include DINOv2~\cite{oquab2023dinov2}, DINOv3~\cite{oquab2024dinov3}, and BEiT~\cite{bao2022beit}.
Masked-image reconstruction uses ViT-MAE~\cite{he2022mae}; neighbour-based masking uses ViT-MSN~\cite{assran2023msn}.
InternViT towers inherit multimodal InternVL pretraining at $448{\times}448$ resolution~\cite{chen2024internvl2}.
Large language models (LLMs) are another type of FMs that are trained on a very large corpus of textual data using self-supervised learning with mask language modeling loss~\cite{naveed2025comprehensive,raiaan2024review}. These models can be used for text generation and in different downstream tasks. Compared to LLMs, which only take text input, multimodal large language models (MLLMs) can take different modalities as input (such as image, audio,  etc.) in addition to a text prompt~\cite{zhang2024lvlms}. 

Recently, several papers have explored applications of FMs and MLLMs in face biometric recognition and security. A recent survey \cite{shahreza2025foundation} provides an extensive
review of how FMs and MLLMs are used in face biometrics.
Early studies investigated the application of  MLLMs, such as ChatGPT~\cite{hurst2024gpt}, for face recognition~\cite{hassanpour2024chatgpt}, and predicting soft-biometric attributes, such as gender, age, and ethnicity.  
Jia \textit{et al.} \cite{jia2024can} also explored the application of ChatGPT for zero-shot face deepfake detection.  Shi \textit{et al.} \cite{shi2024shield} investigated chain-of-thoughts prompting for  ChatGPT and Gemini in face presentation attack and deepfake detection. 
Sony~\textit{et al.}~\cite{sony2025benchmarking} benchmarked foundation models across a broader set of zero-shot biometric tasks. 
MLLMs were also benchmarked for face recognition~\cite{ozturk2026demographic,shahreza2026benchmarking} and heterogeneous face recognition~\cite{shahreza2026evaluating}. 
FRoundation adapted CLIP and DINO  backbones for face recognition~\cite{chettaoui2024froundation}. 
FaceLLM~\cite{shahreza2025facellm} adapted InternVL3~\cite{zhu2025internvl3} with the FairFaceGPT dataset for general face analysis tasks.

\subsection{Presentation Attack Detection (PAD)}
\label{sec:rel:pad}

PAD methods use texture, color, and frequency cues to distinguish live skin from presentation attack (e.g., printed or replayed media)~\cite{liu2018learning, DBLP:journals/tifs/LiLCWHK18, Fang_2022_WACV}. 
Pixel-wise supervision~\cite{liu2018learning} and depth-aware architectures~\cite{Fang_2022_WACV} train compact convolutional neural networks end-to-end on small-scale data and can reach strong intra-dataset average classification error rate (ACER). 
Frequency-domain and multi-scale techniques specifically target print-and-replay artifacts \cite{DBLP:journals/pr/FangDKK22}.
Patch-based formulations such as PatchNet~\cite{DBLP:conf/cvpr/WangLYL22} treat spoof detection as fine-grained local discrimination.
These networks serve as practical baselines for compact deployment because of their lightweight, dataset-specific design.
However, they frequently overfit to acquisition conditions and exhibit performance degradation under cross-domain shifts~\cite{DBLP:journals/tcsv/YanZH22,fang2024face}.

Deep neural network-based models achieved strong performance within individual datasets, but experience significant performance degradation when evaluated under cross-dataset shifts~\cite{DBLP:journals/tbbis/WangWDG22,DBLP:journals/tcsv/YanZH22, fang2024face}.
In cross-dataset evaluation, the model is trained on one dataset and tested on another, exposing sensor and illumination shifts~\cite{DBLP:conf/icmcs/LiuCDLZX22,fang2024face}.
Cross-database evaluation reveals domain shifts among MSU-MFSD, CASIA-FASD, Replay-Attack, and OULU-NPU~\cite{wen2015face, patel2016secure, costa2016replay, boulkenafet2017oulu}.
%
Unseen-attack protocols are designed to evaluate model robustness against previously unencountered presentation attacks (PAs) during testing.
Using the leave-one-out protocol on MCIO~\cite{DBLP:conf/cvpr/ShaoLLY19} and limited source domain protocols~\cite{feng2026benchmarking} measure generalization when multiple labeled sources are available at training time.

Foundation models offer an alternative for training PAD networks from scratch on small MCIO datasets.
FoundPAD~\cite{ozgur2025foundpad} adapts CLIP~\cite{radford2021learning} with Low-Rank Adaptation (LoRA)~\cite{hu2021lora}.
FSFM~\cite{wang2025fsfm} pretrains a ViT-B/16~\cite{dosovitskiy2021vit} with face-spoofing-specific self-supervised learning before downstream fine-tuning.
FLIP~\cite{srivatsan2023flip} guides cross-domain PAD with CLIP language embeddings.
CFPL-FAS~\cite{liu2024cfpl}, optimal-transport adaptation~\cite{li2025optimal}, frequency-shortcut modelling~\cite{cao2025towards}, and DADM~\cite{yang2025dadm} engineer cross-domain generalization with prompts, transport, frequency views, or dual alignment.
Gonzalez-Soler~\textit{et al.}~\cite{gonzalez2025foundation} study zero-shot PAD with pretrained ViTs. Komaty~\textit{et al.}~\cite{komaty2025exploring} explored zero-shot and in-context learning for PAD with ChatGPT.  
Feng~\textit{et al.}~\cite{feng2026benchmarking} benchmark frozen and fine-tuned FMs, showing that DINOv2 with register tokens~\cite{darcet2024registers} and PAD-specific augmentations closes much of the gap to larger pipelines. 
In \cite{lorenz2026foundation}, multiple vision encoders were frozen and used as a feature extractor, and then a head was trained for PAD. 
We expand this benchmark by extracting more vision encoders from VLMs and MAD datasets, and compare the results with stronger baselines.

\subsection{Morph Attack Detection (MAD)}
\label{sec:rel:mad}
A face morph fuses two identities' images into a single enrollment photo that a face-recognition matcher accepts as either contributor, a threat first flagged for automated border control and passport issuance~\cite{schwaiger2022man}.
Early morphs align facial landmarks and warp-blend the two source images; landmark-based pipelines remain a strong, low-cost attack~\cite{Sarkar2020,venkatesh2020can}.
Generative models produce higher-fidelity morphs: MIPGAN embeds an identity-prior loss in a StyleGAN generator to keep both contributing identities verifiable\cite{zhang2021mipgan}, while MorDIFF morphs in a diffusion autoencoder's latent space\cite{damer2023mordiff}.
Comparative studies show that GAN-based morphs can threaten matchers as much as, or more than, landmark-based ones while leaving different, harder-to-detect artefacts~\cite{sarkar2022gan,venkatesh2020can}.

We focus on single-image MAD (S-MAD), which flags a morph from one enrollment photo alone, without a trusted live reference capture at the gate.
S-MAD methods extract deep representations tuned to expose blending artefacts: 
OrthoMAD disentangles identity from spoof-relevant features via an orthogonality constraint~\cite{neto2022orthomad}, 
SelfMAD replaces morph-specific supervision with self-supervised pretext tasks for better cross-attack generalization~\cite{ivanovska2025selfmad}, and Zhang~\textit{et al.}~\cite{zhang2024generalized} probe frozen vision-transformer representations for single-image detection.
Colbois~\textit{et al.}~\cite{colbois2024evaluating} evaluate how well features tuned for one morphing technique (landmark-based, GAN-based, or diffusion-based) transfer to another, finding that attack-specific detectors generalize poorly across morph generation methods (\textit{i.e.,} an unseen-attack analogue of the cross-database gap that plagues PAD).

Recently, some studies probe general-purpose MLLMs (such as ChatGPT/GPT-4V or open-weight models) on face MAD. 
Zhang~\textit{et al.}~\cite{zhang2025chatgpt} and Maric~\textit{et al.}~\cite{maricexploring} evaluate ChatGPT-style and open MLLMs on zero-shot MAD, and Patwardhan~\textit{et al.}~\cite{patwardhan2024empowering} adapt an image--text foundation model for interpretable, language-grounded MAD decisions.
Ivanovska~\textit{et al.}~\cite{ivanovska2026emergent} show that open MLLMs exhibit emergent morph sensitivity without any MAD-specific training.

Although investigating MLLMs for both PAD and MAD is an active area of research, to the best of our knowledge, no existing work explores their capabilities across multiple access levels. 
This work aims to fill this gap by presenting a comprehensive study covering: zero-shot prompting, parameter-efficient fine-tuning, next-token logit sampling, and the deployment of both frozen and fine-tuned vision encoders.

%% file: content/03_FMs_4_PAD_MAD.tex
\section{Foundation Models for Presentation and Morph Attack Detection}\label{sec:method}

In this section, we present five different approaches to deploy FMs and MLLMs with increasing access to a model's internal information for PAD and MAD tasks. 
The full pipeline is illustrated in~\cref{fig:generic_pipeline}. 

\subsection{Zero-shot MLLMs}\label{sec:zeroshot}

Our goal, in this approach, is to measure how well multimodal large language models (MLLMs) can perform as face attack detectors without any task-specific training. We therefore treat every model as an off-the-shelf and blackbox model that is queried once per image with a similar instruction prompt, and we convert its
answer into a scalar detection score. 

The presentation attack and morph attack detection share the same abstract structure: given a face image, decide whether it is a \emph{bonafide} sample of a single live subject, or an \emph{attack} sample produced by some attack instrument. We exploit this shared structure to define one protocol for both
tasks. Therefore, a detector assigns each image a \emph{bonafide score} between $0$ and $1$, where higher values indicate stronger evidence that the sample is bonafide. 
Typically, PAD and MAD models predict this score from a network trained on labelled data from the target domain. We instead obtain it by prompting a frozen MLLM without any fine-tuning. The model must therefore rely entirely on the priors acquired during pre-training,
together with whatever notion of ``printed photo'', ``screen replay'' or ``morphed face'' its language model already encodes. This is the minimum setting in which a practitioner with no PAD or MAD training data would deploy an off-the-shelf MLLM.

The most direct way to query an MLLM for a binary decision is to ask for a binary answer, but standardized PAD and MAD metrics are defined over a score distribution and an operating threshold, not over hard labels. We therefore ask the model for a \emph{score} rather than a binary class, which lets us report standard metrics.

We design the following prompts for PAD and MAD, respectively, which differ only in the description of the attack:
\vspace{0.5em}
\noindent\fbox{\parbox{0.97\linewidth}{\small
\textbf{PAD prompt:} \emph{Is the face in this image a real person (bonafide) or a presentation attack (e.g.\ printed photo or screen replay)?
Answer with a single score between 0 and 1, where 1 means definitely bonafide and 0 means definitely an attack. Output only the score number, nothing else.}
}}
\vspace{0.4em}

\noindent\fbox{\parbox{0.97\linewidth}{\small
\textbf{MAD prompt:} \emph{Is the face in this image a genuine photo of a single real person (bonafide) or a morphed face image that blends two different identities (morph attack)? Answer with a single score between 0 and 1, where 1 means definitely bonafide and 0 means definitely a morph. Output only the score
number, nothing else.}
}}
\vspace{0.5em}

The two prompts are deliberately parallel and differ only in how the attack is described, so that any difference between our PAD and MAD results reflects the task and the data rather than the phrasing. Each prompt is paired with exactly one image and issued as an independent request, so no information leaks between
test samples and the order of evaluation is irrelevant. 
We then parse the MLLM output to obtain a score in $[0,1]$ and compute the standardized metrics for PAD and MAD.

\subsection{Language Model Sampling Logits}
\label{sec:logit-sampling}

In our previous approach, we relied on the MLLM to generate a classification score as part of its textual output. 
However, this approach discards the rich probabilistic information contained in the model's output distribution. Inspired by prior work~\cite{ivanovska2026emergent} that probes LLM token probabilities for classification, we instead extract the raw next-token logits and use them directly as classification features.



Given an input image $\mathbf{I}$ and a prompt $\mathbf{P}$, we get the MLLM output token-level probabilities. We use the following query prompt $\mathbf{P}$ for MAD. The PAD query follows the same template, with "morph" replaced by "presentation".


\vspace{0.5em}
\noindent\fbox{\parbox{0.97\linewidth}{\small
\textbf{Prompt:} \emph{Analyze this face image and decide if it is a morph attack or bonafide. Answer with exactly one label: `yes' for bonafide and `no' for morph attack. Do not provide any additional text.}
}}
\vspace{0.4em}

The MLLM, comprising the vision-encoder $f_\theta$ and the language model $g_\phi$, produces a conditional distribution over the next output token: $p(\cdot \mid \mathbf{I}, \mathbf{P}) = g_\phi\!\left(f_\theta(\mathbf{I}), \mathbf{P}\right)$


Rather than sampling from the full output distribution, we identify two disjoint sets of semantically relevant tokens: an affirmative set with $N_{\text{yes}} = 15$ variants $\mathcal{V}^{+} = \{v^{+}_1, \ldots, v^{+}_{15}\}$ (\textit{e.g.}, ``Yes'', ``yes'', ``YES'', ':YES','.Yes','=yes','\_YES', \ldots), and a negative set with $N_{\text{no}} = 25$ variants $\mathcal{V}^{-} = \{v^{-}_1, \ldots, v^{-}_{25}\}$ (\textit{e.g.}, ``No'', ``no'', ``NO'', '=no', '/no', '\_No', ':NO', '"No', '(no', \ldots). We build a logit vector $\mathbf{z}$ with the token values in these two sets. 
The corresponding logit vector for these tokens is given by: $\mathbf{z} = \big[\ell_{v^{+}_1}, \ldots, \ell_{v^{+}_{15}}, \ell_{v^{-}_1}, \ldots, \ell_{v^{-}_{25}}\big] \in \mathbb{R}^{K}$,
where $\ell_v = \left[g_\phi\!\left(f_\theta(\mathbf{x}), \mathbf{p}\right)\right]_v$ denotes logit for token $v$. The combined set size is $K = N_{\text{yes}} + N_{\text{no}} = 40$. 
The logit vector $\mathbf{z}$ serves as the feature representation for downstream classification.
We use a fully connected network as a classifier, consisting of three hidden layers with 128, 64, and 32 units. Each hidden layer is followed by Batch Normalization, ReLU activation, and a dropout rate of 0.3. The classifier determines if the image is an attack or bonafide based on the logit vector $\mathbf{z}$. We train separate classifiers for PAD and MAD tasks.

\subsection{LoRA Adaption MLLMs}
\label{sec:fine-tuning-mllms}
To adapt an MLLM to face-biometric tasks such as MAD and PAD, we developed task-specific instruction-tuning datasets, 
and we used them to fine-tune a pretrained MLLM. 
Our data preparation strategy combined newly generated task-specific question-answer (QA) pairs with an existing general facial-attribute description dataset (FairFaceGPT~\cite{shahreza2025facellm}).

As a foundation for general facial reasoning, we utilize FairFaceGPT~\cite{shahreza2025facellm}, an existing dataset comprising 87,636 QA pairs based on FairFace~\cite{karkkainen2021fairface} dataset. For each image, the dataset provides descriptive QA pairs covering demographic attributes (age, gender, and race); facial structure, skin texture, expression, and emotion; lighting and image quality; face pose; forensic considerations; and general image descriptions. 
Combining this with task-specific data gives the model a broader ability to describe and reason about faces, rather than restricting its training to attack detection.

MAD  and PAD  QA pairs are generated programmatically from the training split, without utilizing any language model. Each image in the training data has a ground-truth label indicating whether it is bonafide (genuine) or an attack. For attack samples, an additional sub-category specifies the presentation-attack instrument for PAD or the morphing technique for MAD.
For each image, we create a set of question templates that include several binary or forced-choice questions designed to assess the bonafide or attack status. 
These questions can take the form of Yes/No queries or choices, such as \textit{Live}/\textit{Spoof} for PAD and \textit{One Identity}/\textit{Two Identities} for MAD. We also include an operational question asking whether the image should be rejected in a specific context, such as an access-control gate for PAD or an identity-document application for MAD.
Additionally, an attack-category question prompts the model to identify the corresponding instrument or technique used in the attack, with bonafide samples being assigned the answer "None.". 
For attack samples, we further include a question that classifies the attack into broader categories: paper versus display for PAD, and landmark-based versus other methods for MAD.

In addition, we create questions that involve image pairs. 
For each pair, we generate a four-option multiple-choice question describing the pair's composition in terms of bonafide and attack images. This covers cases where both images are bonafide, both are attacks, or only one is an attack. When both images are attacks, we additionally generate a Yes/No question to determine whether they share the same instrument or technique. Following that, we create a four-option question to identify which image contains a specific sub-technique. 
We randomly order the answer choices in all multiple-choice questions to prevent the model from relying on the position of the correct answer.

For MAD, we apply this procedure to the FFHQ (bonafide) and FFHQ-Morphs (attack) training split, and group the five morphing techniques into three broader families, \textit{i.e.,} landmark-based (two variants), StyleGAN latent-interpolation (two variants), and diffusion-based (one variant). The resulting dataset includes 150{,}502 MAD QA pairs. After combining these samples with the FairFaceGPT corpus, the final training set contains 238{,}138 samples. \Cref{tab:mad-qa-examples} provides representative examples of the different MAD question types.

\providecommand{\qaicon}[1]{%
  \begin{tikzpicture}[baseline=-0.45ex]
    \node[draw=#1,line width=0.9pt,minimum width=0.38cm,minimum height=0.38cm,inner sep=0pt,fill=gray!12] (b) {};
    \draw[gray!55,line width=0.35pt] (b.center) circle (0.12cm);
    \fill[gray!55] ($(b.center)+(-0.045,0.03)$) circle (0.011cm);
    \fill[gray!55] ($(b.center)+(0.045,0.03)$) circle (0.011cm);
  \end{tikzpicture}%
}
\providecommand{\bonaimg}{\qaicon{green!55!black}}
\providecommand{\attackimg}{\qaicon{red!70!black}}

\begin{table}[t]
  \centering
  \caption{
  MAD. Example instruction-tuning QA pairs generated. 
  \protect\bonaimg~bonafide and \protect\attackimg~attack.}
  \label{tab:mad-qa-examples}
  \footnotesize
  \begin{tabular}{@{}p{0.13\linewidth}p{0.58\linewidth}p{0.16\linewidth}@{}}
    \toprule
    Image(s) & Question & Answer \\
    \midrule
    \rowcolor{gray!10}
    \bonaimg & Is this image a morphing attack? (Yes/No) & No \\
    \attackimg & Does this image show a single, unaltered identity? (Yes/No) & No \\
    \rowcolor{gray!10}
    \attackimg & This image was submitted for an identity document application. Should it be rejected as a morphing attack? (Yes/No) & Yes \\
    \attackimg & Which technique was used to create this image? (None/Landmark/GAN/Diffusion) & Landmark \\
    \rowcolor{gray!10}
    \attackimg & Was this morphing attack produced by landmark warping or by a generative model? (Landmark/Generative) & Landmark \\
    \bonaimg\,\bonaimg & Which statement best describes the images? A)~Both bonafide B)~Both attacks C)~1st bonafide, 2nd attack D)~1st attack, 2nd bonafide & A \\
    \rowcolor{gray!10}
    \attackimg\,\attackimg & Were these two morphing attacks created with the same morphing technique? (Yes/No) & No \\

    \bottomrule
  \end{tabular}
\end{table}

The same procedure applied to the OULU-NPU training split produces 55{,}534 PAD QA pairs. Here, the presentation-attack instruments are grouped into two broader categories---printed photo and video replay---with each category covering two capture devices. After combining the PAD samples with the FairFaceGPT corpus, the final training set contains 143{,}170 samples. \Cref{tab:pad-qa-examples} presents representative examples covering the different PAD question types.

\begin{table}[t]
  \centering
  \caption{
  PAD. Example instruction-tuning QA pairs generated. \protect\bonaimg~bonafide and \protect\attackimg~attack.
  }
  \label{tab:pad-qa-examples}
  \footnotesize
  \begin{tabular}{@{}p{0.13\linewidth}p{0.58\linewidth}p{0.16\linewidth}@{}}
    \toprule
    Image(s) & Question & Answer \\
    \midrule

    \rowcolor{gray!10}
    \attackimg & Is this image a presentation attack? (Yes/No) & Yes \\

    \bonaimg & Is this a bonafide image? (Yes/No) & Yes \\

    \rowcolor{gray!10}
    \bonaimg & This image was captured by a face verification system at an access control gate. Should it be rejected as a presentation attack? (Yes/No) & No \\

    \attackimg & Which presentation attack instrument was used to produce this image? (None/Printed Photo/Video Replay) & Video Replay \\

    \rowcolor{gray!10}
    \attackimg & Was this presentation attack presented on paper or on a digital display? (Paper/Display) & Paper \\

    \bonaimg\,\attackimg & Which statement best describes the images? A)~Both bonafide B)~Both attacks C)~1st attack, 2nd bonafide D)~1st bonafide, 2nd attack & D \\

    \rowcolor{gray!10}
    \attackimg\,\attackimg & Were these two presentation attacks produced with the same attack instrument? (Yes/No) & Yes \\

    \bottomrule
  \end{tabular}
\end{table}

Building on these instruction-tuning datasets, we fine-tune the InterVL3~\cite{zhu2025internvl3} model with 8B parameters, as a pretrained MLLM, for MAD and PAD in a parameter-efficient manner. We employ Low-Rank Adaptation (LORA)~\cite{hu2021lora} to efficiently adapt the model while keeping most of the pretrained components fixed. Specifically, we keep the vision-encoder frozen and apply LoRA to the language-side transformer, thereby preserving the pretrained visual representations while enabling task-specific adaptation. LoRA introduces trainable low-rank matrices into the attention and feed-forward layers of the transformer. Given a pretrained weight matrix $W \in \mathbb{R}^{d \times k}$, the adapted weight matrix  is defined as
\begin{equation}
\tilde{W} = W + \Delta W = W + \frac{\alpha}{r} AB,
\label{eq:lora}
\end{equation}
where $A \in \mathbb{R}^{d \times r}$ and $B \in \mathbb{R}^{r \times k}$ are trainable low-rank matrices, $r$ denotes the rank of the adaptation, and $\alpha$ is a scaling factor that controls the magnitude of the LoRA update. The factor $\alpha/r$ scales the low-rank update relative to the pretrained weights, allowing the strength of the adaptation to be controlled independently of the pretrained parameterization.
During fine-tuning, only the LoRA parameters  $A$ and $B$ are updated, while the pretrained weights $W$ and the vision-encoder remain frozen. This design substantially reduces the number of trainable parameters, enabling efficient adaptation of the pretrained MLLM to face-attack analysis without modifying the pretrained backbone architecture. The same procedure, with an identical LoRA configuration, is applied independently to the MAD and PAD instruction-tuning datasets, resulting in the MADLLM and PADLLM models, respectively.

\subsection{Vision-encoder as a Feature Extractor with Linear Probe}
\label{sec:linearprobe}

Given that MLLMs often use a vision-encoder along with a language model, we can only focus on the vision-encoder part.
This can be further extended to the vision-encoder of vision foundation models, such as CLIP~\cite{radford2021learning}, SigLIP~\cite{siglip2023}, etc. 
In this approach, we utilize only the vision-encoder component as shown in~\cref{fig:generic_pipeline}. 
Given a facial image $\mathbf{I} \in \mathbb{R}^{H \times W \times 3}$ from dataset $\mathcal{D}$, with a binary label $y \in \{0,1\}$ indicating an attack ($0$) or bonafide sample ($1$), 
the image is passed through a frozen vision-encoder $f_{\theta}$ to obtain a fixed-dimensional feature representation: $\mathbf{e}=f_{\theta}(\mathbf{I}) \in \mathbb{R}^{d}$.
where $d$ denotes the embedding dimension, which may vary depending on the selected vision-encoder. The extracted embeddings capture high-level semantic and structural facial characteristics learned during large-scale multimodal pre-training. We freeze the vision-encoder and use the extracted features to train a classifier head for binary attack detection with the linear-probing. 

\subsection{LoRA Fine-tuning the Vision-encoder as a Feature Extractor}\label{sec:lora_vision_encoder}
While in the previous approach we kept the vision-encoder frozen and only used it as a feature extractor, we can also fine-tune it to improve the performance. Similar to our approach for fine-tuning MLLMs on instruction-tuning datasets, we use LoRA as a parameter-efficient approach for fine-tuning the vision-encoder of FMs. 
For PAD and MAD, the LoRA-adapted backbone maps an input face image $\mathbf{I}$ to a task-specific representation, $\mathbf{h} = f_{\theta,\Delta\theta}(\mathbf{I})$, $p(y=1\mid\mathbf{I}) = \sigma\left(\mathbf{w}^{\top}\mathbf{h}+b\right)$, 
where $\theta$ denotes the frozen pretrained parameters,
$\Delta\theta$ the trainable LoRA parameters,  $(\mathbf{w},b)$ the trainable classification head, and $\sigma$ the sigmoid function. 
Thus, LoRA enables the pretrained representation to be adapted jointly with the classifier while modifying only a small fraction of the backbone parameters.

This setup is particularly suitable for PAD and MAD, where labeled training data are comparatively scarce and full fine-tuning of large FMs can be computationally expensive.
Two representative approaches are FoundPAD~\cite{ozgur2025foundpad} for PAD and MADation~\cite{caldeira2025madation} for MAD.
Both methods adapt pretrained CLIP vision encoders, including ViT-B/16 and ViT-L~\cite{radford2021learning}, to their respective tasks using LoRA while concurrently training a classification head.
These works focus on CLIP-based backbones; however, the extent to which their findings generalize to other FM families remains insufficiently explored.
This is particularly relevant given the broad and expanding range of publicly available foundation models~\cite{li2025survey,shahreza2025foundation}.
Hence, we extend prior studies on this approach by systematically evaluating LoRA-based adaptation across 30 foundation models, including the previously investigated CLIP backbones, for both PAD and MAD.

%% file: content/04_experiments.tex
\begin{table}[tbp]
\centering
\caption{Composition of the evaluation datasets.
\emph{Types} is the number of attack types present in the
test partition. FRLL and FERET have no train or development set.}
\label{tab:databases}
\setlength{\tabcolsep}{3.5pt}
\small
\resizebox{\linewidth}{!}{%
\begin{tabular}{llccccccc}
\toprule
& & \multicolumn{2}{c}{Train}
  & \multicolumn{2}{c}{Dev (threshold)}
  & \multicolumn{2}{c}{Test}
  &  No. Attack \\
\cmidrule(lr){3-4}
\cmidrule(lr){5-6}
\cmidrule(lr){7-8}
Task & Database
  & Bonafide & Attack
  & Bonafide & Attack
  & Bonafide & Attack
  & Types \\
\midrule
\multirow{4}{*}{PAD}
 & M: MSU-MFSD        &    798 &  2{,}398 &    759 &  2{,}278 &    599 &  1{,}799 & 2 \\
 & C: CASIA-FASD      &    900 &  2{,}700 &    300 &     900 & 1{,}800 &  5{,}400 & 3  \\
 & I: Replay-Attack   & 1{,}200 &  5{,}999 & 1{,}200 &  6{,}000 & 1{,}600 &  8{,}000 & 1 \\
 & O: OULU-NPU        & 4{,}800 & 19{,}200 & 3{,}600 & 14{,}399 & 2{,}400 &  9{,}600 & 2\\
\midrule
\multirow{4}{*}{MAD}
 & FFHQ  & 6{,}000 & 15{,}000 & 2{,}000 & 5{,}000 & 2{,}000 & 5{,}000 & 5 \\
 & FRGC  & 6{,}924 &  7{,}530 & 2{,}304 & 2{,}540 & 2{,}304 & 2{,}535 & 5  \\
 & FRLL  & \multicolumn{2}{c}{--} & \multicolumn{2}{c}{--} &    204 & 5{,}700 & 5 \\
 & FERET & \multicolumn{2}{c}{--} & \multicolumn{2}{c}{--} &    791 & 1{,}587 & 3\\
\bottomrule
\end{tabular}
}
\end{table}

\section{Experimental Setup}

\subsection{Dataset}

We evaluate on four PAD and four MAD datasets, summarised in~\cref{tab:databases}, spanning $31{,}198$ and $20{,}121$ test images
respectively. 
The PAD benchmarks include multiple RGB datasets that mostly include print and replay attacks (denoted as MCIO): MSU-MFSD~\cite{wen2015face}, CASIA-FASD~\cite{patel2016secure}, Replay-Attack~\cite{costa2016replay}, and OULU-NPU~\cite{boulkenafet2017oulu}. 
Evaluation is conducted for both intra-dataset (\textit{i.e.,} $M \rightarrow M$, $C \rightarrow C$, $I \rightarrow I$, $O \rightarrow O$) and cross-dataset (\textit{i.e.,} $M \rightarrow C$, $M \rightarrow I$, $M \rightarrow O$, ..., $O \rightarrow$ ...) protocols.
MSU-MFSD
(\textbf{M}) contains videos of printed attacks and replay attacks produced using a mobile phone and a tablet.
CASIA-FASD
(\textbf{C}) comprises warped-photo, cut-photo, and replay attacks conducted under diverse illumination conditions.
Replay-Attack
(\textbf{I}) offers high-resolution print-and-replay spoof attacks performed on a fixed display setup.
OULU-NPU
(\textbf{O}) comprises six mobile phone sensors and multiple print-and-replay attack types, providing the greatest sensor diversity among the datasets.
Collectively, M/C/I/O encompasses laboratory and mobile capture scenarios, multiple presentation attack types, and the cross-dataset shift characteristic of PAD benchmarking~\cite{DBLP:conf/icmcs/LiuCDLZX22,fang2024face}.

For MAD, we use Flicker-Faces HQ (FFHQ) \cite{karras2019style}, Face Research Lab London (FRLL) \cite{debruine2017face}, Face Recognition Grand Challenge (FRGC) \cite{phillips2005overview}, and FERET \cite{phillips1998feret} datasets for evaluation.
FERET and FRGC are widely used benchmark datasets for Morph Attack Detection (MAD) due to their controlled acquisition conditions, including frontal facial pose, neutral expressions, and constrained illumination. 
FRLL provides high-quality frontal face images, whereas FFHQ contains unconstrained facial images collected from Flickr and exhibits substantial variations in pose, illumination, background, and facial appearance. These characteristics make FFHQ particularly suitable for assessing cross-dataset robustness and generalization.
For each dataset,  we consider morph samples generated using multiple morphing algorithms. Similar to \cite{colbois2024evaluating} our morph data in FFHQ, FRGC, and FRLL include landmark-based morphs, GAN-based morphs (SG2-W and SG2-W+), and diffusion-based morphs (MorDIFF), while FERET contains morphs generated using OpenCV (OC), FaceMorpher (FM), and StyleGAN2-based (SG2) methods~\cite{Sarkar2020}. This diversity allows us to evaluate MAD systems across different morphing paradigms and under both controlled and unconstrained conditions.
Together, these datasets provide diverse benchmarks covering multiple morph techniques and acquisition conditions, enabling systematic evaluation of model robustness and generalization. 
\Cref{tab:databases} provides a detailed overview of dataset splits and sample distribution.

\subsection{Evaluation metrics}

We evaluate both PAD and MAD using the Average Classification Error Rate (ACER). 
For simplicity, we  select the operating threshold on the development split using the equal error rate (EER) criterion and then apply it to the test split.
All reported ACER values are in percent (lower is better).

\subsection{Models}
\label{subsec:experiments:models}

We consider $16$ open-weight MLLMs on PAD and MAD, chosen to cover the dominant architectural families and a range of capacities from $2$B to $13$B
parameters: the Qwen-VL family (Qwen2-VL-2B/7B~\cite{wang2024qwen2vl}, Qwen2.5-VL-3B/7B~\cite{qwen2025qwen25technicalreport}), InternVL3-8B~\cite{zhu2025internvl3}, Idefics family (  
Idefics-9B-Instruct~\cite{laurencon2023obelics}, 
Idefics-2-8B~\cite{laurencon2024matters}, 
Idefics3-8B-Llama3~\cite{laurencon2024building}), Ristretto-3B~\cite{ristretto3b}, the Ovis1.5~\cite{lu2024ovis} family with two different language
backbones (Llama3-8B~\cite{grattafiori2024llama} and Gemma2-9B~\cite{team2024gemma}), the LLaVA-NeXT~\cite{liu2023improvedllava} family with three backbones (Vicuna-7B, Vicuna-13B, Mistral-7B), and Valley2~\cite{wu2025valley2}. 
Including several same-family, different-scale pairs (Qwen2-VL-2B vs.\ 7B, Qwen2.5-VL-3B vs.\ 7B, LLaVA-NeXT-Vicuna-7B vs.\ 13B) and several same-family, different-backbone pairs
(Ovis1.5, LLaVA-NeXT) allows us to separate the effect of capacity from the language backbone. 
All models use their publicly released instruction-tuned weights.
We evaluate these models in a zero-shot scenario for PAD and MAD.

We also consider 30 vision encoders from vision foundation models and MLLMs. We group them according to their main pretraining paradigm and objective:

\begin{itemize}

\item \textit{MLLM Vision Encoders.}
These encoders serve as the visual backbone of MLLMs and are subsequently aligned with a language model through a learned projector. This group includes the vision encoders of Qwen2-VL~\cite{wang2024qwen2vl}, Qwen2.5-VL~\cite{bai2025qwen25vl}, Qwen3-VL~\cite{bai2025qwen3vl}, LLaVA-NeXT~\cite{liu2023improvedllava}, LLaVA-OneVision~\cite{li2024llavaonevision}, DeepSeek-VL2~\cite{wu2024deepseekvl2}, and InternVL~\cite{chen2024internvl}. For example, InternVL connects InternViT to the language model through an MLP projector (\cref{fig:generic_pipeline}) and jointly optimizes the multimodal architecture for vision--language alignment and language modeling.

\item \textit{Vision--Language Contrastive Encoders.}
CLIP~\cite{radford2021learning} learns aligned image--text representations through contrastive pretraining on large-scale image--text pairs. SigLIP~\cite{siglip2023} replaces CLIP's softmax-based contrastive loss with an independent sigmoid loss, while EVA-CLIP~\cite{sun2023eva} combines masked-image pretraining with subsequent large-scale image--text contrastive training.

\item \textit{Self-Supervised Discriminative Encoders.}
DINOv2~\cite{oquab2023dinov2} combines self-distillation with iBOT-style masked prediction~\cite{zhou2021ibot}, while DINOv3~\cite{oquab2024dinov3} extends this paradigm with improved training at larger scale. MSN~\cite{assran2022masked} learns representations by matching masked views to prototype assignments obtained from unmasked views.

\item \textit{Masked Image Modeling Encoders.}
BEiT~\cite{bao2022beit} predicts discrete visual tokens for masked patches, whereas ViT-MAE~\cite{he2022mae} reconstructs their RGB values directly. Both learn visual representations from heavily masked inputs without requiring semantic class labels.

\end{itemize}

\subsection{Baselines}
We compare against recent PAD and MAD baselines. 
For PAD, we use DeepPixBiS~\cite{george2019deep}, FSFM-FAS~\cite{wang2025fsfm}, FLIP~\cite{srivatsan2023flip}, and FoundPAD~\cite{ozgur2025foundpad}. 
DeepPixBiS uses pixel-wise supervision on DenseNet-161 features, while FSFM-FAS pretrains a ViT-B/16 with face-specific self-supervision. 
FLIP adapts CLIP ViT-B/16 using visual and language-guided objectives (FLIP-V, FLIP-IT, and FLIP-MCL). 
FoundPAD adapts frozen CLIP ViT-B/16 and ViT-L/14 encoders using LoRA on the attention $Q,V$ projections.
For MAD, we use OrthoMAD~\cite{neto2022orthomad}, IDistill~\cite{caldeira2023idistill}, SelfMAD~\cite{ivanovska2025selfmad}, and MADation~\cite{caldeira2025madation}. 
OrthoMAD learns orthogonal identity representations with a ResNet-18, while IDistill extends this approach through identity distillation. 
SelfMAD trains an HRNet-W18 using synthetic morph-generic artefacts without morph supervision. 
MADation adapts CLIP ViT-B/16 and ViT-L/14 encoders with LoRA on the attention $Q,V$ projections.

\section{Evaluation Results}
In this section, we separately evaluate the performance of each method described in~\cref{sec:method}. 

\subsection{Zero-shot MLLMs}
We benchmark $16$ open-weight MLLMs mentioned in Section~\ref{subsec:experiments:models} on PAD and MAD, separately.

\subsubsection{Zero-shot MLLM for PAD}
Table~\ref{tab:zeroshot:pad_acer} reports ACER for different open-weight MLLMs for PAD on the MCIO benchmark. 
As the results in this table show, the best average ACER over the MCIO datasets is $31.1\%$
(LLaVA-NeXT-Mistral-7B), followed by $31.3\%$ (InternVL3-8B). 
Several models also achieve a chance level close to $50\%$ in our zero-shot evaluation.
Therefore, an off-the-shelf MLLM prompted for a liveness judgment is not a reliable PAD system.

According to this table, the performance is database-dependent, and the ordering of
models changes across databases. While InternVL3-8B reaches $9.2\%$ ACER on CASIA-FASD, it falls
to $47.3\%$ ACER on Replay-Attack and to $39.5\%$ ACER on
OULU-NPU. 
No model is best on all four datasets. 
This means that a zero-shot MLLM detector selected on one dataset gives no guarantee on another.

\begin{table}[tbp]
\centering
\caption{PAD ACER$\downarrow$ (\%)  of MLLMs in zero-shot evaluation.}
\label{tab:zeroshot:pad_acer}
\resizebox{\linewidth}{!}{%
\begin{tabular}{lccccc}
\toprule
Model & M & C & I & O & Avg. \\
\midrule
Idefics-9B-Instruct~\cite{laurencon2023obelics}             &           51.1 &         50.0 &           50.0 &          49.8 &           50.2 \\
LLaVA-NeXT-Vicuna-7B~\cite{liu2023improvedllava}            &           50.0 &          50.0 &           50.0 &          50.0 &           50.0 \\
Qwen2-VL-2B-Instruct~\cite{wang2024qwen2vl}                         &           50.7 &          49.2 &           45.3 &          49.8 &           48.8 \\
Qwen2.5-VL-7B-Instruct~\cite{qwen2025qwen25technicalreport} &           48.0 &          47.1 &           47.3 &          50.0 &           48.1 \\
Valley2~\cite{wu2025valley2}                                &           50.0 &          43.8 &           48.2 &          50.0 &           48.0 \\
LLaVA-NeXT-Vicuna-13B~\cite{liu2023improvedllava}           &           44.9 &          49.1 &           50.7 &          46.9 &           47.9 \\
Ovis1.5-Gemma2-9B~\cite{lu2024ovis}                         &           49.8 &          36.7 &           46.1 &          50.0 &           45.6 \\
Idefics3-8B-Llama3~\cite{laurencon2024building}             &           48.4 &          27.5 &           46.6 &          49.6 &           43.0 \\
Ristretto-3B~\cite{ristretto3b}                             &           39.9 &          47.6 &           48.8 &          34.7 &           42.7 \\
Ovis1.5-Llama3-8B~\cite{lu2024ovis}                         &           44.7 &          47.1 &           50.0 &          25.5 &           41.8 \\
Idefics2-8B~\cite{laurencon2024matters}                     &           49.2 &          14.0 &           47.1 &          49.6 &           40.0 \\
Qwen2.5-VL-3B-Instruct~\cite{qwen2025qwen25technicalreport} &           32.9 &          28.4 &           52.1 &          42.4 &           39.0 \\
Qwen2-VL-7B-Instruct~\cite{wang2024qwen2vl}                         &           39.4 &          38.9 &           43.0 &          32.3 &           38.4 \\
FaceLLM-8B~\cite{shahreza2025facellm}                       &           35.6 &          26.2 &           44.7 &          \textbf{22.2} &           32.2 \\
InternVL3-8B~\cite{zhu2025internvl3}                        &           \textbf{29.1} &           \textbf{9.2} &           47.3 &          39.5 &           31.3 \\
LLaVA-NeXT-Mistral-7B~\cite{liu2023improvedllava}           &           35.5 &          10.4 &           \textbf{38.0} &          40.6 &           \textbf{31.1} \\
\bottomrule
\end{tabular}}
\end{table}

\begin{table}[tbp]
\centering
\caption{MAD ACER$\downarrow$ (\%)  of MLLMs in zero-shot evaluation.}
\label{tab:zeroshot:mad_acer}
  \setlength{\tabcolsep}{4.3pt}
\resizebox{0.96\linewidth}{!}{%
\begin{tabular}{lccccc}
\toprule
Model & \scalebox{0.9}{FFHQ} & \scalebox{0.9}{FRGC} & \scalebox{0.9}{FRLL} & \scalebox{0.875}{FERET} & Avg. \\
\midrule
Qwen2-VL-2B-Instruct~\cite{wang2024qwen2vl}                         &          50.1 &          54.1 &          48.6 &          49.6 &          50.6 \\
LLaVA-NeXT-Vicuna-7B~\cite{liu2023improvedllava}            &          50.0 &          50.0 &          50.0 &          50.0 &          50.0 \\
Qwen2.5-VL-7B-Instruct~\cite{qwen2025qwen25technicalreport} &          49.9 &          49.1 &          50.0 &          50.0 &          49.7 \\
Qwen2.5-VL-3B-Instruct~\cite{qwen2025qwen25technicalreport} &          49.1 &          47.6 &          50.0 &          47.1 &          48.5 \\
LLaVA-NeXT-Vicuna-13B~\cite{liu2023improvedllava}  &          50.0 &          45.4 &          48.8 &          47.3 &          47.9 \\
Idefics-9B-Instruct~\cite{laurencon2023obelics}             &          53.2 &          41.8 &          40.4 &          50.6 &          46.5 \\
Idefics3-8B-Llama3~\cite{laurencon2024building}             &          38.2 &          46.1 &          49.9 &          49.7 &          46.0 \\
Idefics2-8B~\cite{laurencon2024matters}                     &          34.8 &          40.7 &          48.7 &          49.2 &          43.4 \\
Qwen2-VL-7B-Instruct~\cite{wang2024qwen2vl}                         &          27.7 &          43.9 &          49.8 &          46.2 &          41.9 \\
Valley2~\cite{wu2025valley2}                                &          40.0 &          50.0 &          44.5 &          40.9 &          43.9 \\
FaceLLM-8B~\cite{shahreza2025facellm}                       &          23.9 &          35.8 &          43.7 &          34.1 &          34.4 \\
Ristretto-3B~\cite{ristretto3b}                             &          42.7 &          36.2 &          35.4 &          \textbf{15.2} &          32.4 \\
LLaVA-NeXT-Mistral-7B~\cite{liu2023improvedllava}           &          18.4 &          29.1 &          41.1 &          38.9 &          31.9 \\
Ovis1.5-Gemma2-9B~\cite{lu2024ovis}                         &          26.4 &          \textbf{15.7} &          24.8 &          31.8 &          24.7 \\
Ovis1.5-Llama3-8B~\cite{lu2024ovis}                         &          \textbf{15.2} &          19.6 &          36.5 &          23.0 &          23.6 \\
InternVL3-8B~\cite{zhu2025internvl3}                        &          16.8 &          29.8 &          \textbf{11.3} &          19.5 &          \textbf{19.4} \\
\bottomrule
\end{tabular}}
\end{table}

\subsubsection{Zero-shot MLLM for MAD}
Table~\ref{tab:zeroshot:mad_acer} reports the corresponding MAD results in terms of ACER in zero-shot evaluation. 
As the results in this table show, MLLMs do not perform well for MAD in the zero-shot setting. 
Compared with Table~\ref{tab:zeroshot:pad_acer}, the average ACER is lower than for PAD.  InternVL3-8B is the best model for MAD with an ACER of $19.4\%$, whereas it had an average ACER of $31.3\%$ for PAD.
 Therefore, the morphing artefacts appear more accessible to generic visual priors than the capture artefacts that distinguish a replayed video from a live face. 
The next models in MAD are Ovis1.5-Llama3-8B and Ovis1.5-Gemma2-9B with ACER of $23.60\%$ and $24.67\%$, respectively.
Some models, such as Qwen2.5-VL-7B and LLaVA-NeXT-Vicuna-7B, however, are close to random chance.

\subsection{Language Model Logits Sampling}
As described in~\cref{sec:logit-sampling}, we can use the probability of the next token at the output of the language model to get a score for each image. 
We consider top-performing MLLMs and evaluate their performance with logit sampling. We also compare our results with the logit sampling approach proposed by 
Ivanovska and Štruc~\cite{ivanovska2026emergent} and zero-shot evaluation. 
Only \textit{ours} logit sampling approach involves training a separate classifier for each source dataset for every backbone, but the zero-shot and prior work\cite{ivanovska2026emergent} settings do not retrain per source.

\subsubsection{Logit Sampling for PAD}
\label{sec:experiments-logit-sampling-pad}
We evaluate top-performing MLLMs using our logit sampling approach. 
\Cref{tab:pad-comparison-dev} presents the PAD performance in terms of ACER for {InternVL3-8B}~\cite{zhu2025internvl3}, {FaceLLM-8B}~\cite{shahreza2025facellm}, {Ovis1.5-Llama3-8B}~\cite{lu2024ovis}, and {Qwen2-VL-7B-Instruct}~\cite{wang2024qwen2vl}. 

Our logit-sampling strategy improves performance with the InternVL3-8B and FaceLLM-8B backbones, achieving the lowest overall Avg ACER of $31.0\%$ and $31.3\%$ among the three settings, respectively. 
For InternVL3-8B, this improves on $32.9\%$ for Zero-shot and $35.9\%$ for prior work, and for FaceLLM, the corresponding values are $39.8\%$ and $36.1\%$. This advantage is consistent across all four train/dev protocols for both models. 
However, for Ovis1.5-Llama3-8B and Qwen2-VL-7B-Instruct, the prior work method performs better, achieving overall Avg ACER of $38.1\%$ and $35.9\%$, respectively, compared with $41.8\%$ and $37.4\%$ for ours.
Our approach also achieves the lowest intra-dataset (ID) ACER among all settings for all four backbones, ranging from $16.2\%$ to $24.9\%$, indicating a consistently strong fit to the training source.
In contrast, the cross-dataset (CD) ACER remains comparable to the other two settings, suggesting that the improvement in ``Overall Avg ACER'' is primarily driven by the gains in ID performance.

\begin{table}[tbp]
\centering
\caption{PAD ACER$\downarrow$ (\%) comparison of three decision-extraction settings.  Overall results summarize intra-dataset (\colorbox{yellow!30}{ID}), cross-dataset (\colorbox{green!30}{CD}), and average performance. 
\textbf{Bold}: best per column within each block, while our method is additionally highlighted in blue. 
Full results:~\cref{tab:pad-comparison-dev_supple}
}
\label{tab:pad-comparison-dev}
\resizebox{\linewidth}{!}{%
\begin{tabular}{ll cccc ccc}
\toprule
\multirow{2}{*}{Model} & \multirow{2}{*}{Method} &
\multicolumn{4}{c}{Train/Dev} &
\multicolumn{3}{c}{Overall} \\
\cmidrule(lr){3-6} \cmidrule(lr){7-9}
& & M  & C  & I  & O  &
\cellcolor{yellow!30}ID & \cellcolor{green!30}CD & Avg \\
\midrule

\multirow{3}{*}{
InternVL3-8B~\cite{zhu2025internvl3}}
& Zero-shot
& 33.9 & 31.9 & 31.9 & 33.9
& 31.3 & \textbf{33.4} & 32.9 \\

& \scalebox{0.9}{
I\&S~\cite{ivanovska2026emergent}}
& 35.2 & 35.2 & 34.3 & 39.1
& 30.1 & 37.9 & 35.9 \\
 \rowcolor[HTML]{D6ECFF} \cellcolor{white}
& Ours
& \textbf{30.2} & \textbf{30.8} & \textbf{31.9} & \textbf{31.2}
& \textbf{16.2} & 36.0 & \textbf{31.0} \\

\midrule

\multirow{3}{*}{Ovis1.5-Llama3-8B~\cite{lu2024ovis}}
& Zero-shot
& 41.9 & 41.9 & 49.8 & \textbf{41.9}
& 41.8 & 44.6 & 43.9 \\

& \scalebox{0.9}{I\&S~\cite{ivanovska2026emergent}}
& \textbf{37.9} & \textbf{35.5} & \textbf{36.0} & 42.8
& 35.1 & \textbf{39.0} & \textbf{38.1} \\
 \rowcolor[HTML]{D6ECFF} \cellcolor{white}
& Ours
& 41.4 & 39.9 & 42.6 & 43.3
& \textbf{24.9} & 47.4 & 41.8 \\

\midrule

\multirow{3}{*}{\scalebox{0.9}{Qwen2-VL-7B-Instruct~\cite{wang2024qwen2vl}}}
& Zero-shot
& \textbf{38.4} & 48.1 & 38.4 & 38.4
& 40.5 & 41.0 & 40.8 \\

& \scalebox{0.9}{I\&S~\cite{ivanovska2026emergent}}
& 34.9 & 32.8 & \textbf{31.3} & 44.4
& 30.3 & \textbf{37.7} & \textbf{35.9} \\
 \rowcolor[HTML]{D6ECFF} \cellcolor{white}
& Ours
& \textbf{30.1} & \textbf{28.2} & 42.9 & 48.4
& \textbf{20.0} & 43.2 & 37.4 \\

\midrule

\multirow{3}{*}{FaceLLM-8B~\cite{shahreza2025facellm}}
& Zero-shot
& 37.0 & 42.7 & 42.7 & 37.0
& 32.2 & 42.4 & 39.8 \\

& \scalebox{0.9}{I\&S~\cite{ivanovska2026emergent}}
& 34.0 & 36.0 & 35.3 & 39.0
& 29.6 & \textbf{7.8 }& 36.1 \\
 \rowcolor[HTML]{D6ECFF} \cellcolor{white}
& Ours
& \textbf{27.4} & \textbf{29.4} & \textbf{34.7} & \textbf{33.7}
& \textbf{20.9} & 34.8 & \textbf{31.3} \\

\bottomrule
\end{tabular}%
}
\vspace{-5pt}
\end{table}

\begin{table}[tbp]
\centering
\caption{
MAD ACER$\downarrow$ (\%) comparison of three decision-extraction settings.  Overall results summarize intra-dataset (\colorbox{yellow!30}{ID}), cross-dataset (\colorbox{green!30}{CD}), and average performance. \textbf{Bold}: best per column within each block, while our method is additionally highlighted in blue.Full results:~\cref{tab:mad-comparison-dev_supple}
}
\label{tab:mad-comparison-dev}
\resizebox{\linewidth}{!}{%
\begin{tabular}{ll ccc cc}
\toprule
\multirow{2}{*}{Model} & \multirow{2}{*}{Method} &
\multicolumn{2}{c}{Train/Dev} &
\multicolumn{3}{c}{Overall} \\
\cmidrule(lr){3-4} \cmidrule(lr){5-7}
& & FFHQ & FRGC &
\cellcolor{yellow!30}ID & \cellcolor{green!30}CD & Avg \\
\midrule

\multirow{3}{*}{InternVL3-8B~\cite{zhu2025internvl3}}
& Zero-shot
& 19.4 & 19.4
& 14.1 & 21.1 & 19.4 \\

& \scalebox{0.9}{I\&S~\cite{ivanovska2026emergent}}
& 13.0 & 13.1
& 14.4 & \textbf{12.7} & 13.1 \\
 \rowcolor[HTML]{D6ECFF} \cellcolor{white}
& Ours
& \textbf{12.4} & \textbf{9.7}
& \textbf{3.5} & 13.5 & \textbf{11.0} \\

\midrule

\multirow{3}{*}{Ovis1.5-Llama3-8B~\cite{lu2024ovis}}
& Zero-shot
& 23.6 & 23.6
& 25.9 & 22.8 & 23.6 \\

& \scalebox{0.9}{I\&S~\cite{ivanovska2026emergent}}
& 17.4 & 23.0
& 15.5 & 21.7 & 20.2 \\
 \rowcolor[HTML]{D6ECFF} \cellcolor{white}
& Ours
& \textbf{10.1} & \textbf{21.0}
& \textbf{1.7} & \textbf{20.2} & \textbf{15.6} \\

\midrule

\multirow{3}{*}{Qwen2-VL-7B-Instruct~\cite{wang2024qwen2vl}}
& Zero-shot
& 41.9 & 41.9
& 38.8 & 43.0 & 41.9 \\

& \scalebox{0.9}{I\&S~\cite{ivanovska2026emergent}}
& 22.5 & \textbf{15.2}
& 14.2 & \textbf{20.4} & 18.9 \\
 \rowcolor[HTML]{D6ECFF} \cellcolor{white}
& Ours
& \textbf{10.0} & 22.7
& \textbf{0.3} & 21.7 & \textbf{16.4} \\

\midrule

\multirow{3}{*}{FaceLLM-8B~\cite{shahreza2025facellm}}
& Zero-shot
& 34.4 & 34.4
& 33.8 & 34.6 & 34.4 \\

& \scalebox{0.9}{I\&S~\cite{ivanovska2026emergent}}
& 16.8 & 17.4
& 17.9 & 16.9 & 17.1 \\
 \rowcolor[HTML]{D6ECFF} \cellcolor{white}
& Ours
& \textbf{8.6} & \textbf{7.5}
& \textbf{3.6} & \textbf{9.5} & \textbf{8.0} \\

\bottomrule
\end{tabular}%
}
\vspace{-5pt}
\end{table}

\subsubsection{Logit Sampling for MAD}

\Cref{tab:mad-comparison-dev} reports MAD ACER (\%) for the same four MLLMs and three settings as \cref{tab:pad-comparison-dev}: {Zero-shot}, the prior work of \cite{ivanovska2026emergent}, and  our logit-sampling approach (\cref{sec:logit-sampling}).
Surprisingly, for our logit-sampling strategy, MAD achieves the lowest Overall Avg ACER for all backbones, with $11.0\%$ for InternVL3-8B, $8.0\%$ for FaceLLM, $15.6\%$ for Ovis1.5-Llama3-8B, and $16.4\%$ for Qwen2-VL-7B-Instruct. The gain is largest for FaceLLM, where our method reduces the Overall Avg ACER from $34.4\%$ for Zero-shot to $8.0\%$, more than a fourfold reduction. Our logit-sampling method achieves the lowest intra-dataset (ID) ACER for all backbones. However, unlike PAD, the advantage extends beyond intra-dataset performance: cross-dataset (CD) ACER also improves over Zero-shot for every backbone, decreasing from $21.1-43.0\%$ to $9.5-21.7\%$ and is competitive with the prior method. 

These results indicate that our logit-sampling approach enhances cross-dataset generalization, improving performance beyond the training-source fit. 
Together with the results in \cref{tab:pad-comparison-dev}, these findings indicate that logit sampling effectiveness varies with the task and backbone. 
While logit sampling consistently improves MAD performance across all evaluated backbones, the improvements are more significant for MAD than for PAD, particularly in cross-dataset generalization.

\subsection{LoRA Adaptation of MLLMs}
As described in~\cref{sec:fine-tuning-mllms}, we generate QA datasets for PAD and MAD, and train separate MLLMs for each task. 
We use pretrained InternVL3-8B as the base model, which is among the top-performing MLLMs in our zero-shot evaluation (\cref{tab:zeroshot:pad_acer,tab:zeroshot:mad_acer}), with LoRA parameters $\alpha = 32$ and $r = 16$ from~\cref{eq:lora}.

\subsubsection{LoRA Adapting MLLM for PAD}
We construct a task-specific QA dataset for PAD and use it to fine-tune a pretrained MLLM. Specifically, we fine-tune InternVL3-8B on a combination of the PAD QA dataset and the FairFaceGPT dataset. The trained model, called PADLLM, is now specialized for PAD. 
We evaluate PADLLM's performance by prompting the model to output a score, as in our zero-shot evaluation described in Section~\ref{sec:zeroshot}.
Table~\ref{tab:fine-tuning:pad_acer} compares the performance of PADLLM with previous MLLMs from the literature.
As the results in this table show, PADLLM outperforms all MLLMs across MCIO datasets. 
The average ACER improves from $31.1\%$ in LLaVA-NeXT-Mistral-7B to $13.6\%$ in PADLLM.
In addition, we observe performance improvements on all datasets: MSU-MFSD (from $29.1\%$ to $16.2\%$), CASIA-FASD (from $9.2\%$ to $8.3\%$), Replay-Attack (from $38.0\%$ to $26.6\%$), and OULU-NPU (from $22.2\%$ to $3.2\%$).
Compared with the base model (InternVL-8B), PADLLM performs better on all datasets, demonstrating the effectiveness of our fine-tuning.

\subsubsection{LoRA Adapting MLLM for MAD}
Similarly, we generate a QA dataset for MAD and fine-tune a pretrained MLLM (InternVL-8B) with our MAD QA dataset along with the FairFaceGPT dataset. 
The resulting model, referred to as MADLLM, is specifically fine-tuned for morphing attack detection (MAD). We evaluate MADLLM's performance by prompting the model to output a score, as in our zero-shot evaluation.
Table~\ref{tab:fine-tuning:mad_acer} reports the performance of MADLLM and compares it with previous MLLMs from the literature.
As the results in this table show, MADLLM improves the performance of MLLMs on all datasets: FFHQ (from $15.2\%$ to $6.1\%$), FRGC (from $15.7\%$ to $0.7\%$), FRLL (from $11.3\%$ to $1.2\%$), and FERET (from $15.2\%$ to $3.7\%$) morph images. 
Similar to PADLLM, MADLLM outperforms its base model (InternVL3-8B) on all datasets, demonstrating the effectiveness of our fine-tuning.

\begin{table}[tbp]
\centering
\caption{PAD ACER$\downarrow$ (\%)  Comparison of the performance of fine-tuned and off-the-shelf MLLMs under zero-shot evaluation.}
\label{tab:fine-tuning:pad_acer}
\resizebox{0.875\linewidth}{!}{%
\begin{tabular}{lccccc}
\toprule
Model & M & C & I & O & Avg. \\
\midrule
Idefics3-8B-Llama3~\cite{laurencon2024building}             &           48.4 &          27.5 &           46.6 &          49.6 &           43.0 \\
Ristretto-3B~\cite{ristretto3b}                             &           39.9 &          47.6 &           48.8 &          34.7 &           42.7 \\
Ovis1.5-Llama3-8B~\cite{lu2024ovis}                         &           44.7 &          47.1 &           50.0 &          25.5 &           41.8 \\
Idefics2-8B~\cite{laurencon2024matters}                     &           49.2 &          14.0 &           47.1 &          49.6 &           40.0 \\
Qwen2.5-VL-3B-Instruct~\cite{qwen2025qwen25technicalreport} &           32.9 &          28.4 &           52.1 &          42.4 &           39.0 \\
Qwen2-VL-7B-Instruct~\cite{wang2024qwen2vl}                         &           39.4 &          38.9 &           43.0 &          32.3 &           38.4 \\
FaceLLM-8B~\cite{shahreza2025facellm}                       &           35.6 &          26.2 &           44.7 &          {22.2} &           32.2 \\
InternVL3-8B~\cite{zhu2025internvl3}                        &           {29.1} &           {9.2} &           47.3 &          39.5 &           31.3 \\
LLaVA-NeXT-Mistral-7B~\cite{liu2023improvedllava}           &           35.5 &          10.4 &           {38.0} &          40.6 &           {31.1} \\
 \rowcolor[HTML]{D6ECFF}  \textbf{PADLLM-8B [ours]}              &  \textbf{16.2} &  \textbf{8.3} &  \textbf{26.6} &  \textbf{3.2} &  \textbf{13.6} \\
\bottomrule
\end{tabular}}
\vspace{-5pt}
\end{table}

\begin{table}[tbp]
\centering
\caption{MAD ACER$\downarrow$ (\%) Comparison of the performance of fine-tuning with off-the-shelf MLLMs in  zero-shot evaluation.}
\label{tab:fine-tuning:mad_acer}
\resizebox{0.975\linewidth}{!}{%
\begin{tabular}{lccccc}
\toprule
Model & FFHQ & FRGC & FRLL & FERET & Avg. \\
\midrule
Idefics2-8B~\cite{laurencon2024matters}                     &          34.8 &          40.7 &          48.7 &          49.2 &          43.4 \\
Qwen2-VL-7B-Instruct~\cite{wang2024qwen2vl}                         &          27.7 &          43.9 &          49.8 &          46.2 &          41.9 \\
Valley2~\cite{wu2025valley2}                                &          40.0 &          50.0 &          44.5 &          40.9 &          43.9 \\
FaceLLM-8B~\cite{shahreza2025facellm}                       &          23.9 &          35.8 &          43.7 &          34.1 &          34.4 \\
Ristretto-3B~\cite{ristretto3b}                             &          42.7 &          36.2 &          35.4 &          15.2 &          32.4 \\
LLaVA-NeXT-Mistral-7B~\cite{liu2023improvedllava}           &          18.4 &          29.1 &          41.1 &          38.9 &          31.9 \\
Ovis1.5-Gemma2-9B~\cite{lu2024ovis}                         &          26.4 &          {15.7} &          24.8 &          31.8 &          24.7 \\
Ovis1.5-Llama3-8B~\cite{lu2024ovis}                         &          {15.2} &          19.6 &          36.5 &          23.0 &          23.6 \\
InternVL3-8B~\cite{zhu2025internvl3}                        &          16.8 &          29.8 &          {11.3} &          {19.5} &          {19.4} \\
 \rowcolor[HTML]{D6ECFF}  \textbf{MADLLM-8B [ours]}              &  \textbf{6.1} &  \textbf{0.7} &  \textbf{1.2} &  \textbf{3.7} &  \textbf{2.9} \\
\bottomrule
\end{tabular}}
\vspace{-5pt}
\end{table}

\subsection{Linear Probe a Vision-encoder as Feature Extractor}
We use the 30 vision encoders of FMs (compare~\cref{subsec:experiments:models}) as feature extractors and train classifiers on vision-encoder features for PAD and MAD, respectively.

\subsubsection{Linear Probe a Vision-encoder for PAD}

\Cref{tab:intra-dataset-acer} summarizes the performance of vision encoders for presentation attack detection (PAD) on MCIO datasets, providing both intra-dataset (yellow) and cross-dataset (green) evaluations.
A separate comparison of dedicated detectors appears in \cref{sec:discussion}.
Larger models consistently outperform smaller ones.
InternViT-6B achieves the lowest average intra-dataset ACER at 1.6\%.
DINOv3-H+ achieves the second-lowest intra-dataset ACER at 5.7\%, followed by CLIP ViT-B/16 at 5.9\%.

However, cross-dataset error remains high for nearly all frozen probes.
When trained on MSU-MFSD, InternViT-6B achieves 4.4\% ACER on its own corpus but 14.7-45.3\% on the three held-out datasets. 
Increasing model scale improves performance but does not eliminate the cross-dataset gap. 
InternViT-6B outperforms InternViT-300M (average ACER from 41.8\% to 24.4\%).
Consequently, the intra-dataset and cross-dataset rankings diverge. InternViT-6B achieves the lowest intra-dataset (1.6\%) and lowest average (24.4\%) error, whereas CLIP ViT-B/16 attains the lowest cross-dataset error (31.5\%).

Within the CLIP family, the two base models demonstrate the best transfer performance: B/16 achieves 31.5\%, and B/32 achieves 31.6\% mean cross-dataset ACER, both outperforming ViT-L/14 (34.0\%) and ViT-H/14 (38.7\%).
For B/32, this improved cross-dataset robustness is accompanied by a weaker intra-dataset fit (12.6\% compared to 5.9\%), indicating that its coarser $32{\times}32$ patch tokenization may reduce separability while enhancing robustness to MCIO shift under a linear readout.
Frozen CLIP backbones have also previously ranked among the strongest cross-dataset extractors in VFM-FAS benchmarks~\cite{srivatsan2023flip,liu2024cfpl,feng2026benchmarking}.
Low intra-dataset ACER does not reliably predict cross-dataset PAD performance, even for models with billions of parameters.

\input{content/tables/pad_frozen_matrix}

\subsubsection{Linear Probe a Vision-encoder for MAD}

\Cref{tab:mad-ffhq-frgc-acer} presents test ACER (\%) for both intra-dataset and cross-dataset evaluations on the MAD datasets. Each encoder is trained on either FFHQ or FRGC and evaluated across all four corpora (FFHQ, FRGC, FRLL, and FERET). Each Train/Dev column reports the mean over the four target datasets for a given training source. The Overall columns distinguish between the two intra-dataset pairs, FFHQ$\rightarrow$FFHQ and FRGC$\rightarrow$FRGC, and the six cross-dataset pairs. FRLL and FERET are not used for training and therefore appear only as cross-dataset targets. Per-target results are provided in \cref{tab:app:mad-ffhq-frgc-acer}.

Among the frozen foundation models, EVA-CLIP-8B~\cite{sun2023eva} demonstrates the strongest overall performance, achieving $5.0\%$ ACER across the eight train/test pairs and $6.4\%$ on cross-dataset evaluations.
Contrastively pretrained encoders are generally competitive. For example, CLIP ViT-H/14~\cite{radford2021learning} ($12.1\%$) and SigLIP-large~\cite{siglip2023} ($15.1\%$) both outperform all self-supervised encoders.
However, model capacity alone does not dictate performance outcomes.
InternViT-6B~\cite{chen2024internvl2} achieves a $13.9\%$ average ACER, lagging behind several substantially smaller encoders. ViT-MSN~\cite{assran2023msn} continues to be among the weakest frozen representations for MAD.

\input{content/tables/mad_frozen_matrix}

Similar to PAD, low intra-dataset error does not reliably predict cross-dataset performance.
For example, InternViT-6B achieves $0.1\%$ intra-dataset ACER but $18.5\%$ across the six cross-dataset pairs. When trained on FFHQ, it transfers almost perfectly to FRGC ($0.0\%$) and FRLL ($0.1\%$), but its error increases to $31.1\%$ on FERET. In contrast, the dedicated MADation ViT-L/14 baseline exhibits much smaller shifts on the same targets ($3.5\%$, $4.2\%$, and $0.4\%$; \cref{tab:app:mad-ffhq-frgc-acer}).
Therefore, a frozen foundation model may outperform a specialist detector on several source-to-target pairs, yet still fail significantly under specific corpus shifts.

Transfer performance is also highly target-dependent. DINOv3~\cite{oquab2024dinov3} and BEiT~\cite{bao2022beit} exhibit substantially different error rates across FRGC, FRLL, and FERET (\cref{tab:app:mad-ffhq-frgc-acer}).

Overall, the MAD results replicate the central pattern observed for PAD: achieving strong source-domain separability is considerably easier than attaining robust cross-dataset transfer.
Large-scale pretraining can yield highly discriminative frozen representations. 
However, neither increased model scale nor low intra-dataset ACER alone ensures robustness to previously unseen morphing corpora.

\subsection{Adaptation of Vision-Encoders as Feature Extractor}
\label{sec:loravisionencoder}

The vision-encoder of FMs can also be adapted for PAD and MAD tasks.
We evaluate the same vision encoders for each task using consistent Low-Rank Adaptation (LoRA) parameters, specifically $r{=}8$ and $\alpha{=}16$.

\subsubsection{Adaptation of Vision-Encoders for PAD}

\Cref{tab:lora-intra-dataset-acer} presents the multi-class intra-dataset evaluation (MCIO) following LoRA adaptation.
Intra-dataset PAD performance approaches saturation. The best model-zoo encoder, CLIP ViT-B/32, achieves an average intra-dataset (ID) average classification error rate (ACER) of only $0.3\%$, with many encoders remaining below $1\%$.
Consequently, model ranking is determined primarily by cross-dataset transfer performance rather than by source-domain separability.

The strongest overall results are achieved by LLaVA-NeXT-7B-ViT and CLIP ViT-B/16, both attaining a mean cross-dataset ACER of $19.8\%$, followed by DINOv2-registers-giant at $20.4\%$.

However, a substantial gap persists between intra-dataset (ID) and cross-dataset (CD) performance.
For example, LLaVA-NeXT-7B-ViT achieves $0.4\%$ ID ACER but $19.8\%$ CD ACER. Transfer performance also depends strongly on the source dataset.
adapted on Replay-Attack, the same encoder averages $3.5\%$, against $21.4\%$ when adapted on Oulu-NPU.
Model scale alone does not account for these differences, as comparatively small CLIP encoders outperform several vision towers with multi-billion parameters.
Thus, LoRA adaptation largely addresses intra-dataset PAD under this protocol, while robustness to acquisition-domain shift remains the principal limitation.

\input{content/tables/pad_lora_matrix}

\input{content/tables/mad_lora_matrix}

\subsubsection{Adaptation of Vision-Encoders for MAD}

\Cref{tab:lora-mad-intra-dataset-acer} demonstrates a pronounced adaptation effect for MAD.
Most encoders achieve near-zero intra-dataset error, with model differences primarily determined by cross-dataset generalization. CLIP ViT-L/14 attains the strongest overall performance, reporting a $3.7\%$ average ACER and $4.9\%$ mean cross-dataset ACER. This is followed by CLIP ViT-H/14 ($7.1\%$ CD) and EVA-CLIP-8B ($8.4\%$ CD).

Adaptation is most beneficial when training on FFHQ, as several LoRA-adapted encoders transfer to FRGC, FRLL, and FERET with error rates of only a few percent.

Transfer performance remains asymmetric. Models trained on FFHQ generalize substantially better than those trained on FRGC. For example, InternViT-6B averages $0.5\%$ from FFHQ compared to $21.8\%$ from FRGC, and CLIP ViT-L/14 achieves $1.1\%$ versus $6.3\%$ (\cref{tab:app:lora-mad-intra-dataset-acer}).
This asymmetry is observed across multiple encoder families, indicating that comprehensive source-domain coverage remains important even after adaptation.

Overall, LoRA demonstrates greater effectiveness for MAD than for PAD. The best mean cross-dataset ACER decreases to $4.9\%$ for MAD, compared with $19.8\%$ for PAD. These results suggest that morphing artefacts are more readily captured by lightweight adaptation of generic pretrained representations, whereas presentation attacks remain more strongly associated with dataset-specific acquisition conditions.

\FloatBarrier

%% file: content/tables/pad_frozen_matrix.tex
\begin{table}[tbp]
\centering
\caption{
{LP for PAD.} Test ACER$\downarrow$ (\%) of frozen vision encoders with linear probes.
Overall results summarize intra-dataset (\colorbox{yellow!30}{ID}), cross-dataset (\colorbox{green!30}{CD}), and average performance. \textbf{Bold}: best per column within each block.
Full results:~\cref{tab:app:intra-dataset-acer}.
}
\label{tab:intra-dataset-acer}
\resizebox{\linewidth}{!}{%
\begin{tabular}{lccccrrr}
\toprule
\multirow{2}{*}{Model}
& \multicolumn{4}{c}{Train/Dev}
& \multicolumn{3}{c}{Overall} \\
\cmidrule(lr){2-5}
\cmidrule(lr){6-8}
& M & C & I & O
& \cellcolor{yellow!30}{ID}
& \cellcolor{green!30}{CD}
& Avg \\
\midrule
ViT-MSN-base~\cite{assran2023msn}                  & 56.0 & 57.0 & 54.3          & 47.2          & 46.5                      & 56.0                     & 53.6 \\
ViT-MSN-large~\cite{assran2023msn}                 & 47.0 & 47.1 & 46.5          & 46.5          & 45.2                      & 47.3                     & 46.8 \\
InternViT-300M~\cite{chen2024internvl2}            & 40.2 & 43.2 & 41.8          & 41.8          & 23.9                      & 47.7                     & 41.8 \\
ViT-MAE-base~\cite{he2022mae}                      & 35.6 & 43.8 & 42.7          & 44.7          & 18.5                      & 49.4                     & 41.7 \\
SigLIP-base~\cite{siglip2023}                      & 36.7 & 39.8 & 37.5          & 43.8          & 22.5                      & 45.1                     & 39.5 \\
Qwen3-VL-2B-ViT~\cite{bai2025qwen3vl}              & 43.4 & 32.6 & 42.4          & 39.5          & 23.2                      & 44.9                     & 39.5 \\
DeepSeek-VL2-7B-ViT~\cite{wu2024deepseekvl2}       & 32.9 & 35.0 & 40.6          & 41.8          & 17.2                      & 44.4                     & 37.6 \\
DeepSeek-VL2-small-ViT~\cite{wu2024deepseekvl2}    & 30.4 & 34.6 & 40.3          & 41.0          & 14.7                      & 43.9                     & 36.6 \\
Qwen2.5-VL-3B-ViT~\cite{bai2025qwen25vl}           & 34.7 & 34.8 & 42.0          & 34.4          & 16.2                      & 43.2                     & 36.5 \\
SigLIP-large~\cite{siglip2023}                     & 34.7 & 38.6 & 34.7          & 37.4          & 19.3                      & 42.0                     & 36.3 \\
ViT-MAE-huge~\cite{he2022mae}                      & 29.4 & 40.0 & 41.1          & 34.4          & 16.7                      & 42.7                     & 36.2 \\
DINOv2-base~\cite{oquab2023dinov2}                 & 35.3 & 34.6 & 30.7          & 41.4          & 13.2                      & 42.9                     & 35.5 \\
Qwen3-VL-32B-ViT~\cite{bai2025qwen3vl}             & 32.6 & 33.6 & 38.5          & 36.2          & 16.1                      & 41.6                     & 35.2 \\
BEiT-base~\cite{bao2022beit}                       & 35.6 & 35.7 & 34.0          & 34.9          & 9.2                       & 43.7                     & 35.0 \\
Qwen2.5-VL-7B-ViT~\cite{bai2025qwen25vl}           & 30.1 & 33.6 & 37.9          & 37.6          & 17.2                      & 40.7                     & 34.8 \\
DINOv2-registers-base~\cite{darcet2024registers}   & 30.5 & 35.3 & 31.5          & 40.1          & 14.6                      & 40.9                     & 34.4 \\
LLaVA-NeXT-7B-ViT~\cite{liu2023improvedllava}          & 35.6 & 30.8 & 26.8          & 40.7          & 17.3                      & 38.9                     & 33.5 \\
DINOv3-base~\cite{oquab2024dinov3}                 & 32.2 & 31.4 & 38.4          & 30.8          & 9.4                       & 41.1                     & 33.2 \\
LLaVA-OneVision-7B-ViT~\cite{li2024llavaonevision} & 28.6 & 30.8 & 32.5          & 40.0          & 13.9                      & 39.3                     & 33.0 \\
DINOv3-huge+~\cite{oquab2024dinov3}                & 31.4 & 29.9 & 35.3          & 34.4          & 5.7                       & 41.8                     & 32.7 \\
DINOv2-registers-giant~\cite{darcet2024registers}  & 31.4 & 26.7 & 33.9          & 35.7          & 8.6                       & 39.7                     & 31.9 \\
DINOv2-giant~\cite{oquab2023dinov2}                & 30.7 & 28.8 & 30.5          & 37.5          & 8.4                       & 39.7                     & 31.9 \\
CLIP ViT-H/14~\cite{radford2021learning}           & 32.2 & 27.2 & 31.7          & 35.4          & 10.5                      & 38.7                     & 31.6 \\
EVA-CLIP-8B~\cite{sun2023eva}                      & 31.4 & 24.5 & 31.3          & 34.3          & 10.2                      & 37.1                     & 30.4 \\
BEiT-large~\cite{bao2022beit}                      & 24.4 & 27.6 & 31.6          & 33.6          & 9.1                       & 36.0                     & 29.3 \\
CLIP ViT-L/14~\cite{radford2021learning}           & 35.5 & 19.5 & 30.1          & 27.6          & 10.6                      & 34.0                     & 28.2 \\
Qwen2-VL-72B-ViT~\cite{wang2024qwen2vl}            & 20.8 & 26.3 & 33.7          & 31.5          & 8.5                       & 34.6                     & 28.1 \\
CLIP ViT-B/32~\cite{radford2021learning}           & 35.0 & 21.3 & 27.3          & \bestcell{23.7} & 12.6                      & 31.6                     & 26.8 \\
CLIP ViT-B/16~\cite{radford2021learning}           & 29.7 & 20.2 & \bestcell{19.3} & 31.0           & 5.9    & \bestcell{31.5} & 25.1          \\
InternViT-6B~\cite{chen2024internvl2}              & \bestcell{20.3} & \bestcell{19.3} & 26.0          & 32.0 & \bestcell{1.6} & 32.0          & \bestcell{24.4} \\ 
\bottomrule
\end{tabular}
}
\end{table}

%% file: content/tables/mad_frozen_matrix.tex
\begin{table}[tbp]
\centering
    \caption{
    {LP for MAD.} Test ACER$\downarrow$ (\%) of frozen vision encoders with linear probes. 
    Overall results summarize intra-dataset (\colorbox{yellow!30}{ID}), cross-dataset (\colorbox{green!30}{CD}), and average performance. \textbf{Bold}: best per column within each block. 
    Full results:~\cref{tab:app:mad-ffhq-frgc-acer}.
    }
\label{tab:mad-ffhq-frgc-acer}
\resizebox{0.85\linewidth}{!}{%
\begin{tabular}{lccrrr}
\toprule
\multirow{2}{*}{Model}                                       & \multicolumn{2}{c}{Train/Dev} & \multicolumn{3}{c}{Overall}                                         \\ \cmidrule(lr){2-3}
\cmidrule(lr){4-6}
                                                             & FFHQ          & FRGC          & \cellcolor{yellow!30}{ID} & \cellcolor{green!30}{CD} & Avg          \\ \midrule
ViT-MSN-large~\cite{assran2023msn}                           & 40.7          & 35.7          & 24.0                      & 42.9                     & 38.2         \\
ViT-MSN-base~\cite{assran2023msn}                            & 44.3          & 20.5          & 19.2                      & 36.8                     & 32.4         \\
InternViT-300M~\cite{chen2024internvl2}                      & 23.3          & 33.3          & 5.5                       & 35.9                     & 28.3         \\
DINOv3-H+~\cite{oquab2024dinov3}                             & 31.6          & 22.5          & 4.5                       & 34.5                     & 27.0         \\
ViT-MAE-huge~\cite{he2022mae}                                & 26.8          & 23.3          & 8.4                       & 30.6                     & 25.1         \\
DINOv2-R-giant~\cite{darcet2024registers}                    & 30.1          & 19.1          & 4.9                       & 31.2                     & 24.6         \\
ViT-MAE-base~\cite{he2022mae}                                & 26.0          & 21.9          & 10.4                      & 28.5                     & 24.0         \\
DINOv2-base~\cite{oquab2023dinov2}                           & 26.5          & 21.2          & 4.6                       & 30.2                     & 23.8         \\
\scalebox{0.85}{LLaVA-OneVision-7B-ViT~\cite{li2024llavaonevision}} & 25.3         & 20.2          & 0.9          & 30.1          & 22.8          \\
LLaVA-NeXT-7B-ViT~\cite{liu2023improvedllava}                    & 26.7          & 18.3          & 0.9                       & 29.7                     & 22.5         \\
SigLIP-base~\cite{siglip2023}                                & 22.3          & 22.6          & 6.3                       & 27.9                     & 22.5         \\
DINOv2-R-base~\cite{darcet2024registers}                     & 23.9          & 21.0          & 7.1                       & 27.5                     & 22.4         \\
DINOv3-large~\cite{oquab2024dinov3}                          & 28.4          & 16.3          & 4.7                       & 28.3                     & 22.4         \\
BEiT-large~\cite{bao2022beit}                                & 16.6          & 27.3          & 4.7                       & 27.7                     & 21.9         \\
BEiT-base~\cite{bao2022beit}                                 & 18.9          & 23.8          & 6.5                       & 26.3                     & 21.3         \\
DINOv2-giant~\cite{oquab2023dinov2}                          & 23.7          & 17.1          & 4.7                       & 25.6                     & 20.4         \\
\scalebox{0.85}{DeepSeek-VL2-small-ViT~\cite{wu2024deepseekvl2}}    & 19.9         & 20.0          & 1.4          & 26.1          & 20.0          \\
Qwen3-VL-2B-ViT~\cite{bai2025qwen3vl}                        & 21.2          & 18.1          & 5.8                       & 24.3                     & 19.6         \\
Qwen2.5-VL-3B-ViT~\cite{bai2025qwen25vl}                     & 13.0          & 26.1          & 4.2                       & 24.7                     & 19.5         \\
DINOv3-base~\cite{oquab2024dinov3}                           & 22.0          & 15.4          & 5.6                       & 23.1                     & 18.7         \\
Qwen2-VL-72B-ViT~\cite{wang2024qwen2vl}                      & 12.2          & 25.1          & 2.4                       & 24.0                     & 18.6         \\
Qwen3-VL-32B-ViT~\cite{bai2025qwen3vl}                       & 14.5          & 21.1          & 3.4                       & 22.6                     & 17.8         \\
Qwen2.5-VL-7B-ViT~\cite{bai2025qwen25vl}                     & 13.0          & 21.5          & 3.5                       & 21.9                     & 17.3         \\
CLIP ViT-B/32~\cite{radford2021learning}                     & 13.1          & 19.0          & 2.7                       & 20.5                     & 16.1         \\
SigLIP-large~\cite{siglip2023}                               & 15.9          & 14.3          & 3.7                       & 18.9                     & 15.1         \\
\scalebox{0.9}{DeepSeek-VL2-7B-ViT~\cite{wu2024deepseekvl2}} & 15.5          & 14.5          & 0.6                       & 19.8                     & 15.0         \\
InternViT-6B~\cite{chen2024internvl2}                        & 7.9           & 19.9          & \bestcell{0.1}              & 18.5                     & 13.9         \\
CLIP ViT-L/14~\cite{radford2021learning}                     & 10.2          & 16.8          & 1.4                       & 17.5                     & 13.5         \\
CLIP ViT-H/14~\cite{radford2021learning}                     & 9.2           & 14.9          & 0.6                       & 15.9                     & 12.1         \\
EVA-CLIP-8B~\cite{sun2023eva}                                & \bestcell{1.9}  & \bestcell{8.1}  & 0.6                   & \bestcell{6.4}             & \bestcell{5.0} \\ 
\bottomrule
\end{tabular}
}
\end{table}

%% file: content/tables/pad_lora_matrix.tex
\begin{table}[tbp]
\caption{
{LoRA for PAD.} 
Test ACER$\downarrow$ (\%) after LoRA adaptation of the vision encoder attention.
Overall results summarize intra-dataset (\colorbox{yellow!30}{ID}), cross-dataset (\colorbox{green!30}{CD}), and average performance. 
\textbf{Bold}: best per column within each block. 
Full results:~\cref{tab:app:lora-intra-dataset-acer}.
}
\label{tab:lora-intra-dataset-acer}
\resizebox{\linewidth}{!}{%
\begin{tabular}{lccccrrr}
\toprule
\multirow{2}{*}{Model}
& \multicolumn{4}{c}{Train/Dev}
& \multicolumn{3}{c}{Overall} \\
\cmidrule(lr){2-5}
\cmidrule(lr){6-8}
& M & C & I & O
& \cellcolor{yellow!30}{ID}
& \cellcolor{green!30}{CD}
& Avg \\
\midrule
InternViT-300M~\cite{chen2024internvl2}            & 32.4          & 29.3          & 35.9          & 34.8 & 4.6                       & 42.6                     & 33.1 \\
ViT-MSN-large~\cite{assran2023msn}                 & 26.3          & 24.2          & 38.4          & 37.3 & 5.2                       & 40.3                     & 31.6 \\
ViT-MAE-base~\cite{he2022mae}                      & 18.1          & 29.8          & 39.7          & 36.3 & 3.8                       & 40.0                     & 31.0 \\
BEiT-large~\cite{bao2022beit}                      & 22.0          & 27.0          & 37.2          & 34.9 & 1.9                       & 39.7                     & 30.3 \\
InternViT-6B~\cite{chen2024internvl2}              & 21.5          & 27.8          & 37.9          & 32.8 & 0.7                       & 39.8                     & 30.0 \\
ViT-MAE-huge~\cite{he2022mae}                      & 23.2          & 27.3          & 34.7          & 33.3 & 2.7                       & 38.6                     & 29.6 \\
Qwen3-VL-2B-ViT~\cite{bai2025qwen3vl}              & 31.6          & 17.1          & 38.2          & 31.0 & 2.0                       & 38.6                     & 29.5 \\
ViT-MSN-base~\cite{assran2023msn}                  & 14.6          & 30.4          & 38.2          & 34.3 & 2.9                       & 38.2                     & 29.4 \\
Qwen3-VL-32B-ViT~\cite{bai2025qwen3vl}             & 25.7          & 28.8          & 37.8          & 24.9 & 5.1                       & 37.4                     & 29.3 \\
DINOv2-giant~\cite{oquab2023dinov2}                & 25.6          & 24.6          & 29.8          & 36.8 & 1.8                       & 38.4                     & 29.2 \\
Qwen2.5-VL-3B-ViT~\cite{bai2025qwen25vl}           & 23.9          & 31.5          & 28.4          & 30.5 & 3.2                       & 37.0                     & 28.6 \\
DeepSeek-VL2-7B-ViT~\cite{wu2024deepseekvl2}       & 22.8          & 31.8          & 36.8          & 20.8 & 2.0                       & 36.7                     & 28.1 \\
BEiT-base~\cite{bao2022beit}                       & 17.1          & 26.5          & 37.8          & 30.7 & 1.1                       & 37.0                     & 28.0 \\
DINOv2-registers-base~\cite{darcet2024registers}   & 20.3          & 22.4          & 34.7          & 32.9 & 1.9                       & 36.1                     & 27.6 \\
Qwen2.5-VL-7B-ViT~\cite{bai2025qwen25vl}           & 23.9          & 25.2          & 39.5          & 20.7 & 1.2                       & 36.0                     & 27.3 \\
LLaVA-OneVision-7B-ViT~\cite{li2024llavaonevision} & 22.7          & 21.8          & 30.8          & 29.2 & 1.2                       & 34.4                     & 26.1 \\
SigLIP-large~\cite{siglip2023}                     & 17.7          & 26.2          & 33.1          & 27.4 & 1.4                       & 34.4                     & 26.1 \\
SigLIP-base~\cite{siglip2023}                      & 19.8          & 23.2          & 27.4          & 30.4 & 1.2                       & 33.2                     & 25.2 \\
Qwen2-VL-72B-ViT~\cite{wang2024qwen2vl}            & 14.4          & 24.7          & 34.7          & 25.5 & 2.0                       & 32.4                     & 24.8 \\
DINOv2-base~\cite{oquab2023dinov2}                 & 26.7          & 23.4          & 20.6          & 27.4 & 1.4                       & 32.3                     & 24.5 \\
DINOv3-huge+~\cite{oquab2024dinov3}                & 21.5          & \bestcell{12.5} & 18.9          & 35.2 & 0.4                       & 29.2                     & 22.0 \\
DINOv3-base~\cite{oquab2024dinov3}                 & 19.0          & 26.8          & 30.7          & 10.4 & 0.7                       & 28.7                     & 21.7 \\
EVA-CLIP-8B~\cite{sun2023eva}                      & 12.3          & 26.0          & 14.6          & 27.9 & 0.7                       & 26.7                     & 20.2 \\
CLIP ViT-L/14~\cite{radford2021learning}           & 12.1          & 17.0          & 10.1          & 32.4 & 0.4                       & 23.7                     & 17.9 \\
DeepSeek-VL2-small-ViT~\cite{wu2024deepseekvl2}    & 16.0          & 25.0          & 5.3           & 23.4 & 0.7                       & 23.0                     & 17.4 \\
CLIP ViT-H/14~\cite{radford2021learning}           & \bestcell{12.0} & 20.5          & 6.4           & 29.8 & 0.7                       & 22.7                     & 17.2 \\
CLIP ViT-B/32~\cite{radford2021learning}           & 25.3          & 23.6          & 7.7          & \bestcell{9.1}  & \bestcell{0.3} & 21.8          & 16.4          \\
DINOv2-registers-giant~\cite{darcet2024registers}  & 20.0          & 22.1          & 9.2           & 11.4 & 1.4                       & 20.4                     & 15.7 \\
CLIP ViT-B/16~\cite{radford2021learning}           & 14.9          & 18.2          & 9.3           & 17.5 & 0.4                       & \bestcell{19.8}            & 15.0 \\
LLaVA-NeXT-7B-ViT~\cite{liu2023improvedllava}          & 17.3          & 17.4          & \bestcell{3.5} & 21.4          & 0.4          & \bestcell{19.8} & \bestcell{14.9} \\ 
\bottomrule
\end{tabular}
}
\end{table}

%% file: content/tables/mad_lora_matrix.tex
\begin{table}[tbp]
\centering
\caption{
{LoRA for MAD.} Test ACER$\downarrow$ (\%) after LoRA adaptation of the vision encoder on the features (compare~\cref{tab:feature_overview}). 
Overall results summarize intra-dataset (\colorbox{yellow!30}{ID}), cross-dataset (\colorbox{green!30}{CD}), and average performance.
\textbf{Bold}: best per column.
Full results:~\cref{tab:app:lora-mad-intra-dataset-acer}.
}
\label{tab:lora-mad-intra-dataset-acer}
\resizebox{0.85\linewidth}{!}{%
\begin{tabular}{lccrrr}
\toprule
\multirow{2}{*}{Model}                        & \multicolumn{2}{c}{Train/Dev} & \multicolumn{3}{c}{Overall}                               \\ \cmidrule(lr){2-3}
\cmidrule(lr){4-6}
                                              & FFHQ          & FRGC          & \cellcolor{yellow!30}{ID} & \cellcolor{green!30}{CD} & Avg          \\ 
\midrule
InternViT-300M~\cite{chen2024internvl2}       & 6.2           & 32.9          & \bestcell{0.0}              & 26.1                     & 19.6         \\
BEiT-base~\cite{bao2022beit}                  & 8.0           & 30.9          & 0.2                       & 25.8                     & 19.4         \\
ViT-MAE-base~\cite{he2022mae}                 & 5.8           & 30.2          & 0.9                       & 23.7                     & 18.0         \\
Qwen2-VL-72B-ViT~\cite{wang2024qwen2vl}       & 3.6           & 32.0          & 0.3                       & 23.7                     & 17.8         \\
SigLIP-large~\cite{siglip2023}                & 4.9           & 29.6          & 0.1                       & 23.0                     & 17.3         \\
Qwen3-VL-2B-ViT~\cite{bai2025qwen3vl}         & 2.3           & 29.4          & 0.1                       & 21.1                     & 15.8         \\
ViT-MAE-huge~\cite{he2022mae}                 & 5.7           & 24.7          & \bestcell{0.0}              & 20.2                     & 15.2         \\
Qwen2.5-VL-3B-ViT~\cite{bai2025qwen25vl}      & 2.6           & 27.5          & 0.5                       & 19.9                     & 15.0         \\
Qwen2.5-VL-7B-ViT~\cite{bai2025qwen25vl}      & 2.1           & 27.4          & 0.3                       & 19.6                     & 14.8         \\
ViT-MSN-large~\cite{assran2023msn}            & 4.6           & 23.2          & 0.7                       & 18.3                     & 13.9         \\
BEiT-large~\cite{bao2022beit}                 & 1.7           & 25.6          & 0.1                       & 18.2                     & 13.7         \\
DINOv3-base~\cite{oquab2024dinov3}            & 0.6           & 25.1          & 0.1                       & 17.1                     & 12.8         \\
DINOv2-base~\cite{oquab2023dinov2}            & 2.2           & 22.7          & 0.1                       & 16.6                     & 12.5         \\
DINOv3-H+~\cite{oquab2024dinov3}              & 0.7           & 24.0          & \bestcell{0.0}              & 16.5                     & 12.4         \\
InternViT-6B~\cite{chen2024internvl2}         & \bestcell{0.5}  & 21.8          & \bestcell{0.0}              & 14.9                     & 11.2         \\
DINOv2-R-base~\cite{darcet2024registers}      & 0.9           & 20.7          & 0.1                       & 14.4                     & 10.8         \\
\scalebox{0.875}{LLaVA-OneVision-7B-ViT~\cite{li2024llavaonevision}} & 0.8  & 20.7          & \bestcell{0.0} & 14.4          & 10.8          \\
CLIP ViT-B/32~\cite{radford2021learning}      & 5.7           & 14.9          & 0.2                       & 13.7                     & 10.3         \\
DINOv2-giant~\cite{oquab2023dinov2}           & \bestcell{0.5}  & 18.5          & \bestcell{0.0}              & 12.7                     & 9.5          \\
ViT-MSN-base~\cite{assran2023msn}             & 4.9           & 13.9          & 0.7                       & 12.4                     & 9.4          \\
DINOv2-R-giant~\cite{darcet2024registers}     & 1.6           & 17.2          & 0.1                       & 12.5                     & 9.4          \\
\scalebox{0.875}{DeepSeek-VL2-small-ViT~\cite{wu2024deepseekvl2}}    & 5.7  & 12.5          & \bestcell{0.0} & 12.2          & 9.1           \\
LLaVA-NeXT-7B-ViT~\cite{liu2023improvedllava}     & 2.1           & 15.3          & \bestcell{0.0}              & 11.6                     & 8.7          \\
SigLIP-base~\cite{siglip2023}                 & 2.0           & 14.9          & 0.1                       & 11.2                     & 8.5          \\
Qwen3-VL-32B-ViT~\cite{bai2025qwen3vl}        & 2.2           & 14.2          & 0.4                       & 10.8                     & 8.2          \\
DINOv3-large~\cite{oquab2024dinov3}           & 0.8           & 14.3          & \bestcell{0.0}              & 10.1                     & 7.5          \\
\scalebox{0.925}{DeepSeek-VL2-7B-ViT~\cite{wu2024deepseekvl2}} & 4.1  & 9.7           & \bestcell{0.0} & 9.2           & 6.9           \\
EVA-CLIP-8B~\cite{sun2023eva}                 & 1.6           & 11.0          & \bestcell{0.0}              & 8.4                      & 6.3          \\
CLIP ViT-H/14~\cite{radford2021learning}      & 0.8           & 9.9           & \bestcell{0.0}              & 7.1                      & 5.3          \\
CLIP ViT-L/14~\cite{radford2021learning}      & 1.1           & \bestcell{6.3}  & 0.1                       & \bestcell{4.9}             & \bestcell{3.7} \\ 
\bottomrule
\end{tabular}
}
  \vspace{-5pt}
\end{table}

%% file: content/05_discussions.tex
\section{Discussion}
\label{sec:discussion}

\subsection{How much access to the model is needed?}
 
\Cref{tab:summary_pad,tab:summary_mad} represents top-performing models from the five different approaches we used in our study and compares them against baselines. For PAD, the best overall average ACER falls from $32.9\%$ under
zero-shot prompting to $31.0\%$ with logit sampling, $24.4\%$ for the best
frozen encoder with a linear probe, $16.3\%$ for PADLLM, and $14.9\%$ for a
LoRA-adapted encoder; for MAD the corresponding values are $19.4\%$, $8.0\%$,
$5.0\%$, $2.9\%$ and $3.7\%$.

On both tasks, zero-shot evaluation leads to poor performance, however logit sampling through next-token distribution can improve the average ACER ($31.0\%$ instead of $32.9\%$ for PAD, and $19.4\%$ instead of $11.0\%$ for MAD for InternVL3-8B). 
In addition, a frozen vision-encoder with a linear probe outperforms an MLLM with logit sampling  ($24.4\%$ against $31.0\%$ for PAD, $5.0\%$
against $8.0\%$ for MAD). Parameter-efficient adaptation of  MLLM or only vision-encoder can further improve the performance of the model, achieving the best performance for MAD and PAD, respectively, and outperforming baselines from the literature.
 
For MAD a frozen vision-encoder with a
linear head (EVA-CLIP-8B, $5.0\%$) already outperforms the baseline from the literature, and backbone adaptation improves $1.3\%$ ACER. 
For PAD, a frozen vision-encoder is clearly insufficient: the best linear probe (InternViT-6B, $24.4\%$) remains $7.1\%$  behind FLIP-MCL, and adaptation of
the vision-encoder backbone is necessary, not optional. 
Once the backbone is trained, however, it is $1.4\%$ different with fine-tuning MLLM
for PAD ($14.9$-$16.3\%$) and $0.8\%$  for MAD ($2.9$-$3.7\%$), so the choice between fine-tuning an MLLM or a vision-encoder is governed by computation cost and application rather than by accuracy.

 \begin{table}[tbp]
\centering
\caption{PAD ACER$\downarrow$ (\%) for different methods  on  MCIO datasets, \colorbox{yellow!30}{ID} is the intra-dataset and \colorbox{green!30}{CD} is the cross-dataset performance, and Avg is the overall aggregated average. 
Full results:~\cref{tab:app:summary_pad_full_detail}.
}
\label{tab:summary_pad}
\footnotesize
\resizebox{0.95\linewidth}{!}{%
\begin{tabular}{llccc}
\toprule
Method & Model & \cellcolor{yellow!30}ID & \cellcolor{green!30}CD & Avg \\
\midrule
\multirow{3}{*}{Zero-shot}
& FaceLLM-8B~\cite{shahreza2025facellm}             & 32.2 & 42.4 & 39.8 \\
& InternVL3-8B~\cite{zhu2025internvl3}               & {31.3} & {33.4} & {32.9} \\
& Ovis1.5-Llama3-8B~\cite{lu2024ovis}  & 41.8 & 44.6 & 43.9 \\
\midrule
\multirow{3}{*}{Logit Sampling}
& InternVL3-8B~\cite{zhu2025internvl3}               & {16.2} & 36.0 & {31.0} \\
& FaceLLM-8B~\cite{shahreza2025facellm}               & 20.9& {34.8} & 31.3 \\
& Qwen2-VL-7B-Instruc \cite{wang2024qwen2vl}          & {20.0} & {43.2}& {37.4}\\
\midrule
MLLM (LoRA) & PADLLM-8B [ours]            & {13.6}  & \textbf{17.2 }& {16.3 }\\
\midrule
\multirow{3}{*}{Vision Encoder (LP)}
& InternViT-6B~\cite{chen2024internvl2}              & {1.6 }& 32.0 & {24.4}\\
& CLIP ViT-L/14~\cite{radford2021learning}            & 10.6 & 34.0 & 28.2 \\
& CLIP ViT-B/32~\cite{radford2021learning}            & 12.6 & {31.6} & 26.8 \\
\midrule
\multirow{3}{*}{Vision Encoder (LoRA)}
& LLaVA-NeXT-7B-ViT~\cite{liu2023improvedllava}           & \textbf{0.4 }& {19.8} & \textbf{14.9 }\\
& CLIP ViT-B/16~\cite{radford2021learning}            & \textbf{0.4 }& {19.8} & 15.0 \\
& DINOv2-registers-giant~\cite{darcet2024registers}   & 1.4 & 20.4 & 15.7 \\
\midrule
\multicolumn{5}{c}{PAD Baselines} \\
\midrule
\multirow{7}{*}{PAD Baselines}
& DeepPixBiS~\cite{george2019deep}           & 3.4 & 43.1 & 33.2 \\ 
& FSFM-FAS~\cite{wang2025fsfm}               & 4.2 & 37.3 & 29.0 \\
& FoundPAD ViT-B/16~\cite{ozgur2025foundpad} & 10.7 & 34.7 & 28.7 \\
& FoundPAD ViT-L/14~\cite{ozgur2025foundpad} & 8.2 & 32.6 & 26.5 \\
& FLIP-V~\cite{srivatsan2023flip}             & 1.5  & 28.1 & 21.5 \\
& FLIP-IT~\cite{srivatsan2023flip}            & \textbf{0.5}  & 25.8 & 19.5 \\
& FLIP-MCL~\cite{srivatsan2023flip}           & 0.7  & \textbf{22.9} & \textbf{17.3} \\
\bottomrule
\end{tabular}
}
\end{table}

\begin{table}[tbp]
\centering
\caption{MAD ACER$\downarrow$ (\%) for different methods, all train/dev on FFHQ/FRGC, all evaluated on four FFHQ, FRGC, FRLL, and FERET datasets; 
\colorbox{yellow!30}{ID} is the intra-dataset and \colorbox{green!30}{CD} is the cross-dataset performance, and Avg is the overall aggregated average. 
Full results:~\cref{tab:app:summary_mad_full_detail}.
}
\label{tab:summary_mad}
\footnotesize
\resizebox{0.95\linewidth}{!}{%
\begin{tabular}{llccc}
\toprule
Method & Model & \cellcolor{yellow!30}ID & \cellcolor{green!30}CD & Avg \\
\midrule
\multirow{3}{*}{Zero-shot}
& FaceLLM-8B~\cite{shahreza2025facellm}             & 33.8 & 34.6 & 34.4 \\
& InternVL3-8B~\cite{zhu2025internvl3}               & {14.1} & {21.1} & {19.4 }\\
& Ovis1.5-Llama3-8B~\cite{lu2024ovis}  & 25.9 & 22.8 & 23.6 \\
\midrule
\multirow{3}{*}{Logit Sampling}
& InternVL3-8B~\cite{zhu2025internvl3}               & 3.5 & 13.5 & 11.0 \\
& FaceLLM-8B~\cite{shahreza2025facellm}               & 3.6& {9.5} & {8.0} \\
& Ovis1.5-Llama3-8B~\cite{lu2024ovis}                                     & {1.7} & {20.2}& {15.6}\\
\midrule
MLLM (LoRA) & MADLLM-8B [ours]            & {3.7}  &\textbf{ 2.7} & \textbf{2.9} \\
\midrule
\multirow{3}{*}{Vision Encoder (LP)}
& CLIP ViT-L/14~\cite{radford2021learning}          & 1.4 & 17.5 & 13.5 \\
&   CLIP ViT-H/14~\cite{radford2021learning}            & {0.6} & 15.9 & 12.1 \\
& EVA-CLIP-8B~\cite{sun2023eva} &{0.6 }& {6.4 }&{5.0}\\
\midrule
\multirow{3}{*}{Vision Encoder (LoRA)}
& CLIP ViT-L/14~\cite{radford2021learning}          & 0.1 & {4.9 }& {3.7 }\\
&   CLIP ViT-H/14~\cite{radford2021learning}            & \textbf{0.0 }& 7.1 & 5.3 \\
& EVA-CLIP-8B~\cite{sun2023eva}  & \textbf{0.0 }& 8.4 & 6.3 \\
\midrule
\multicolumn{5}{c}{MAD Baselines} \\
\midrule
\multirow{5}{*}{MAD Baselines}
& OrthoMAD~\cite{neto2022orthomad}                & 9.3 & 34.8 & 28.5 \\
& IDistill~\cite{caldeira2023idistill}            & 3.3 & 26.4 & 20.7 \\
&  MADation ViT-L/14~\cite{caldeira2025madation}  & \textbf{0.2}   & 19.2 & 14.5 \\
& MADation ViT-B/16~\cite{caldeira2025madation}             & 0.4  & 15.5 & 11.7 \\
&  SelfMAD (HRNet-W18)~\cite{ivanovska2025selfmad}             & 6.5  & \textbf{12.9} & \textbf{11.3} \\
\bottomrule
\end{tabular}
}
\end{table}

\subsection{Generic representations in foundation models versus dedicated detectors}
 
How far generic pretraining goes against purpose-built detectors depends on the
task, and the difference is larger than the leading rows of
\cref{tab:summary_pad,tab:summary_mad} suggest. Counting over the full model
zoo in \Cref{tab:intra-dataset-acer,tab:mad-ffhq-frgc-acer}, no frozen encoder out of $30$ reaches the average ACER of FLIP-MCL
($17.3\%$) for PAD, and only one (\textit{i.e.,} EVA-CLIP-8B) beats SelfMAD ($11.3\%$) for
MAD. 
After LoRA adaptation, $5$ of $30$ outperform clear FLIP-MCL for PAD, against
$16$ of $30$ that clear SelfMAD for MAD (\Cref{tab:lora-intra-dataset-acer,tab:lora-mad-intra-dataset-acer}). 
The claim that general-purpose representations surpass task-specific design is
thus a fairly generic statement for MAD, but for PAD it holds only for a
particular handful of backbones and only after adaptation.
 
Among the baselines of PAD, FLIP-MCL,
FLIP-IT and FLIP-V reach $22.9$, $25.8$ and $28.1\%$ cross-dataset ACER, ahead
of FoundPAD at $32.6$-$34.7\%$, whereas FSFM-FAS and DeepPixBiS are among the
lowest intra-dataset errors of the block ($4.2$ and $3.4\%$) with the weakest
transfer of any method in either table ($37.3$ and $43.1\%$). 
For MAD, 
MADation ViT-L/14 gets near-zero intra-dataset error ($0.2\%$) but $19.2\%$
cross-dataset, while SelfMAD is the only baseline in either table whose transfer
error stays close to its source error ($6.5\%$ vs. $12.9\%$), which we
attribute to its self-supervised morph synthesis rather than to its backbone.
 
\subsection{Scale provides separability, not adaptability}
 
Across the models evaluated in \Cref{tab:intra-dataset-acer,tab:mad-ffhq-frgc-acer,tab:lora-intra-dataset-acer,tab:lora-mad-intra-dataset-acer}, the number of parameters correlates with frozen performance but
not with adapted performance. Over the $30$ vision encoders, the Spearman correlation
between vision-tower size (reported in \cref{tab:feature_overview})  and overall average ACER
is $\rho=-0.36$ ($p=0.06$) for PAD and $\rho=-0.34$ ($p=0.07$) for MAD under
linear probing, but falls to $\rho=-0.11$ ($p=0.58$) and $\rho=-0.19$ ($p=0.32$)
after LoRA adaptation. 
Therefore, scaling the number of parameters in vision-encoder can lead to  separability of a fixed representation; but it does not improve performance when adapting the vision-encoder.
 
The individual models can have significantly different performance between linear probe and adaptation.  For example, InternViT-6B ($5.5$B) is the best frozen PAD
encoder ($24.4\%$) but ranks $26$th of $30$ after adaptation ($30.0\%$). In contrast, 
LLaVA-NeXT-7B-ViT ($0.32$B) moves from $14$th to $1$st (from $33.5\%$ to $14.9\%$ ACER);
for MAD. 
In fact, rank agreement
between the two approaches with vision-encoder (linear probe and adaptation) is correspondingly weak: $\rho=0.51$ ($p=0.004$) for
PAD and $\rho=0.33$ ($p=0.07$) for MAD. 
Therefore, selecting a backbone by linear probe alone would have chosen the wrong model for both tasks. 
 
 
Which pretraining objective transfers best remains task-dependent. As observed
in \cref{sec:loravisionencoder}: self-distillation (DINOv2/v3) and
discrete-token prediction (BEiT) are among the strongest frozen PAD probes,
whereas the large contrastive
vision-language encoders lead for MAD.

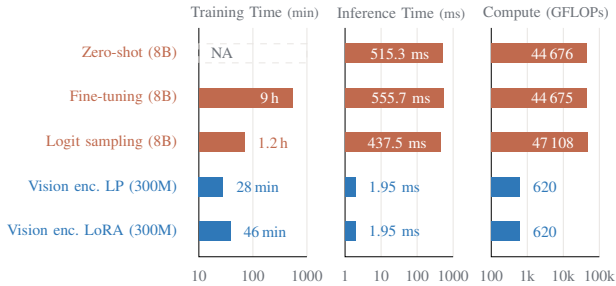
\begin{figure}[t] 
    \centering
    \resizebox{\linewidth}{!}{%
        \input{content/figures/time_plot}
    }
    \caption{
    Training and inference computational cost for our five different approaches on InternVL3-8B on the FFHQ dataset.
   Inference on 1,000 samples using an NVIDIA H100 GPU, and a logarithmic scale is used.
   }\label{fig:time_plot}
\end{figure}
 
\subsection{Complexity and Computational Cost}

\Cref{fig:time_plot} shows the complexity and computational cost of training and inference using InternVL-8B across our five approaches. 
As can be seen, a vision-encoder is roughly $264\times$ faster at
inference than the full $8$B MLLM ($1.95$\,ms against $515.3$\,ms) and needs
about $72\times$ fewer operations ($620$ against $44{,}676$ GFLOPs) in inference. In addition,  adapting the vision-encoder compared to MLLM
 is an order of magnitude faster ($46$\,min against $9$\,h). Because the
three MLLM settings share one backbone, their inference costs are comparable, and therefore the choice among them is driven by accuracy and
interpretability rather than by compute.
 
Based on the performance in \Cref{tab:summary_pad,tab:summary_mad} and \Cref{fig:time_plot}, we can conclude that where throughput or edge
deployment matters (\textit{e.g.,} automated gates, mobile capture, etc.), the LoRA-adapted vision-encoder is the clear choice: for PAD it is
simultaneously more accurate than PADLLM ($14.9\%$ against $16.3\%$) and two
orders of magnitude faster, and for MAD it is within $0.8\%$ average ACER of MADLLM at a fraction of the cost. 
In contrast, PADLLM and MADLLM
can additionally provide descriptions and
answer the operational accept/reject questions, therefore fine-tuning MLLM approach is justified when a human-readable
justification is part of the application, such as document issuance, border-control
adjudication, forensic reporting, etc. However, prompting an
off-the-shelf MLLM is not a deployable option given the high ACER values.

\subsection{Limitations and Future Work}
There are several limitations in our study that require further investigation in future.
\textit{(i) Unseen attacks.} Our PAD evaluation is limited to the MCIO benchmark, which primarily covers print and replay attacks and does not include other presentation instrument or deepfake-based attacks. In addition, the protocol holds out a {dataset} rather than an attack instrument, so generalization on presentation instrument is unexplored. Similarly, the MAD evaluation varies the corpus but keeps the morphing generator largely fixed, so the smaller cross-dataset gaps we report for MAD should not be read as robustness to unseen morphing methods. Our evaluation for MAD is also limited to the digital domain, and the performance of the detectors on print-scan morph attacks remains unexplored in this work.
\textit{(ii) Frame-level PAD.} The MCIO benchmark includes video datasets, but we operate on individual frames and therefore discard the temporal and physiological cues (such as blink, motion, etc.) that are known to aid liveness detection, which places a ceiling on our PAD results.
%
%
%
\textit{(iii) Evaluation Metric.} All results are reported as ACER at a threshold corresponding to EER on the  development set. However, other operating threshold may be used in practice, and exploring different thresholds can be studied in future work.
\textit{(iv) Parameter-efficient adaptation.} Parameter-efficient adaptation is restricted to a single LoRA configuration applied unchanged to MLLM and vision encoders of very different width and depth, rather than a per-model or broader comparison of parameter-efficient fine-tuning methods. 
\textit{(v) Interpretability of MLLMs.}
Beyond accuracy, PADLLM and MADLLM can generate textual descriptions alongside the score and answer questions
that were built into the instruction-tuning mixture. However, we only evaluated
the decision, not the explanations in our study. We did not measure whether the model is generating valid textual descriptions or may hallucinate. Systematic
factuality evaluation is therefore a necessary complement to the detection
results reported in our experiments.
\textit{(vi) Demographic fairness.} FairFaceGPT is demographically balanced
and was included partly for that reason, but we did not measure differential
error rates across demographic groups. 
\textit{(vii) Possible pretraining contamination.} FFHQ, FERET, FRGC, CASIA-FASD and Replay-Attack are publicly distributed, and we cannot exclude that some of their images appear in the pretraining corpora of the evaluated foundation models; the zero-shot results in particular should be interpreted with this in mind.

%% file: content/figures/time_plot.tex
\definecolor{costEightB}{RGB}{192,106,78}   
\definecolor{costThreeM}{RGB}{47,120,184}   
\definecolor{costGrey}{RGB}{110,115,122}
\definecolor{costRule}{RGB}{232,229,226}

\begin{tikzpicture}[
  font=\scriptsize,
  valin/.style={anchor=east, text=white, font=\scriptsize, xshift=-2.5pt},
  valout/.style={anchor=west, font=\scriptsize, xshift=2.5pt},
  tick/.style={costGrey, font=\scriptsize, anchor=north, yshift=-2.5pt},
  head/.style={costGrey, anchor=south west, font=\scriptsize},
  mlab/.style={anchor=east, font=\scriptsize},
  ratio/.style={costThreeM, anchor=north west, font=\scriptsize}
]
\def\W{1.7}\def\P{2.3}\def\BH{0.30}
\def\ra{0}\def\rb{-0.70}\def\rc{-1.40}\def\rd{-2.10}\def\re{-2.80}
\def\ytop{0.40}\def\ybot{-3.20}

\node[mlab, costEightB] at (-0.22,\ra) {Zero-shot (8B)};
\node[mlab, costEightB] at (-0.22,\rb) {Fine-tuning (8B)};
\node[mlab, costEightB] at (-0.22,\rc) {Logit sampling\ (8B)};
\node[mlab, costThreeM] at (-0.22,\rd) {Vision enc.\ LP (300M)};
\node[mlab, costThreeM] at (-0.22,\re) {Vision enc.\ LoRA (300M)};

\begin{scope}[xshift=0cm]
  \node[head] at (-0.25,\ytop) {Training Time \scalebox{0.9}{(min)}};
  \foreach \x in {0.5,1} \draw[costRule] (\x*\W,\ytop) -- (\x*\W,\ybot);
  \draw (0,\ytop) -- (0,\ybot) -- (\W,\ybot);
  \node[tick] at (0,\ybot) {10}; \node[tick] at (0.5*\W,\ybot) {100};
  \node[tick] at (\W,\ybot) {1000};
  \draw[costRule, dashed] (0,\ra-\BH/2) rectangle (\W,\ra+\BH/2);
  \node[costGrey, anchor=west, xshift=2pt] at (0,\ra) {NA};
  \fill[costEightB] (0,\rb-\BH/2) rectangle (1.471,\rb+\BH/2);
  \node[valin] at (1.471,\rb) {9\,h};
  \fill[costEightB] (0,\rc-\BH/2) rectangle (0.718,\rc+\BH/2);
  \node[valout, costEightB] at (0.718,\rc) {1.2\,h};
  \fill[costThreeM] (0,\rd-\BH/2) rectangle (0.380,\rd+\BH/2);
  \node[valout, costThreeM] at (0.380,\rd) {28\,min};
  \fill[costThreeM] (0,\re-\BH/2) rectangle (0.495,\re+\BH/2);
  \node[valout, costThreeM] at (0.495,\re) {46\,min};
\end{scope}

\begin{scope}[xshift=\P cm]
  \node[head] at (-0.25,\ytop) {Inference\ Time \scalebox{0.9}{(ms)}};
  \foreach \x in {0.3333,0.6667,1} \draw[costRule] (\x*\W,\ytop) -- (\x*\W,\ybot);
  \draw (0,\ytop) -- (0,\ybot) -- (\W,\ybot);
  \node[tick] at (0,\ybot) {1}; \node[tick] at (0.3333*\W,\ybot) {10};
  \node[tick] at (0.6667*\W,\ybot) {100}; \node[tick] at (\W,\ybot) {1000};
  \fill[costEightB] (0,\ra-\BH/2) rectangle (1.537,\ra+\BH/2);
  \node[valin] at (1.537,\ra) {515.3 ms};
  \fill[costEightB] (0,\rb-\BH/2) rectangle (1.556,\rb+\BH/2);
  \node[valin] at (1.556,\rb) {555.7 ms};
  \fill[costEightB] (0,\rc-\BH/2) rectangle (1.496,\rc+\BH/2);
  \node[valin] at (1.496,\rc) {437.5 ms};
  \fill[costThreeM] (0,\rd-\BH/2) rectangle (0.164,\rd+\BH/2);
  \node[valout, costThreeM] at (0.164,\rd) {1.95 ms};
  \fill[costThreeM] (0,\re-\BH/2) rectangle (0.164,\re+\BH/2);
  \node[valout, costThreeM] at (0.164,\re) {1.95 ms};
\end{scope}

\begin{scope}[xshift=2*\P cm]
  \node[head] at (-0.25,\ytop) {Compute \scalebox{0.95}{(GFLOPs)}};
  \foreach \x in {0.3333,0.6667,1} \draw[costRule] (\x*\W,\ytop) -- (\x*\W,\ybot);
  \draw (0,\ytop) -- (0,\ybot) -- (\W,\ybot);
  \node[tick] at (0,\ybot) {100}; \node[tick] at (0.3333*\W,\ybot) {1k};
  \node[tick] at (0.6667*\W,\ybot) {10k}; \node[tick] at (\W,\ybot) {100k};
  \fill[costEightB] (0,\ra-\BH/2) rectangle (1.501,\ra+\BH/2);
  \node[valin] at (1.501,\ra) {44\,676};
  \fill[costEightB] (0,\rb-\BH/2) rectangle (1.501,\rb+\BH/2);
  \node[valin] at (1.501,\rb) {44\,675};
  \fill[costEightB] (0,\rc-\BH/2) rectangle (1.514,\rc+\BH/2);
  \node[valin] at (1.514,\rc) {47\,108};
  \fill[costThreeM] (0,\rd-\BH/2) rectangle (0.449,\rd+\BH/2);
  \node[valout, costThreeM] at (0.449,\rd) {620};
  \fill[costThreeM] (0,\re-\BH/2) rectangle (0.449,\re+\BH/2);
  \node[valout, costThreeM] at (0.449,\re) {620};
\end{scope}
\end{tikzpicture}

%% file: content/06_conclusions.tex
\section{Conclusion}
In this paper, we asked whether general-purpose foundation models (FMs) and multimodal large language models (MLLMs) encode PAD-relevant and MAD-relevant information, and how best to use these models for these tasks. 
To answer this, we studied five approaches with increasing access to the internal information of the model, including zero-shot prompting, parameter-efficient fine-tuning with task-specific data resulting new MLLMs  (PADLLM and MADLLM), using next-token logit probabilities at the output of the MLLM, linear probing of frozen vision encoders, and fine-tuning of vision encoders of FMs and MLLMs for PAD and MAD. 
We used 16 open-weight MLLMs and 30 vision encoders, evaluating them on four PAD datasets (MSU-MFSD, CASIA-FASD, Replay-Attack, and OULU-NPU) and four MAD datasets (FFHQ, FRGC, FRLL, and FERET). 
The results indicate that general-purpose pretrained
representations contain substantial attack-relevant information. Using only a linear head on a frozen vision-encoder, EVA-CLIP-8B already outperforms all
dedicated MAD baselines. With LoRA adaptation, the top-performing generic encoders surpass task-specific PAD and MAD architectures on both tasks (14.9\% versus
17.3\% average ACER for PAD, 3.7\% versus 11.3\% for MAD). However, for PAD, only a small subset of adapted encoders outperform the strongest specialist.
Meanwhile, our results indicate that prompting an off-the-shelf MLLM is not a viable detector, given that the best zero-shot averages are 31.1\% for PAD and 19.4\% for MAD, with many models near chance. 
We also showed that logit sampling can improve the performance of pretrained MLLMs for PAD and MAD.
Our task-specific models, PADLLM and MADLLM, further reduce the average ACER of MLLMs from 31.1\% to 13.6\% and from 19.4\% to 2.9\%, respectively, while additionally
providing textual reasoning.
Our experiments with a vision-encoder also show that the pretraining objective matters more than parameter count, and the objective that transfers best is task-dependent: self-distillation and discrete-token pretraining favour PAD, large contrastive vision-language pretraining favours MAD, while prototype-based pretraining yields near-chance
features for both.

%% file: content/supplementary.tex
\clearpage
\appendices

\FloatBarrier
\section{Linear Probe and LoRA fine-tuning on PAD and MAD}

\Cref{tab:feature_overview} lists the 30 frozen backbones grouped by features, family, and scale for linear probing and LoRA fine-tuning.
For feature extraction, ViT-based encoders use the final \texttt{CLS} (classification) token when available. CLIP, SigLIP, and BEiT use the layer-normalized \texttt{CLS} representation, while Qwen, LLaVA, and DeepSeek vision encoders use mean-pooled patch tokens.
EVA-CLIP uses its final image embedding. These representations are used as input to the linear classification head.
For LoRA, adapters are inserted into the query (Q) and value (V) projections of each attention block. The table further distinguishes frozen backbone parameters from trainable LoRA and classification-head parameters.

This complements the linear-probe results for PAD and MAD~\cref{tab:intra-dataset-acer,tab:mad-ffhq-frgc-acer} and the LoRA fine-tuning results~\cref{tab:lora-intra-dataset-acer,tab:lora-mad-intra-dataset-acer} in the main paper. The corresponding full results are provided in the appendix in~\cref{tab:app:intra-dataset-acer,tab:app:mad-ffhq-frgc-acer,tab:app:lora-intra-dataset-acer,tab:app:lora-mad-intra-dataset-acer}, reporting results separately for each train--test dataset pair in addition to the ID and CD averages.
The model performance is ranked under ``Overall Average'' ACER and the best one is at the bottom of these tables.

\Cref{tab:feature_overview} lists the 30 frozen backbones grouped by family and scale.
ViT-based encoders use the final CLS token, while VLM vision towers use mean-pooled patch tokens unless otherwise specified.

\begin{table*}[bp]
\centering
\caption{
LP and LoRA.
Parameter sizes are marked as frozen \iconfrozen\hspace{0.5mm} ($10^9$) or trainable
\icontrainable\hspace{0.5mm} ($10^3$).
Feature denotes the pooling used to obtain one vector from each encoder (compare \cref{tab:intra-dataset-acer,tab:mad-ffhq-frgc-acer,tab:lora-intra-dataset-acer,tab:lora-mad-intra-dataset-acer}).
}
\label{tab:feature_overview}
\begin{tabular}{llllccrrr}
\toprule
\multirow{2}{*}{Model}
& \multirow{2}{*}{Input}
& \multirow{2}{*}{Supervision}
& \multirow{2}{*}{Pretraining objective}
& \multicolumn{2}{c}{Features}
& \multicolumn{3}{c}{Parameters} \\
\cmidrule(l){5-6} \cmidrule(r){7-9}
& & & &
{LoRA} & {FC}
& {\iconfrozen $\times 10^9$}
& {\icontrainable(}{LoRA},
& {FC)$\times 10^3$} \\
\midrule
DINOv2-base             & $224^2$          & SSL           & Self-distillation     & Q,V & CLS      & 0.09 & 294.9  & 0.8 \\
DINOv2-R-base           & $224^2$          & SSL           & Self-distillation     & Q,V & CLS      & 0.09 & 294.9  & 0.8 \\
DINOv2-giant            & $224^2$          & SSL           & Self-distillation     & Q,V & CLS      & 1.14 & 1966.1 & 1.5 \\
DINOv2-R-giant          & $224^2$          & SSL           & Self-distillation     & Q,V & CLS      & 1.14 & 1966.1 & 1.5 \\
DINOv3-base             & $224^2$          & SSL           & Self-distillation     & Q,V & CLS      & 0.09 & 294.9  & 0.8 \\
DINOv3-large            & $224^2$          & SSL           & Self-distillation     & Q,V & CLS      & 0.30 & 786.4  & 1.0 \\
DINOv3-H+               & $224^2$          & SSL           & Self-distillation     & Q,V & CLS      & 0.84 & 1310.7 & 1.3 \\
ViT-MAE-base            & $224^2$          & SSL           & Masked autoencoding   & Q,V & CLS      & 0.09 & 294.9  & 0.8 \\
ViT-MAE-huge            & $224^2$          & SSL           & Masked autoencoding   & Q,V & CLS      & 0.63 & 1310.7 & 1.3 \\
ViT-MSN-large           & $224^2$          & SSL           & Masked neighbours     & Q,V & CLS      & 0.30 & 786.4  & 1.0 \\
ViT-MSN-base            & $224^2$          & SSL           & Masked neighbours     & Q,V & CLS      & 0.09 & 294.9  & 0.8 \\
BEiT-base               & $224^2$          & SSL           & Masked image modeling & Q,V & LN CLS   & 0.09 & 294.9  & 0.8 \\
BEiT-large              & $224^2$          & SSL           & Masked image modeling & Q,V & LN CLS   & 0.30 & 786.4  & 1.0 \\
\midrule
SigLIP-base             & $224^2$          & VLM & Sigmoid contrastive   & Q,V & LN CLS   & 0.09 & 294.9  & 0.8 \\
SigLIP-large            & $256^2$          & VLM & Sigmoid contrastive   & Q,V & LN CLS   & 0.32 & 786.4  & 1.0 \\
CLIP ViT-B/32           & $224^2$          & VLM & Img--txt contrastive  & Q,V & LN CLS   & 0.09 & 294.9  & 0.8 \\
CLIP ViT-L/14           & $224^2$          & VLM & Img--txt contrastive  & Q,V & LN CLS   & 0.30 & 786.4  & 1.0 \\
CLIP ViT-H/14           & $224^2$          & VLM & Img--txt contrastive  & Q,V & LN CLS   & 0.63 & 1310.7 & 1.3 \\
EVA-CLIP-8B             & $224^2$          & VLM & Img--txt contrastive  & Q,V & Img emb. & 8.22 & 5505.0 & 1.3 \\
\midrule
Qwen2.5-VL-3B-ViT      & $224^2$          & MLLM          & Multimodal pretraining      & Q,V & Mean pool & 0.67 & 1310.7 & 2.0 \\
Qwen2.5-VL-7B-ViT      & $224^2$          & MLLM          & Multimodal pretraining      & Q,V & Mean pool & 0.68 & 1310.7 & 3.6 \\
Qwen2-VL-72B-ViT       & $224^2$          & MLLM          & Multimodal pretraining      & Q,V & Mean pool & 0.70 & 1310.7 & 8.2 \\
Qwen3-VL-2B-ViT        & $224^2$          & MLLM          & Multimodal pretraining      & Q,V & Mean pool & 0.41 & 786.4  & 2.0 \\
Qwen3-VL-32B-ViT       & $224^2$          & MLLM          & Multimodal pretraining      & Q,V & Mean pool & 0.60 & 995.3  & 5.1 \\
LLaVA-NeXT-7B-ViT      & $336^2{\times}5$ & MLLM          & Img--txt contrastive        & Q,V & Mean pool & 0.32 & 786.4  & 4.1 \\
LLaVA-OneVision-7B-ViT & $384^2{\times}5$ & MLLM          & Sigmoid contrastive         & Q,V & Mean pool & 0.41 & 958.5  & 3.6 \\
DeepSeek-VL2-small-ViT & $384^2{\times}5$ & MLLM          & Vision--language pretrain.  & Q,V & Mean pool & 0.65 & 995.3  & 2.0 \\
DeepSeek-VL2-7B-ViT    & $384^2{\times}5$ & MLLM          & Vision--language pretrain.  & Q,V & Mean pool & 0.78 & 995.3  & 2.6 \\
InternViT-300M         & $448^2$          & MLLM          & Vision--language alignment  & Q,V & CLS       & 0.30 & 786.4  & 1.0 \\
InternViT-6B           & $448^2$          & MLLM          & Vision--language alignment  & Q,V & CLS       & 5.54 & 4608.0 & 3.2 \\    
\bottomrule
\end{tabular}
\end{table*}

\input{content/tables/main_pad_frozen_matrix.tex}
\input{content/tables/main_mad_frozen_matrix.tex}

\input{content/tables/main_pad_lora_matrix.tex}
\input{content/tables/main_mad_lora_matrix.tex}

\section{MLLM Logit Sampling}

The results for PAD and MAD are detailed in \Cref{tab:pad-comparison-dev_supple} and \Cref{tab:mad-comparison-dev_supple}. These tables present performance metrics across three decision-extraction settings: Zero-shot, Ivanovska \textit{et al.} (I\&S)~\cite{ivanovska2026emergent}, and our logit-sampling approach, evaluated across various MLLM backbones and training/development protocols.

For PAD, we report results for the M, C, I, and O protocols, while MAD results focus on the FFHQ and FRGC settings. The tables include intra-dataset (ID), cross-dataset (CD), and overall averages, providing insight into performance across different protocols and datasets.

These detailed findings complement the aggregate comparisons in the main paper (\Cref{tab:pad-comparison-dev} and \Cref{tab:mad-comparison-dev}), facilitating a more granular assessment of each strategy across datasets and evaluation protocols.

\begin{table*}[tbp]
\centering
\caption{PAD ACER$\downarrow$ (\%), 
comparing three decision-extraction settings: {Zero-shot}, {Ivanovska and \v{S}truc (I\&S)~\cite{ivanovska2026emergent}}, and our logit-sampling approach. 
Overall is the average over all 16 cells, \colorbox{yellow!30}{ID} over the 4 intra-dataset cells, and \colorbox{green!30}{CD} over the 12 cross-dataset cells. 
For each backbone, the best value across the three settings is {bold}, while our method is additionally highlighted in blue.
}
\label{tab:pad-comparison-dev_supple}
\resizebox{\textwidth}{!}{%
\begin{tabular}{ll ccccc ccccc ccccc ccccc ccc}
\toprule
\multirow{2}{*}{Model} & \multirow{2}{*}{Method} & \multicolumn{5}{c}{Train/Dev on M} & \multicolumn{5}{c}{Train/Dev on C} & \multicolumn{5}{c}{Train/Dev on I} & \multicolumn{5}{c}{Train/Dev on O} & \multicolumn{3}{c}{Overall} \\
\cmidrule(lr){3-7} \cmidrule(lr){8-12} \cmidrule(lr){13-17} \cmidrule(lr){18-22} \cmidrule(lr){23-25}
 & & \cellcolor{yellow!30}M & \cellcolor{green!30}C & \cellcolor{green!30}I & \cellcolor{green!30}O & Avg & \cellcolor{green!30}M & \cellcolor{yellow!30}C & \cellcolor{green!30}I & \cellcolor{green!30}O & Avg & \cellcolor{green!30}M & \cellcolor{green!30}C & \cellcolor{yellow!30}I & \cellcolor{green!30}O & Avg & \cellcolor{green!30}M & \cellcolor{green!30}C & \cellcolor{green!30}I & \cellcolor{yellow!30}O & Avg & \cellcolor{yellow!30}ID & \cellcolor{green!30}CD & Avg \\
\midrule
\multirow{3}{*}{InternVL3-8B~\cite{zhu2025internvl3}}
 & Zero-shot & 29.1 & {18.9} & 48.0 & 39.5 & 33.9 & {26.8} & 9.2 & 47.3 & {44.3} & 31.9 & \textbf{26.8} & \textbf{9.2} & 47.3 & 44.3 & \textbf{31.9} & \textbf{29.1} & \textbf{18.9} & 48.0 & 39.5 & 33.9 & 31.3 & \textbf{33.4} & 32.9 \\
 & \scalebox{0.9}{I\&S~\cite{ivanovska2026emergent}}   & 26.7 & 24.0 & 47.3 & 42.6 & 35.2 & 43.5 & 8.0 & 40.4 & 48.7 & 35.2 & 29.9 & 17.3 & 45.2 & 44.8 & 34.3 & 33.7 & 34.0 & 48.0 & 40.5 & 39.1 & 30.1 & 37.9 & 35.9 \\
  \rowcolor[HTML]{D6ECFF} \cellcolor{white}
 & Ours      & \textbf{15.1} & 32.7 & \textbf{38.8} & \textbf{34.2} & \textbf{30.2} & 38.2 & \textbf{4.4} & \textbf{35.0} & 45.6 & \textbf{30.8} & 36.3 & 47.7 & \textbf{16.5} & \textbf{27.0} & \textbf{31.9} & 31.3 & 29.0 & \textbf{35.8} & \textbf{28.6} & \textbf{31.2} & \textbf{16.2} & 36.0 & \textbf{31.0} \\
\midrule
 \multirow{3}{*}{Ovis1.5-Llama3-8B~\cite{lu2024ovis}}
 & Zero-shot & 44.7 & 47.1 & 50.3 & \textbf{25.5} & 41.9 & 44.7 & 47.1 & 50.3 & \textbf{25.5} & 41.9 & 50.0 & 49.2 & 50.0 & 50.0 & 49.8 & 44.7 & 47.1 & 50.3 & \textbf{25.5} & \textbf{41.9} & 41.8 & 44.6 & 43.9 \\
 & \scalebox{0.9}{I\&S~\cite{ivanovska2026emergent}}   & 37.8 & \textbf{25.7} & \textbf{41.1} & 47.0 & \textbf{37.9} & \textbf{37.6} & 17.0 & \textbf{39.0} & 48.4 & \textbf{35.5} & \textbf{36.3} & \textbf{19.8} & 40.2 & \textbf{48.0} & \textbf{36.0} & \textbf{41.5} & \textbf{37.8} & \textbf{46.6} & 45.5 & 42.8 & 35.1 & \textbf{39.0} & \textbf{38.1} \\
 \rowcolor[HTML]{D6ECFF} \cellcolor{white}
 & Ours & \textbf{32.4} & 42.8 & 49.4 & 41.1 & 41.4 & 46.4 & \textbf{8.6} & 55.9 & 48.6 & 39.9 & 44.7 & 48.0 & \textbf{24.9} & 52.7 & 42.6 & 45.8 & 41.2 & 52.5 & 33.8 & 43.3 & \textbf{24.9} & 47.4 & 41.8 \\\midrule
 \multirow{3}{*}{\scalebox{0.9}{Qwen2-VL-7B-Instruct~\cite{wang2024qwen2vl}}}
 & Zero-shot & 39.4 & 38.9 & 43.0 & \textbf{32.3} & 38.4 & 48.0 & 47.1 & 47.3 & 50.0 & 48.1 & 39.4 & 38.9 & 43.0 & \textbf{32.3} & 38.4 & \textbf{39.4} & \textbf{38.9} & \textbf{43.0} & \textbf{32.3} & \textbf{38.4} & 40.5 & 41.0 & 40.8 \\
 & \scalebox{0.9}{I\&S~\cite{ivanovska2026emergent}}   & 33.0 & \textbf{20.8} & \textbf{40.3} & 45.5 & 34.9 & 33.2 & 13.9 & \textbf{36.6} & 47.6 & 32.8 & \textbf{34.0} & \textbf{9.9} & 33.0 & 49.0 & \textbf{31.3} & 43.4 & 43.1 & 50.0 & 41.4 & 44.4 & 30.3 & \textbf{37.7} & \textbf{35.9} \\
  \rowcolor[HTML]{D6ECFF} \cellcolor{white}
 & Ours  & \textbf{17.4} & 22.7 & 44.4 & 35.9 & \textbf{30.1} & \textbf{28.4} & \textbf{6.9} & 38.7 & \textbf{38.8} & \textbf{28.2} & 46.3 & 51.6 & \textbf{22.8} & 50.7 & 42.9 & 45.0 & 60.6 & 54.9 & 32.9 & 48.4 & \textbf{20.0} & 43.2 & 37.4 \\
\midrule
\multirow{3}{*}{FaceLLM-8B~\cite{shahreza2025facellm}}
 & Zero-shot & 35.6 & 39.8 & 50.4 & \textbf{22.2} & 37.0 & 49.6 & 26.2 & 44.7 & 50.0 & 42.7 & 49.6 & 26.2 & 44.7 & 50.0 & 42.7 & 35.6 & 39.8 & 50.4 & \textbf{22.2} & 37.0 & 32.2 & 42.4 & 39.8 \\
 & \scalebox{0.9}{I\&S~\cite{ivanovska2026emergent}}   & \textbf{24.3} & \textbf{21.3} & 48.3 & 42.0 & 34.0 & 38.1 & 6.5 & 51.6 & 47.9 & 36.0 & 35.5 & \textbf{7.8} & 51.0 & 47.0 & 35.3 & 34.8 & 34.1 & 50.6 & 36.5 & 39.0 & 29.6 & \textbf{7.8} & 36.1 \\
  \rowcolor[HTML]{D6ECFF} \cellcolor{white}
 & Ours      & 24.5 & 25.5 & \textbf{29.3} & 30.2 & \textbf{27.4} & \textbf{33.2} & \textbf{5.6} & \textbf{36.3} & \textbf{42.6} & \textbf{29.4} & \textbf{31.7} & 47.6 & \textbf{26.7} & \textbf{32.8} & \textbf{34.7} & \textbf{30.0} & \textbf{29.8} & \textbf{48.0} & 26.8 & \textbf{33.7} & \textbf{20.9} & 34.8 & \textbf{31.3} \\
\bottomrule
\end{tabular}%
}
\end{table*}

\begin{table*}[tbp]
\centering
\caption{MAD ACER$\downarrow$ (\%) comparing three decision-extraction settings: {Zero-shot}, {Ivanovska and \v{S}truc (I\&S)~\cite{ivanovska2026emergent}}, and our logit-sampling approach. 
Overall is the average over all 8 cells, \colorbox{yellow!30}{ID} over the 2 intra-dataset cells, and \colorbox{green!30}{CD} over the 6 cross-dataset cells. For each backbone, the best value across the three settings is \textbf{bold}, while our method is additionally highlighted in blue.}
  \setlength{\tabcolsep}{7.5pt}
\label{tab:mad-comparison-dev_supple}
\resizebox{\textwidth}{!}{%
\begin{tabular}{ll ccccc ccccc ccc}
\toprule
\multirow{2}{*}{Model} & \multirow{2}{*}{Method} & \multicolumn{5}{c}{Train/Dev on FFHQ} & \multicolumn{5}{c}{Train/Dev on FRGC} & \multicolumn{3}{c}{Overall} \\
\cmidrule(lr){3-7} \cmidrule(lr){8-12} \cmidrule(lr){13-15}
 & & \cellcolor{yellow!30}FFHQ & \cellcolor{green!30}FRGC & \cellcolor{green!30}FRLL & \cellcolor{green!30}FERET & Avg & \cellcolor{green!30}FFHQ & \cellcolor{yellow!30}FRGC & \cellcolor{green!30}FRLL & \cellcolor{green!30}FERET & Avg & \cellcolor{yellow!30}ID & \cellcolor{green!30}CD & Avg \\
\midrule
\multirow{3}{*}{InternVL3-8B~\cite{zhu2025internvl3}}
 & Zero-shot & 16.8 & 29.8 & 11.3 & 19.5 & 19.4 & 29.8 & 11.3 & 19.5 & 16.8 & 19.4 & 14.1 & 21.1 & 19.4 \\
 & \scalebox{0.9}{I\&S~\cite{ivanovska2026emergent}}   & 20.3 & \textbf{5.7} & \textbf{8.6} & 17.5 & 13.0 & 20.4 & 5.6 & 8.4 & 18.1 & 13.1 & 14.4 & \textbf{12.7} & 13.1 \\
  \rowcolor[HTML]{D6ECFF} \cellcolor{white}
 & Ours      & \textbf{3.8} & 8.7 & 28.0 & \textbf{9.1} & \textbf{12.4} & \textbf{20.0} & \textbf{0.0} & \textbf{3.2} & \textbf{15.4} & \textbf{9.7} & \textbf{3.5} & 13.5 & \textbf{11.0} \\
\midrule
 \multirow{3}{*}{Ovis1.5-Llama3-8B~\cite{lu2024ovis}}
 & Zero-shot & 15.2 & 19.6 & 36.5 & 23.0 & 23.6 & 19.6 & 36.6 &\textbf{23.0} & \textbf{15.2} & 23.6 & 25.9 & 22.8 & 23.6 \\
 & \scalebox{0.9}{I\&S~\cite{ivanovska2026emergent}}   & 15.3 & 28.6 & \textbf{7.02} & \textbf{18.6} & 17.4 & 22.5 & 15.7 & 23.3 & 30.4 & 23.0 & 15.5 & 21.7 & 20.2 \\
  \rowcolor[HTML]{D6ECFF} \cellcolor{white}
 & Ours      & \textbf{3.3} & \textbf{8.2} & 9.4 & 19.4 & \textbf{10.1} & \textbf{19.5} & \textbf{0.0} & 36.3 & 28.3 & \textbf{21.0} & \textbf{1.7} & \textbf{20.2} & \textbf{15.6} \\
\midrule
 \multirow{3}{*}{Qwen2-VL-7B-Instruct~\cite{wang2024qwen2vl}}
 & Zero-shot & 27.7 & 43.9 & 49.8 & 46.2 & 41.9 & 43.9 & 49.8 & 46.2 & 27.7 & 41.9 & 38.8 & 43.0 & 41.9 \\
 & \scalebox{0.9}{I\&S~\cite{ivanovska2026emergent}}   & 23.8 & 26.2 & 21.0 & \textbf{19.16} & 22.5 & 29.9 & 4.6 & \textbf{7.2} & \textbf{19.0} & \textbf{15.2} & 14.2 & \textbf{20.4} & 18.9 \\
  \rowcolor[HTML]{D6ECFF} \cellcolor{white}
 & Ours      & \textbf{0.6} & \textbf{4.8} & \textbf{8.0} & 26.8 & \textbf{10.0} & \textbf{24.0} & \textbf{0.0} & 47.3 & 19.4 & 22.7 & \textbf{0.3} & 21.7 & \textbf{16.4} \\
\midrule
\multirow{3}{*}{FaceLLM-8B~\cite{shahreza2025facellm}}
 & Zero-shot & 23.9 & 35.8 & 43.7 & 34.1 & 34.4 & 35.8 & 43.7 & 34.1 & 23.9 & 34.4 & 33.8 & 34.6 & 34.4 \\
 & \scalebox{0.9}{I\&S~\cite{ivanovska2026emergent}}   & 21.0 & 15.1 & \textbf{14.0} & 17.2 & 16.8 & 21.3 & 14.7 & 14.7 & 18.8 & 17.4 & 17.9 & 16.9 & 17.1 \\
  \rowcolor[HTML]{D6ECFF} \cellcolor{white}
 & Ours   & \textbf{4.5} & \textbf{7.7} & 14.6 & \textbf{7.6} & \textbf{8.6} & \textbf{15.2} & \textbf{0.1} & \textbf{2.6} & \textbf{11.9} & \textbf{7.5} & \textbf{3.6} & \textbf{9.5} & \textbf{8.0} \\
\bottomrule
\end{tabular}%
}
\end{table*}

\section{Full table of Discussion}

The main paper presents a compact summary of the PAD and MAD results in \cref{tab:summary_pad} and \cref{tab:summary_mad} for all methods used in the experiments. For completeness, the appendix provides the full results underlying these summaries in \cref{tab:app:summary_pad_full_detail} and \cref{tab:app:summary_mad_full_detail}, including the individual train/test results, intra-dataset (ID) and cross-dataset (CD) averages, and the overall average.

\begin{table*}[htbp]
\centering
\caption{PAD ACER$\downarrow$ (\%) for different methods, all train/dev on MCIO, all evaluation on four MCIO datasets; \colorbox{yellow!30}{ID} over the intra-dataset (Train = Test) and \colorbox{green!30}{CD} over the remaining cross datasets, and avg is the overall aggregated average.}
\label{tab:app:summary_pad_full_detail}
\footnotesize
\resizebox{\textwidth}{!}{%
\begin{tabular}{llcccccccccccccccccccrrrr}
\toprule
\multirow{2}{*}{Method}
& \multirow{2}{*}{Model}
& \multicolumn{5}{c}{Train on M}
& \multicolumn{5}{c}{Train on C}
& \multicolumn{5}{c}{Train on I}
& \multicolumn{5}{c}{Train on O}
& \multicolumn{3}{c}{Overall} \\
\cmidrule(lr){3-7}
\cmidrule(lr){8-12}
\cmidrule(lr){13-17}
\cmidrule(lr){18-22}
\cmidrule(lr){23-25}
& & \cellcolor{yellow!30}{\textbf{M}} & \cellcolor{green!30}{C} & \cellcolor{green!30}{I} & \cellcolor{green!30}{O} & Avg
& \cellcolor{green!30}{M} & \cellcolor{yellow!30}{\textbf{C}} & \cellcolor{green!30}{I} & \cellcolor{green!30}{O} & Avg
& \cellcolor{green!30}{M} & \cellcolor{green!30}{C} & \cellcolor{yellow!30}{\textbf{I}} & \cellcolor{green!30}{O} & Avg
& \cellcolor{green!30}{M} & \cellcolor{green!30}{C} & \cellcolor{green!30}{I} & \cellcolor{yellow!30}{\textbf{O}} & Avg
& \cellcolor{yellow!30}{ID} & \cellcolor{green!30}{CD} & Avg \\
\midrule

\multirow{3}{*}{Zero-shot}
& FaceLLM-8B~\cite{shahreza2025facellm}
& 35.6 & 39.8 & 50.4 & {22.2} & 37.0 & 49.6 & 26.2 & 44.7 & 50.0 & 42.7 & 49.6 & 26.2 & 44.7 & 50.0 & 42.7 & 35.6 & 39.8 & 50.4 & {22.2} & 37.0 & 32.2 & 42.4 & 39.8 \\
& InternVL3-8B~\cite{zhu2025internvl3}
& 29.1 & {18.9} & 48.0 & 39.5 & 33.9 & {26.8} & 9.2 & 47.3 & {44.3} & 31.9 & {26.8} & {9.2} & 47.3 & 44.3 & {31.9} & {29.1} & {18.9} & 48.0 & 39.5 & 33.9 & 31.3 & {33.4} & 32.9 \\
& Ovis1.5-Llama3-8B~\cite{lu2024ovis}
& 44.7 & 47.1 & 50.3 & {25.5} & 41.9 & 44.7 & 47.1 & 50.3 & {25.5} & 41.9 & 50.0 & 49.2 & 50.0 & 50.0 & 49.8 & 44.7 & 47.1 & 50.3 & {25.5} & {41.9} & 41.8 & 44.6 & 43.9 \\

\midrule
\multirow{3}{*}{Logit Sampling}
& InternVL3-8B~\cite{zhu2025internvl3}
&{15.1 }& 32.7&38.8 &34.2 &30.2 &38.2 &{4.4 }&{35.0 }&45.6 &30.8 &36.3 &47.7 &{16.5 }&{27.0} &{31.9} &31.3 &{29.0} &{35.8 } & 28.6 &{31.2} & {16.2} & 36.0 & {31.0} \\
& FaceLLM-8B~\cite{shahreza2025facellm}
&24.5 &25.5 &{29.3} &{30.2} &{27.4} &33.2 &5.6 &36.3 &42.6 &29.4 &{31.7} &{47.6} &26.7 &32.8 &34.7 &{30.0 }&29.8 &48.0 &{26.8 }&33.7 & 20.9 & {34.8} & 31.3 \\
& Qwen2-VL-7B-Instruc~\cite{wang2024qwen2vl}
& 17.4&{22.7} &44.4 &35.9 &30.1 &{28.4} &6.9 &38.7 &{38.8} &{28.2} &46.3 &51.6 &22.8 &50.7 &42.9 &45.0 &60.0 &54.9 &32.9 &48.4 & {20.0} & {43.2} & {37.4} \\

\midrule
 MLLM (LoRA)
& PADLLM-8B [ours] & 
{16.2} & \textbf{14.4} & \textbf{37.1} & 3.2 & 17.7 & \textbf{17.5} & 8.3 & {26.6} & {7.3} & \textbf{14.9} & {17.5} & \textbf{8.3} & 26.6 & 7.3 & {14.9} & {16.2} & {14.4} & 37.1 & 3.2 & {17.7} & 13.6 & \textbf{17.2} & {16.3} \\

\midrule
\multirow{3}{*}{Vision Encoder (LP)}
& InternViT-6B~\cite{chen2024internvl2}
& {4.4} & {16.6} & 45.3 &{14.7} &{20.3} & 28.1 & {1.8} & {21.3} & 25.9 & {19.3} & 28.5 & {25.3} & {0.1} & 50.3 & {26.0} & 48.5 & 38.9 & 40.4 & {0.3} & 32.0& {1.6} & 32.0 & {24.4} \\
& CLIP ViT-L/14~\cite{radford2021learning}
&19.0 &44.1 &{35.0} & 43.9 & 35.5 &{27.3} & 6.9 & 24.8 & 19.2 & 19.5 & 29.2 & 35.5 & 9.5 & 46.3 & 30.1 & {19.6} & 36.8 & 47.0 & 7.1 & 27.6 & 10.6 & 34.0 & 28.2 \\
& CLIP ViT-B/32~\cite{radford2021learning}
&21.8 &48.3 &48.1 &21.8 &35.0 &27.4 &4.4 &34.8 &{18.7} &21.3 &{26.6 }&36.7 &18.6 &{27.4} &27.3 &25.7 &{24.0}&{39.3 }&5.7 & {23.7}& 12.6 & {31.6} & 26.8 \\

\midrule
\multirow{3}{*}{Vision Encoder (LoRA)}
& LLaVA-NeXT-7B-ViT~\cite{liu2023improvedllava}
& 1.3& {19.6}&47.3 &1.3 &17.3 & {29.8 }& \textbf{0.0 }&\textbf{20.1} &19.8 &{17.4 }&\textbf{3.2 }&\textbf{8.3} &\textbf{0.2 }&\textbf{2.4} &\textbf{3.5} & 30.5& 27.7& 27.4&\textbf{0.0} &21.4 & \textbf{0.4} & {19.8} & \textbf{14.9 } \\
& CLIP ViT-B/16~\cite{radford2021learning}
&\textbf{0.4} &20.4 &{38.2} &\textbf{0.5} &\textbf{14.9} &33.1 &1.1 &20.5 &\textbf{18.0} &18.2 &14.1 &17.9 &\textbf{0.2 }&5.1 &9.3 &18.1 &17.3 &34.7 &\textbf{0.0} &17.5 & \textbf{0.4} & {19.8} & 15.0 \\
& DINOv2-registers-giant~\cite{darcet2024registers}
& 2.9&21.5 &49.5 &6.2 &20.0 &40.1 &0.2 &21.6 &26.5 &22.1 &9.5 &17.6 &1.7 &8.2 &9.2 &\textbf{14.5 }&\textbf{6.8 }&\textbf{23.4} &0.8 &\textbf{11.4} & 1.4 & 20.4 & 15.7 \\

\midrule
\multicolumn{25}{c}{PAD Baselines} \\
\midrule


\multirow{7}{*}{PAD Baselines}
& DeepPixBiS~\cite{george2019deep}           & 7.9 & 47.1 & 49.7 & 36.5 & 35.3 & 33.7 & 2.0 & 30.7 & 30.2 & 24.2 & 47.4 & 61.3 & \bestcell{0.0} & 50.9 & 39.9 & 46.6 & 38.4 & 45.1 & 3.8 & 33.5 & 3.4 & 43.1 & 33.2 \\
& FSFM-FAS~\cite{wang2025fsfm}               & 2.8 & 39.0 & \bestcell{11.1} & 30.6 & 20.9 & 43.8 & 1.7 & 52.3 & 39.0 & 34.2 & 33.2 & 49.5 & 0.2 & 54.0 & 34.2 & 39.3 & 22.0 & 33.8 & 12.0 & 26.8 & 4.2 & 37.3 & 29.0 \\
& FoundPAD ViT-B/16~\cite{ozgur2025foundpad} & 19.1 & 48.6 & 49.3 & 45.1 & 40.5 & 25.4 & 10.5 & 32.9 & 20.7 & 22.4 & 39.7 & 25.8 & 11.6 & 32.5 & 27.4 & 26.1 & 27.7 & 43.2 & 1.7 & 24.7 & 10.7 & 34.7 & 28.7 \\
& FoundPAD ViT-L/14~\cite{ozgur2025foundpad} & 14.0 & 50.0 & 50.0 & 49.5 & 40.9 & 31.4 & 5.7 & 25.9 & 23.8 & 21.7 & 24.1 & 37.0 & 10.5 & 38.3 & 27.5 & 17.2 & \bestcell{13.4} & 30.8 & 2.8 & \bestcell{16.0} & 8.2 & 32.6 & 26.5 \\
& FLIP-V~\cite{srivatsan2023flip}            & \bestcell{0.0} & 24.6 & 38.1 & 18.7 & 20.3 & 30.9 & \bestcell{0.4} & \bestcell{20.3} & 8.0 & 14.9 & 25.5 & 36.5 & 0.8 & 22.0 & 21.2 & 37.3 & 29.7 & 46.1 & 4.7 & 29.4 & 1.5 & 28.1 & 21.5 \\
& FLIP-IT~\cite{srivatsan2023flip}           & 0.4 & 25.3 & 44.1 & 20.8 & 22.7 & 34.9 & 0.8 & 31.1 & 22.7 & 22.4 & 22.4 & \bestcell{16.8} & 0.8 & \bestcell{14.8} & \bestcell{13.7} & \bestcell{16.9} & 29.8 & 30.6 & \bestcell{0.1} & 19.3 & \bestcell{0.5} & 25.8 & 19.5 \\
& FLIP-MCL~\cite{srivatsan2023flip}          & 1.6 & \bestcell{20.8} & 38.0 & \bestcell{14.8} & \bestcell{18.8} & \bestcell{20.7} & 0.7 & 27.9 & \bestcell{5.0} & \bestcell{13.6} & \bestcell{21.9} & 31.1 & 0.3 & 18.2 & 17.9 & 32.0 & 25.9 & \bestcell{18.2} & \bestcell{0.1} & 19.1 & 0.7 & \bestcell{22.9} & \bestcell{17.3} \\

\bottomrule
\end{tabular}%
}
\end{table*}

\begin{table*}[htbp]
\centering
\caption{
MAD ACER$\downarrow$ (\%) for different methods, all train/dev on FFHQ/FRGC, all evaluated on four FFHQ, FRGC, FRLL, and FERET datasets; \colorbox{yellow!30}{ID} over the intra-dataset (Train = Test) and \colorbox{green!30}{CD} over the remaining cross datasets, and avg is the overall aggregated average.
}
\label{tab:app:summary_mad_full_detail}
\footnotesize
\resizebox{\linewidth}{!}{%
\begin{tabular}{llccccrccccrrrr}
\toprule
\multirow{2}{*}{Method}
& \multirow{2}{*}{Model}
& \multicolumn{5}{c}{Train on FFHQ}
& \multicolumn{5}{c}{Train on FRGC}
& \multicolumn{3}{c}{Overall} \\
\cmidrule(lr){3-7} \cmidrule(lr){8-12} \cmidrule(lr){13-15}
& & \cellcolor{yellow!30}{\textbf{\scalebox{0.825}{FFHQ}}}
& \cellcolor{green!30}{\scalebox{0.9}{FRGC}}
& \cellcolor{green!30}{\scalebox{0.9}{FRLL}}
& \cellcolor{green!30}{\scalebox{0.85}{FERET}}
& Avg
& \cellcolor{green!30}{\scalebox{0.9}{FFHQ}}
& \cellcolor{yellow!30}{\textbf{\scalebox{0.825}{FRGC}}}
& \cellcolor{green!30}{\scalebox{0.9}{FRLL}}
& \cellcolor{green!30}{\scalebox{0.85}{FERET}}
& Avg
& \cellcolor{yellow!30}{ID}
& \cellcolor{green!30}{CD}
& Avg \\
\midrule

\multirow{3}{*}{Zero-shot}
& FaceLLM-8B~\cite{shahreza2025facellm}
& 23.9 & 35.8 & 43.7 & 34.1 & 34.4 & 35.8 & 43.7 & 34.1 & 23.9 & 34.4 & 33.8 & 34.6 & 34.4 \\
& InternVL3-8B~\cite{zhu2025internvl3}
& {16.8} & 29.8 & {11.3} & {19.5} & {19.4} & 29.8 & {11.3} & {19.5} & 16.8 & {19.4} & {14.1} & {21.1} & {19.4} \\
& Ovis1.5-Llama3-8B~\cite{lu2024ovis}
& 15.2 & {19.6} & 36.5 & 23.0 & 23.6 & {19.6} & 36.6 &{23.0} & {15.2} & 23.6 & 25.9 & 22.8 & 23.6 \\

\midrule
\multirow{3}{*}{Logit Sampling}
& InternVL3-8B~\cite{zhu2025internvl3}
&3.8 &8.7 &28.0 &9.1 &12.4 &20.0 &\textbf{0.0} &3.2 &15.4 &9.7 & 3.5 & 13.5 & 11.0 \\
& FaceLLM-8B~\cite{shahreza2025facellm}
& 4.5& {7.7}&14.6 &{7.6} &{8.6} &{15.2} &0.1 &{2.6} &{11.9} &{7.5} & 3.6& {9.5} & {8.0} \\
& Ovis1.5-Llama3-8B~\cite{lu2024ovis}
&{3.3} &8.2 &{9.4} &19.4 &10.1 &19.5 &\textbf{0.0} &36.3 &28.3 &21.0 & {1.7} & {20.2}& {15.6}\\

\midrule
MLLM (LoRA) & MADLLM-8B [ours]
& 6.1 & \textbf{0.7} & {1.2} & {3.7} & 2.9 & \textbf{0.7} & 1.2 & 3.7 & \textbf{6.1} & \textbf{2.9} & 3.7 & \textbf{2.7} & \textbf{2.9} \\

\midrule
\multirow{3}{*}{Vision Encoder (LP)}
& CLIP ViT-L/14~\cite{radford2021learning}
&2.8 & 14.9& 13.6& 9.5& 10.2&23.7 &\textbf{0.0} &34.6 &8.8 &16.8 & 1.4 & 17.5 & 13.5 \\
& CLIP ViT-H/14~\cite{radford2021learning}
&{1.2 }&11.4 &1.7 &22.6 &9.2 &{22.3} &\textbf{0.0 }&16.2 &21.0 &14.9 & {0.6} & 15.9 & 12.1 \\
& EVA-CLIP-8B~\cite{sun2023eva}
& 1.3& {2.4}& {0.5}& {3.4}& {1.9}&24.9 &\textbf{0.0} &{1.5} &{6.0} &{8.1} & {0.6 }& {6.4 }&{5.0}\\

\midrule
\multirow{3}{*}{Vision Encoder (LoRA)}
& CLIP ViT-L/14~\cite{radford2021learning}
&0.2 &3.1 &\textbf{0.4} &\textbf{0.6} &1.1 &{11.5} &\textbf{0.0} &5.4 &8.1 &{6.3} & 0.1 & {4.9 }& {3.7 }\\
& CLIP ViT-H/14~\cite{radford2021learning}
& \textbf{0.0} &{1.0} &1.0 &1.0 &\textbf{0.8} &25.9 &\textbf{0.0} & \textbf{0.0 }&13.8 &9.9 & \textbf{0.0 }& 7.1 & 5.3 \\
& EVA-CLIP-8B~\cite{sun2023eva}
& 0.1& 1.1& 1.6& 3.7& 1.6&37.8 &\textbf{0.0} &\textbf{0.0} &{6.4} &11.0 & \textbf{0.0 }& 8.4 & 6.3 \\
\midrule
\multicolumn{15}{c}{MAD Baselines} \\
\midrule
\multirow{5}{*}{MAD Baselines}
& OrthoMAD~\cite{neto2022orthomad}                & 18.5 & 18.1 & \bestcell{2.9} & 14.6 & 13.5 & 50.6 & 0.2 & 63.1 & 59.8 & 43.4 & 9.3 & 34.8 & 28.5 \\
& IDistill~\cite{caldeira2023idistill}            & 6.6 & 8.3 & 5.6 & 33.2 & 13.4 & 40.2 & \bestcell{0.0} & 30.7 & 40.6 & 27.9 & 3.3 & 26.4 & 20.7 \\
& MADation ViT-L/14~\cite{caldeira2025madation}   & \bestcell{0.3} & \bestcell{3.5} & 4.2 & \bestcell{0.4} & \bestcell{2.1} & 45.5 & \bestcell{0.0} & 46.3 & 15.3 & 26.8 & \bestcell{0.2} & 19.2 & 14.5 \\
& MADation ViT-B/16~\cite{caldeira2025madation}   & 0.9 & 5.5 & 14.7 & 9.0 & 7.5 & 42.4 & \bestcell{0.0} & 9.6 & 11.5 & 15.9 & 0.4 & 15.5 & 11.7 \\
& SelfMAD \scalebox{0.875}{(HRNet-W18)~\cite{ivanovska2025selfmad}} & 12.7 & 10.0 & 11.6 & 13.1 & 11.8 & \bestcell{29.5} & 0.3 & \bestcell{5.3} & \bestcell{7.8} & \bestcell{10.7} & 6.5 & \bestcell{12.9} & \bestcell{11.3} \\
\bottomrule
\end{tabular}
}
\end{table*}

%% file: content/tables/main_pad_frozen_matrix.tex
\begin{table*}[htbp]
  \centering
\caption{
{LP for PAD.} Test ACER$\downarrow$ (\%) of frozen vision encoders with linear probes.
Overall results summarize intra-dataset (\colorbox{yellow!30}{ID}), cross-dataset (\colorbox{green!30}{CD}), and average performance. \textbf{Bold}: best per column within each block.
}
  \label{tab:app:intra-dataset-acer}
  \footnotesize
  \setlength{\tabcolsep}{4.5pt}
  \resizebox{\linewidth}{!}{%
  \begin{tabular}{lccccccccccccccccccccrrr}
    \toprule
    & \multicolumn{5}{c}{Train on M} & \multicolumn{5}{c}{Train on C} & \multicolumn{5}{c}{Train on I} & \multicolumn{5}{c}{Train on O} & \multicolumn{3}{c}{Overall} \\
    \cmidrule(lr){2-6} \cmidrule(lr){7-11} \cmidrule(lr){12-16} \cmidrule(lr){17-21} \cmidrule(lr){22-24}
    Model & \cellcolor{yellow!30}{\textbf{M}} & \cellcolor{green!30}{C} & \cellcolor{green!30}{I} & \cellcolor{green!30}{O} & Avg & \cellcolor{green!30}{M} & \cellcolor{yellow!30}{\textbf{C}} & \cellcolor{green!30}{I} & \cellcolor{green!30}{O} & Avg & \cellcolor{green!30}{M} & \cellcolor{green!30}{C} & \cellcolor{yellow!30}{\textbf{I}} & \cellcolor{green!30}{O} & Avg & \cellcolor{green!30}{M} & \cellcolor{green!30}{C} & \cellcolor{green!30}{I} & \cellcolor{yellow!30}{\textbf{O}} & Avg & \cellcolor{yellow!30}{ID} & \cellcolor{green!30}{CD} & Avg \\
    \midrule
    \multicolumn{24}{c}{Model Zoo} \\
    \midrule
    ViT-MSN-base~\cite{assran2023msn}                  & 47.9 & 58.6 & 61.6 & 56.0 & 56.0 & 54.1 & 51.7 & 64.0 & 58.0 & 57.0 & 49.7 & 62.4 & 50.6 & 54.5 & 54.3 & 52.0 & 52.6 & 48.2 & 36.1 & 47.2 & 46.5 & 56.0 & 53.6 \\
    ViT-MSN-large~\cite{assran2023msn}                 & 49.8 & 52.4 & 46.8 & 39.0 & 47.0 & 50.9 & 51.0 & 47.3 & 39.2 & 47.1 & 50.2 & 53.4 & 42.7 & 39.6 & 46.5 & 52.8 & 46.9 & 48.8 & 37.5 & 46.5 & 45.2 & 47.3 & 46.8 \\
    InternViT-300M~\cite{chen2024internvl2}            & 22.2 & 40.4 & 51.2 & 47.2 & 40.2 & 39.9 & 34.5 & 50.5 & 47.9 & 43.2 & 50.1 & 50.8 & 16.1 & 50.2 & 41.8 & 49.9 & 41.0 & 53.7 & 22.7 & 41.8 & 23.9 & 47.7 & 41.8 \\
    ViT-MAE-base~\cite{he2022mae}                      & 20.6 & 30.8 & 48.8 & 42.2 & 35.6 & 47.9 & 14.3 & 64.8 & 48.1 & 43.8 & 48.3 & 58.0 & 7.4 & 57.1 & 42.7 & 54.4 & 37.1 & 55.7 & 31.6 & 44.7 & 18.5 & 49.4 & 41.7 \\
    SigLIP-base~\cite{siglip2023}                      & 22.4 & 39.9 & 46.0 & 38.4 & 36.7 & 48.2 & 14.3 & 48.7 & 48.0 & 39.8 & 43.1 & 38.9 & 16.5 & 51.7 & 37.5 & 50.0 & 39.4 & 49.2 & 36.8 & 43.8 & 22.5 & 45.1 & 39.5 \\
    Qwen3-VL-2B-ViT~\cite{bai2025qwen3vl}              & 28.0 & 49.0 & 50.0 & 46.5 & 43.4 & 34.4 & 18.0 & 31.9 & 46.2 & 32.6 & 50.9 & 46.6 & 21.0 & 51.0 & 42.4 & 45.2 & 36.3 & 50.4 & 26.0 & 39.5 & 23.2 & 44.9 & 39.5 \\
    DeepSeek-VL2-7B-ViT~\cite{wu2024deepseekvl2}       & 20.9 & 20.6 & 50.9 & 39.4 & 32.9 & 33.6 & 10.0 & 50.1 & 46.2 & 35.0 & 49.4 & 50.0 & 13.0 & 50.0 & 40.6 & 51.3 & 39.8 & 51.1 & 25.0 & 41.8 & 17.2 & 44.4 & 37.6 \\
    DeepSeek-VL2-small-ViT~\cite{wu2024deepseekvl2}    & 17.8 & 22.3 & 48.4 & 33.1 & 30.4 & 47.7 & 3.5 & 38.1 & 49.3 & 34.6 & 50.5 & 47.1 & 13.8 & 49.9 & 40.3 & 49.4 & 40.2 & 50.4 & 23.9 & 41.0 & 14.7 & 43.9 & 36.6 \\
    Qwen2.5-VL-3B-ViT~\cite{bai2025qwen25vl}           & 19.6 & 35.6 & 50.2 & 33.5 & 34.7 & 42.3 & 6.5 & 43.8 & 46.5 & 34.8 & 53.0 & 44.8 & 21.3 & 48.7 & 42.0 & 41.0 & 36.6 & 42.8 & 17.2 & 34.4 & 16.2 & 43.2 & 36.5 \\
    SigLIP-large~\cite{siglip2023}                     & 27.3 & 32.2 & 50.7 & 28.6 & 34.7 & 48.4 & 9.5 & 48.0 & 48.7 & 38.6 & 45.5 & 28.3 & 24.4 & 40.5 & 34.7 & 47.5 & 37.3 & 48.9 & 15.9 & 37.4 & 19.3 & 42.0 & 36.3 \\
    ViT-MAE-huge~\cite{he2022mae}                      & 16.4 & 47.0 & \bestcell{14.6} & 39.7 & 29.4 & 47.5 & 23.4 & 41.7 & 47.2 & 40.0 & 52.0 & 49.5 & 2.1 & 60.9 & 41.1 & 38.0 & 39.9 & \bestcell{35.0} & 24.8 & 34.4 & 16.7 & 42.7 & 36.2 \\
    DINOv2-base~\cite{oquab2023dinov2}                 & 20.2 & 36.8 & 46.4 & 37.6 & 35.3 & 43.1 & 2.7 & 43.3 & 49.3 & 34.6 & 35.8 & 33.0 & 9.8 & 44.2 & 30.7 & 52.3 & 42.7 & 50.6 & 20.3 & 41.4 & 13.2 & 42.9 & 35.5 \\
    Qwen3-VL-32B-ViT~\cite{bai2025qwen3vl}             & 19.0 & 28.7 & 51.9 & 30.8 & 32.6 & 40.6 & 8.1 & 44.4 & 41.3 & 33.6 & 46.8 & 42.9 & 16.9 & 47.6 & 38.5 & 47.2 & 36.6 & 40.3 & 20.5 & 36.2 & 16.1 & 41.6 & 35.2 \\
    BEiT-base~\cite{bao2022beit}                       & 12.7 & 42.9 & 47.0 & 39.8 & 35.6 & 47.2 & 6.6 & 39.1 & 49.8 & 35.7 & 35.8 & 42.9 & 8.6 & 48.8 & 34.0 & 47.1 & 35.6 & 47.8 & 8.9 & 34.9 & 9.2 & 43.7 & 35.0 \\
    Qwen2.5-VL-7B-ViT~\cite{bai2025qwen25vl}           & 17.2 & 24.2 & 48.3 & 30.9 & 30.1 & 43.8 & 6.1 & 46.5 & 38.2 & 33.6 & 50.4 & 29.2 & 27.6 & 44.4 & 37.9 & 50.9 & 38.5 & 42.7 & 18.1 & 37.6 & 17.2 & 40.7 & 34.8 \\
    DINOv2-registers-base~\cite{darcet2024registers}   & 16.9 & 27.1 & 38.4 & 39.7 & 30.5 & 43.2 & 7.6 & 43.8 & 46.6 & 35.3 & 38.8 & 31.8 & 4.9 & 50.5 & 31.5 & 48.6 & 35.2 & 47.8 & 28.9 & 40.1 & 14.6 & 40.9 & 34.4 \\
    LLaVA-NeXT-7B-ViT~\cite{liu2023improvedllava}          & 14.7 & 44.9 & 43.3 & 39.4 & 35.6 & 31.0 & 12.8 & 40.7 & 38.6 & 30.8 & 33.8 & 31.1 & 17.5 & 24.9 & 26.8 & 51.8 & 36.7 & 50.0 & 24.2 & 40.7 & 17.3 & 38.9 & 33.5 \\
    DINOv3-base~\cite{oquab2024dinov3}                 & 12.8 & 41.1 & 44.6 & 30.2 & 32.2 & 44.9 & 3.8 & 38.4 & 38.4 & 31.4 & 49.7 & 47.9 & 6.1 & 50.0 & 38.4 & 30.4 & 39.6 & 38.2 & 14.9 & 30.8 & 9.4 & 41.1 & 33.2 \\
    LLaVA-OneVision-7B-ViT~\cite{li2024llavaonevision} & 16.8 & 26.3 & 36.1 & 35.1 & 28.6 & 34.1 & 3.9 & 37.5 & 47.5 & 30.8 & 37.6 & 35.6 & 5.8 & 51.2 & 32.5 & 51.1 & 32.9 & 46.7 & 29.3 & 40.0 & 13.9 & 39.3 & 33.0 \\
    DINOv3-huge+~\cite{oquab2024dinov3}                & 8.8 & 48.9 & 43.9 & 24.1 & 31.4 & 40.6 & \bestcell{0.6} & 45.2 & 33.1 & 29.9 & 36.0 & 48.9 & 7.9 & 48.2 & 35.3 & 35.6 & 46.7 & 50.0 & 5.5 & 34.4 & 5.7 & 41.8 & 32.7 \\
    DINOv2-registers-giant~\cite{darcet2024registers}  & 9.5 & 35.4 & 41.4 & 39.1 & 31.4 & 27.1 & 5.0 & 39.5 & 35.0 & 26.7 & 32.7 & 46.3 & 8.4 & 48.2 & 33.9 & 44.4 & 39.9 & 47.2 & 11.4 & 35.7 & 8.6 & 39.7 & 31.9 \\
    DINOv2-giant~\cite{oquab2023dinov2}                & 10.8 & 32.8 & 45.7 & 33.4 & 30.7 & 40.8 & 5.5 & 36.0 & 33.1 & 28.8 & 45.3 & 25.6 & 3.6 & 47.4 & 30.5 & 48.6 & 38.8 & 48.9 & 13.8 & 37.5 & 8.4 & 39.7 & 31.9 \\
    CLIP ViT-H/14~\cite{radford2021learning}           & 15.2 & 35.4 & 41.3 & 36.7 & 32.2 & 30.5 & 3.1 & 34.0 & 41.3 & 27.2 & 35.2 & 31.6 & 16.4 & 43.6 & 31.7 & 47.7 & 39.8 & 47.0 & 7.3 & 35.4 & 10.5 & 38.7 & 31.6 \\
    EVA-CLIP-8B~\cite{sun2023eva}                      & 12.2 & 40.8 & 40.7 & 31.8 & 31.4 & 30.0 & 2.1 & 25.9 & 39.9 & 24.5 & 33.6 & 35.0 & 15.3 & 41.2 & 31.3 & 46.2 & 31.0 & 49.0 & 11.2 & 34.3 & 10.2 & 37.1 & 30.4 \\
    BEiT-large~\cite{bao2022beit}                      & 14.5 & 25.2 & 37.0 & 20.8 & 24.4 & 44.3 & 9.1 & 39.0 & \bestcell{18.2} & 27.6 & 35.0 & 43.7 & 5.0 & 42.7 & 31.6 & 43.5 & 36.5 & 46.6 & 7.9 & 33.6 & 9.1 & 36.0 & 29.3 \\
    CLIP ViT-L/14~\cite{radford2021learning}           & 19.0 & 44.1 & 35.0 & 43.9 & 35.5 & 27.3 & 6.9 & 24.8 & 19.2 & 19.5 & 29.2 & 35.5 & 9.5 & 46.3 & 30.1 & \bestcell{19.6} & 36.8 & 47.0 & 7.1 & 27.6 & 10.6 & 34.0 & 28.2 \\
    Qwen2-VL-72B-ViT~\cite{wang2024qwen2vl}            & 9.6 & \bestcell{12.4} & 38.6 & 22.5 & 20.8 & \bestcell{25.1} & 2.1 & 45.6 & 32.5 & 26.3 & 41.4 & 33.1 & 14.0 & 46.2 & 33.7 & 43.6 & 31.4 & 42.4 & 8.5 & 31.5 & 8.5 & 34.6 & 28.1 \\
    CLIP ViT-B/32~\cite{radford2021learning}           & 21.8 & 48.3 & 48.1 & 21.8 & 35.0 & 27.4 & 4.4 & 34.8 & 18.7 & 21.3 & \bestcell{26.6} & 36.7 & 18.6 & 27.4 & 27.3 & 25.7 & \bestcell{24.0} & 39.3 & 5.7 & \bestcell{23.7} & 12.6 & 31.6 & 26.8 \\
    CLIP ViT-B/16~\cite{radford2021learning}           & 7.4 & 41.5 & 39.7 & 30.3 & 29.7 & 29.8 & 5.5 & 26.5 & 19.0 & 20.2 & 30.9 & \bestcell{18.7} & 6.3 & \bestcell{21.3} & \bestcell{19.3} & 34.1 & 42.9 & 42.8 & 4.2 & 31.0 & 5.9 & \bestcell{31.5} & 25.1 \\
    InternViT-6B~\cite{chen2024internvl2}              & \bestcell{4.4} & 16.6 & 45.3 & \bestcell{14.7} & \bestcell{20.3} & 28.1 & 1.8 & \bestcell{21.3} & 25.9 & \bestcell{19.3} & 28.5 & 25.3 & \bestcell{0.1} & 50.3 & 26.0 & 48.5 & 38.9 & 40.4 & \bestcell{0.3} & 32.0 & \bestcell{1.6} & 32.0 & \bestcell{24.4} \\
    \bottomrule
  \end{tabular}
  }
\end{table*}

%% file: content/tables/main_mad_frozen_matrix.tex
\begin{table*}[tbp]
  \centering
    \caption{
    {LP for MAD.} Cross-dataset test ACER$\downarrow$ (\%) for frozen backbones with linear probes.
    \colorbox{yellow!30}{ID} averages the 2 train\,=\,test cells, 
    \colorbox{green!30}{CD} the remaining 6 cells, and Avg all 8 cells. 
    \textbf{Bold}: best per column within each block.
    }
  \label{tab:app:mad-ffhq-frgc-acer}
  \footnotesize
  \setlength{\tabcolsep}{2.25pt}
  \resizebox{0.75\linewidth}{!}{%
  \begin{tabular}{lccccccccccrrr}
    \toprule
    & \multicolumn{5}{c}{Train on FFHQ} & \multicolumn{5}{c}{Train on FRGC} & \multicolumn{3}{c}{Overall} \\
    \cmidrule(lr){2-6} \cmidrule(lr){7-11} \cmidrule(lr){12-14}
    Model & \cellcolor{yellow!30}{\textbf{\scalebox{0.825}{FFHQ}}} & \cellcolor{green!30}{\scalebox{0.9}{FRGC}} & \cellcolor{green!30}{\scalebox{0.9}{FRLL}} & \cellcolor{green!30}{\scalebox{0.85}{FERET}} & Avg & \cellcolor{green!30}{\scalebox{0.9}{FFHQ}} & \cellcolor{yellow!30}{\textbf{\scalebox{0.825}{FRGC}}} & \cellcolor{green!30}{\scalebox{0.9}{FRLL}} & \cellcolor{green!30}{\scalebox{0.85}{FERET}} & Avg & \cellcolor{yellow!30}{ID} & \cellcolor{green!30}{CD} & Avg \\
    \midrule
    \multicolumn{14}{c}{Model Zoo} \\
    \midrule
    ViT-MSN-large~\cite{assran2023msn}                 & 43.8 & 21.4 & 50.0 & 47.5 & 40.7 & 46.6 & 4.2 & 47.1 & 44.8 & 35.7 & 24.0 & 42.9 & 38.2 \\
    ViT-MSN-base~\cite{assran2023msn}                  & 37.7 & 52.8 & 46.6 & 39.9 & 44.3 & 45.3 & 0.6 & 8.8 & 27.2 & 20.5 & 19.2 & 36.8 & 32.4 \\
    InternViT-300M~\cite{chen2024internvl2}            & 8.8 & 6.7 & 29.5 & 48.2 & 23.3 & 28.7 & 2.1 & 50.0 & 52.5 & 33.3 & 5.5 & 35.9 & 28.3 \\
    DINOv3-H+~\cite{oquab2024dinov3}                   & 8.3 & 40.2 & 34.0 & 43.9 & 31.6 & 40.5 & 0.7 & 26.0 & 22.7 & 22.5 & 4.5 & 34.5 & 27.0 \\
    ViT-MAE-huge~\cite{he2022mae}                      & 16.8 & 46.5 & 16.2 & 27.8 & 26.8 & 48.6 & \bestcell{0.0} & 35.1 & 9.6 & 23.3 & 8.4 & 30.6 & 25.1 \\
    DINOv2-R-giant~\cite{darcet2024registers}          & 9.2 & 41.7 & 38.5 & 30.9 & 30.1 & 46.1 & 0.6 & 0.7 & 29.0 & 19.1 & 4.9 & 31.2 & 24.6 \\
    ViT-MAE-base~\cite{he2022mae}                      & 20.7 & 30.3 & 13.0 & 40.2 & 26.0 & 48.9 & \bestcell{0.0} & 21.6 & 17.1 & 21.9 & 10.4 & 28.5 & 24.0 \\
    DINOv2-base~\cite{oquab2023dinov2}                 & 9.1 & 34.9 & 34.3 & 27.8 & 26.5 & 40.9 & 0.1 & 10.3 & 33.3 & 21.2 & 4.6 & 30.2 & 23.8 \\
    \scalebox{0.85}{LLaVA-OneVision-7B-ViT~\cite{li2024llavaonevision}}  & 1.8 & 9.1 & 46.4 & 44.1 & 25.3 & 41.7 & \bestcell{0.0} & \bestcell{0.0} & 39.1 & 20.2 & 0.9 & 30.1 & 22.8 \\
    LLaVA-NeXT-7B-ViT~\cite{liu2023improvedllava}          & 1.9 & 50.9 & 2.1 & 52.0 & 26.7 & 49.6 & \bestcell{0.0} & 0.3 & 23.4 & 18.3 & 0.9 & 29.7 & 22.5 \\
    SigLIP-base~\cite{siglip2023}                      & 12.5 & 22.2 & 31.4 & 23.2 & 22.3 & 37.1 & 0.1 & 35.0 & 18.2 & 22.6 & 6.3 & 27.9 & 22.5 \\
    DINOv2-R-base~\cite{darcet2024registers}           & 13.8 & 37.9 & 24.2 & 19.5 & 23.9 & 48.9 & 0.3 & 0.3 & 34.4 & 21.0 & 7.1 & 27.5 & 22.4 \\
    DINOv3-large~\cite{oquab2024dinov3}                & 9.0 & 40.8 & 18.6 & 45.3 & 28.4 & 39.6 & 0.3 & 3.7 & 21.7 & 16.3 & 4.7 & 28.3 & 22.4 \\
    BEiT-large~\cite{bao2022beit}                      & 9.4 & 15.1 & 11.8 & 30.0 & 16.6 & 38.5 & \bestcell{0.0} & 37.8 & 33.1 & 27.3 & 4.7 & 27.7 & 21.9 \\
    BEiT-base~\cite{bao2022beit}                       & 12.9 & 31.8 & 15.4 & 15.3 & 18.9 & 38.6 & \bestcell{0.0} & 31.4 & 25.1 & 23.8 & 6.5 & 26.3 & 21.3 \\
    DINOv2-giant~\cite{oquab2023dinov2}                & 9.0 & 31.0 & 35.3 & 19.5 & 23.7 & 42.6 & 0.4 & 1.0 & 24.4 & 17.1 & 4.7 & 25.6 & 20.4 \\
    \scalebox{0.85}{DeepSeek-VL2-small-ViT~\cite{wu2024deepseekvl2}}    & 2.8 & 10.8 & 10.2 & 55.8 & 19.9 & 44.0 & \bestcell{0.0} & \bestcell{0.0} & 36.0 & 20.0 & 1.4 & 26.1 & 20.0 \\
    Qwen3-VL-2B-ViT~\cite{bai2025qwen3vl}              & 11.6 & 4.2 & 31.2 & 38.0 & 21.2 & 26.7 & \bestcell{0.0} & 21.1 & 24.5 & 18.1 & 5.8 & 24.3 & 19.6 \\
    Qwen2.5-VL-3B-ViT~\cite{bai2025qwen25vl}           & 8.4 & 15.6 & 11.0 & 16.9 & 13.0 & 36.9 & \bestcell{0.0} & 47.3 & 20.3 & 26.1 & 4.2 & 24.7 & 19.5 \\
    DINOv3-base~\cite{oquab2024dinov3}                 & 11.0 & 31.2 & 12.6 & 33.3 & 22.0 & 39.3 & 0.2 & 3.9 & 18.2 & 15.4 & 5.6 & 23.1 & 18.7 \\
    Qwen2-VL-72B-ViT~\cite{wang2024qwen2vl}            & 4.8 & 19.2 & 15.1 & 9.5 & 12.2 & 41.0 & \bestcell{0.0} & 34.1 & 25.1 & 25.1 & 2.4 & 24.0 & 18.6 \\
    Qwen3-VL-32B-ViT~\cite{bai2025qwen3vl}             & 6.9 & 13.5 & 25.1 & 12.6 & 14.5 & 35.3 & \bestcell{0.0} & 31.1 & 18.0 & 21.1 & 3.4 & 22.6 & 17.8 \\
    Qwen2.5-VL-7B-ViT~\cite{bai2025qwen25vl}           & 7.0 & 13.3 & 12.2 & 19.6 & 13.0 & 38.4 & \bestcell{0.0} & 29.2 & 18.5 & 21.5 & 3.5 & 21.9 & 17.3 \\
    CLIP ViT-B/32~\cite{radford2021learning}           & 5.3 & 17.2 & 7.3 & 22.7 & 13.1 & 24.6 & 0.1 & 27.7 & 23.7 & 19.0 & 2.7 & 20.5 & 16.1 \\
    SigLIP-large~\cite{siglip2023}                     & 7.4 & 19.5 & 16.1 & 20.6 & 15.9 & 30.5 & 0.1 & 8.8 & 17.6 & 14.3 & 3.7 & 18.9 & 15.1 \\
    \scalebox{0.9}{DeepSeek-VL2-7B-ViT~\cite{wu2024deepseekvl2}}       & 1.1 & 6.0 & 4.2 & 50.8 & 15.5 & 34.2 & \bestcell{0.0} & \bestcell{0.0} & 24.0 & 14.5 & 0.6 & 19.8 & 15.0 \\
    InternViT-6B~\cite{chen2024internvl2}              & \bestcell{0.2} & \bestcell{0.0} & \bestcell{0.1} & 31.1 & 7.9 & \bestcell{12.9} & \bestcell{0.0} & 27.9 & 38.7 & 19.9 & \bestcell{0.1} & 18.5 & 13.9 \\
    CLIP ViT-L/14~\cite{radford2021learning}           & 2.8 & 14.9 & 13.6 & 9.5 & 10.2 & 23.7 & \bestcell{0.0} & 34.6 & 8.8 & 16.8 & 1.4 & 17.5 & 13.5 \\
    CLIP ViT-H/14~\cite{radford2021learning}           & 1.2 & 11.4 & 1.7 & 22.6 & 9.2 & 22.3 & \bestcell{0.0} & 16.2 & 21.0 & 14.9 & 0.6 & 15.9 & 12.1 \\
    EVA-CLIP-8B~\cite{sun2023eva}                      & 1.3 & 2.4 & 0.5 & \bestcell{3.4} & \bestcell{1.9} & 24.9 & \bestcell{0.0} & 1.5 & \bestcell{6.0} & \bestcell{8.1} & 0.6 & \bestcell{6.4} & \bestcell{5.0} \\
    \bottomrule
  \end{tabular}
  }
\end{table*}

%% file: content/tables/main_pad_lora_matrix.tex
\begin{table*}[htbp]
  \centering
\caption{
{PAD for LoRA.} 
Test ACER$\downarrow$ (\%) after LoRA adaptation of the attention of the vision encoder.
Overall results summarize intra-dataset (\colorbox{yellow!30}{ID}), cross-dataset (\colorbox{green!30}{CD}), and average performance. 
\textbf{Bold}: best per column within each block. 
}
  \label{tab:app:lora-intra-dataset-acer}
  \footnotesize
  \setlength{\tabcolsep}{4.5pt}
  \resizebox{\linewidth}{!}{%
  \begin{tabular}{lccccccccccccccccccccrrr}
    \toprule
    & \multicolumn{5}{c}{Train on M} & \multicolumn{5}{c}{Train on C} & \multicolumn{5}{c}{Train on I} & \multicolumn{5}{c}{Train on O} & \multicolumn{3}{c}{Overall} \\
    \cmidrule(lr){2-6} \cmidrule(lr){7-11} \cmidrule(lr){12-16} \cmidrule(lr){17-21} \cmidrule(lr){22-24}
    Model & \cellcolor{yellow!30}{\textbf{M}} & \cellcolor{green!30}{C} & \cellcolor{green!30}{I} & \cellcolor{green!30}{O} & Avg & \cellcolor{green!30}{M} & \cellcolor{yellow!30}{\textbf{C}} & \cellcolor{green!30}{I} & \cellcolor{green!30}{O} & Avg & \cellcolor{green!30}{M} & \cellcolor{green!30}{C} & \cellcolor{yellow!30}{\textbf{I}} & \cellcolor{green!30}{O} & Avg & \cellcolor{green!30}{M} & \cellcolor{green!30}{C} & \cellcolor{green!30}{I} & \cellcolor{yellow!30}{\textbf{O}} & Avg & \cellcolor{yellow!30}{ID} & \cellcolor{green!30}{CD} & Avg \\
    \midrule
    \multicolumn{24}{c}{Model Zoo} \\
    \midrule
    InternViT-300M~\cite{chen2024internvl2}            & 4.2 & 32.3 & 47.7 & 45.4 & 32.4 & 38.9 & 0.9 & 33.4 & 43.9 & 29.3 & 36.4 & 50.8 & 2.0 & 54.5 & 35.9 & 46.3 & 32.6 & 49.2 & 11.1 & 34.8 & 4.6 & 42.6 & 33.1 \\
    ViT-MSN-large~\cite{assran2023msn}                 & 4.2 & 29.8 & 50.0 & 21.2 & 26.3 & 23.0 & 3.5 & 47.0 & 23.5 & 24.2 & 52.4 & 50.5 & \bestcell{0.0} & 50.8 & 38.4 & 50.0 & 36.1 & 50.0 & 13.2 & 37.3 & 5.2 & 40.3 & 31.6 \\
    ViT-MAE-base~\cite{he2022mae}                      & 1.0 & 35.3 & 22.0 & 14.1 & 18.1 & 34.7 & 1.2 & 38.9 & 44.3 & 29.8 & 43.7 & 55.7 & 0.4 & 59.0 & 39.7 & 48.7 & 34.4 & 49.4 & 12.6 & 36.3 & 3.8 & 40.0 & 31.0 \\
    BEiT-large~\cite{bao2022beit}                      & 0.6 & 27.5 & 47.5 & 12.6 & 22.0 & 40.3 & 0.6 & 28.8 & 38.2 & 27.0 & 50.0 & 48.2 & 0.5 & 50.0 & 37.2 & 49.9 & 34.8 & 49.2 & 5.9 & 34.9 & 1.9 & 39.7 & 30.3 \\
    InternViT-6B~\cite{chen2024internvl2}              & 0.6 & 24.1 & 30.3 & 31.0 & 21.5 & 42.5 & 0.1 & 27.0 & 41.5 & 27.8 & 50.0 & 50.0 & 1.5 & 50.0 & 37.9 & 50.0 & 34.1 & 46.8 & 0.4 & 32.8 & 0.7 & 39.8 & 30.0 \\
    ViT-MAE-huge~\cite{he2022mae}                      & 3.5 & 32.8 & 40.3 & 16.2 & 23.2 & 34.9 & 1.4 & 32.2 & 40.8 & 27.3 & 36.9 & 47.9 & 0.3 & 53.6 & 34.7 & 47.3 & 34.0 & 46.4 & 5.4 & 33.3 & 2.7 & 38.6 & 29.6 \\
    Qwen3-VL-2B-ViT~\cite{bai2025qwen3vl}              & 2.8 & 37.2 & 47.4 & 39.1 & 31.6 & 29.8 & 0.3 & 22.3 & \bestcell{15.8} & 17.1 & 50.0 & 50.0 & 2.9 & 50.0 & 38.2 & 48.1 & 33.2 & 40.9 & 1.7 & 31.0 & 2.0 & 38.6 & 29.5 \\
    ViT-MSN-base~\cite{assran2023msn}                  & 1.0 & 26.6 & \bestcell{15.4} & 15.5 & 14.6 & 35.2 & 1.0 & 39.7 & 45.8 & 30.4 & 44.8 & 53.2 & 0.1 & 54.6 & 38.2 & 49.9 & 33.3 & 44.5 & 9.6 & 34.3 & 2.9 & 38.2 & 29.4 \\
    Qwen3-VL-32B-ViT~\cite{bai2025qwen3vl}             & 9.8 & 20.8 & 44.7 & 27.4 & 25.7 & 39.7 & 0.5 & 35.4 & 39.5 & 28.8 & 49.8 & 45.6 & 5.2 & 50.7 & 37.8 & 39.3 & 31.4 & 24.1 & 4.9 & 24.9 & 5.1 & 37.4 & 29.3 \\
    DINOv2-giant~\cite{oquab2023dinov2}                & 2.3 & 27.6 & 49.3 & 23.4 & 25.6 & 37.9 & 0.5 & 28.1 & 32.0 & 24.6 & 39.9 & 34.4 & 0.5 & 44.6 & 29.8 & 49.5 & 44.8 & 49.0 & 3.9 & 36.8 & 1.8 & 38.4 & 29.2 \\
    Qwen2.5-VL-3B-ViT~\cite{bai2025qwen25vl}           & 3.8 & 27.4 & 49.9 & 14.3 & 23.9 & 47.5 & 0.4 & 33.2 & 44.7 & 31.5 & 33.0 & 41.2 & 1.5 & 38.0 & 28.4 & 45.8 & 36.6 & 32.6 & 7.0 & 30.5 & 3.2 & 37.0 & 28.6 \\
    DeepSeek-VL2-7B-ViT~\cite{wu2024deepseekvl2}       & 4.2 & 23.3 & 44.7 & 18.9 & 22.8 & 42.3 & \bestcell{0.0} & 37.3 & 47.7 & 31.8 & 50.0 & 43.9 & 3.5 & 49.9 & 36.8 & 24.4 & 29.4 & 29.0 & 0.5 & 20.8 & 2.0 & 36.7 & 28.1 \\
    BEiT-base~\cite{bao2022beit}                       & 0.2 & 33.1 & 20.8 & 14.3 & 17.1 & 26.0 & 0.6 & 38.6 & 41.0 & 26.5 & 47.4 & 49.8 & 0.4 & 53.3 & 37.8 & 47.8 & 36.2 & 35.5 & 3.3 & 30.7 & 1.1 & 37.0 & 28.0 \\
    DINOv2-registers-base~\cite{darcet2024registers}   & 0.6 & 35.9 & 28.7 & 16.0 & 20.3 & 33.3 & 2.1 & 30.5 & 23.8 & 22.4 & 46.6 & 44.8 & 0.6 & 46.9 & 34.7 & 46.7 & 44.7 & 35.9 & 4.3 & 32.9 & 1.9 & 36.1 & 27.6 \\
    Qwen2.5-VL-7B-ViT~\cite{bai2025qwen25vl}           & 1.9 & 23.7 & 49.0 & 20.9 & 23.9 & 43.3 & 0.2 & 24.9 & 32.4 & 25.2 & 59.2 & 47.5 & 2.4 & 48.8 & 39.5 & 31.1 & 29.9 & 21.6 & 0.3 & 20.7 & 1.2 & 36.0 & 27.3 \\
    LLaVA-OneVision-7B-ViT~\cite{li2024llavaonevision} & 1.8 & 23.0 & 46.3 & 19.6 & 22.7 & 30.7 & 0.4 & 25.7 & 30.4 & 21.8 & 48.1 & 35.6 & 0.7 & 38.6 & 30.8 & 47.1 & 39.1 & 29.0 & 1.8 & 29.2 & 1.2 & 34.4 & 26.1 \\
    SigLIP-large~\cite{siglip2023}                     & 2.7 & 20.6 & 38.0 & 9.7 & 17.7 & 34.9 & 0.7 & 41.4 & 27.8 & 26.2 & 52.0 & 46.1 & 0.9 & 33.5 & 33.1 & 46.9 & 32.9 & 28.7 & 1.2 & 27.4 & 1.4 & 34.4 & 26.1 \\
    SigLIP-base~\cite{siglip2023}                      & 1.0 & 20.7 & 49.8 & 7.8 & 19.8 & 32.2 & 0.7 & 27.8 & 32.0 & 23.2 & 31.6 & 27.5 & 2.0 & 48.6 & 27.4 & 48.8 & 32.4 & 39.5 & 1.2 & 30.4 & 1.2 & 33.2 & 25.2 \\
    Qwen2-VL-72B-ViT~\cite{wang2024qwen2vl}            & 1.8 & 16.1 & 35.5 & 4.0 & 14.4 & 35.5 & 1.2 & 30.0 & 31.9 & 24.7 & 48.8 & 37.6 & 4.4 & 47.8 & 34.7 & 43.8 & 20.5 & 37.5 & 0.4 & 25.5 & 2.0 & 32.4 & 24.8 \\
    DINOv2-base~\cite{oquab2023dinov2}                 & 1.5 & 35.0 & 36.8 & 33.6 & 26.7 & 26.7 & 1.6 & 32.5 & 32.9 & 23.4 & 12.6 & 24.5 & 0.3 & 45.1 & 20.6 & 41.8 & 32.8 & 32.9 & 2.1 & 27.4 & 1.4 & 32.3 & 24.5 \\
    DINOv3-huge+~\cite{oquab2024dinov3}                & \bestcell{0.0} & 32.9 & 48.7 & 4.4 & 21.5 & \bestcell{17.9} & \bestcell{0.0} & \bestcell{15.3} & 16.8 & \bestcell{12.5} & 13.3 & 33.4 & 0.7 & 28.4 & 18.9 & 40.0 & 49.9 & 50.0 & 1.0 & 35.2 & 0.4 & 29.2 & 22.0 \\
    DINOv3-base~\cite{oquab2024dinov3}                 & 0.6 & 23.7 & 48.9 & 2.7 & 19.0 & 45.6 & 0.5 & 28.8 & 32.3 & 26.8 & 45.5 & 32.4 & 1.2 & 43.8 & 30.7 & 14.3 & \bestcell{4.3} & 22.5 & 0.5 & 10.4 & 0.7 & 28.7 & 21.7 \\
    EVA-CLIP-8B~\cite{sun2023eva}                      & 0.6 & 13.4 & 26.4 & 8.7 & 12.3 & 35.0 & 0.5 & 31.2 & 37.1 & 26.0 & 20.6 & 30.2 & 0.9 & 6.7 & 14.6 & 36.5 & 26.2 & 48.0 & 0.9 & 27.9 & 0.7 & 26.7 & 20.2 \\
    CLIP ViT-L/14~\cite{radford2021learning}           & \bestcell{0.0} & 22.4 & 18.0 & 8.0 & 12.1 & 29.9 & 0.1 & 19.2 & 18.7 & 17.0 & 9.2 & 15.3 & \bestcell{0.0} & 15.8 & 10.1 & 37.8 & 40.6 & 49.7 & 1.4 & 32.4 & 0.4 & 23.7 & 17.9 \\
    DeepSeek-VL2-small-ViT~\cite{wu2024deepseekvl2}    & 1.8 & 23.6 & 35.2 & 3.3 & 16.0 & 33.9 & 0.2 & 30.6 & 35.3 & 25.0 & 14.6 & 4.7 & \bestcell{0.0} & \bestcell{2.0} & 5.3 & 20.9 & 32.2 & 39.5 & 1.1 & 23.4 & 0.7 & 23.0 & 17.4 \\
    CLIP ViT-H/14~\cite{radford2021learning}           & 0.3 & \bestcell{11.9} & 32.1 & 3.8 & \bestcell{12.0} & 31.5 & \bestcell{0.0} & 20.4 & 30.3 & 20.5 & 10.8 & \bestcell{3.7} & 2.5 & 8.8 & 6.4 & 44.2 & 26.7 & 48.0 & 0.2 & 29.8 & 0.7 & 22.7 & 17.2 \\
    CLIP ViT-B/32~\cite{radford2021learning}           & 0.8 & 27.3 & 49.6 & 23.5 & 25.3 & 41.0 & 0.1 & 21.4 & 31.6 & 23.6 & 8.3 & 10.0 & 0.2 & 12.3 & 7.7 & \bestcell{7.3} & 16.4 & \bestcell{12.5} & \bestcell{0.0} & \bestcell{9.1} & \bestcell{0.3} & 21.8 & 16.4 \\
    DINOv2-registers-giant~\cite{darcet2024registers}  & 2.9 & 21.5 & 49.5 & 6.2 & 20.0 & 40.1 & 0.2 & 21.6 & 26.5 & 22.1 & 9.5 & 17.6 & 1.7 & 8.2 & 9.2 & 14.5 & 6.8 & 23.4 & 0.8 & 11.4 & 1.4 & 20.4 & 15.7 \\
    CLIP ViT-B/16~\cite{radford2021learning}           & 0.4 & 20.4 & 38.2 & \bestcell{0.5} & 14.9 & 33.1 & 1.1 & 20.5 & 18.0 & 18.2 & 14.1 & 17.9 & 0.2 & 5.1 & 9.3 & 18.1 & 17.3 & 34.7 & \bestcell{0.0} & 17.5 & 0.4 & \bestcell{19.8} & 15.0 \\
     LLaVA-NeXT-7B-ViT~\cite{liu2023improvedllava}          & 1.3 & 19.6 & 47.3 & 1.3 & 17.3 & 29.8 & \bestcell{0.0} & 20.1 & 19.8 & 17.4 & \bestcell{3.2} & 8.3 & 0.2 & 2.4 & \bestcell{3.5} & 30.5 & 27.7 & 27.4 & \bestcell{0.0} & 21.4 & 0.4 & \bestcell{19.8} & \bestcell{14.9} \\
    \bottomrule
  \end{tabular}
  }
\end{table*}

%% file: content/tables/main_mad_lora_matrix.tex
\begin{table*}[htbp]
  \centering
  \caption{
{LoRA for MAD.} Test ACER$\downarrow$ (\%) after LoRA adaptation of the vision encoder on the features (compare~\cref{tab:feature_overview}). 
Overall results summarize intra-dataset (\colorbox{yellow!30}{ID}), cross-dataset (\colorbox{green!30}{CD}), and average performance.
\textbf{Bold}: best per column.
}
  \label{tab:app:lora-mad-intra-dataset-acer}
  \footnotesize
  \setlength{\tabcolsep}{2.25pt}
  \resizebox{0.75\linewidth}{!}{%
  \begin{tabular}{lccccccccccrrr}
    \toprule
    & \multicolumn{5}{c}{Train on FFHQ} & \multicolumn{5}{c}{Train on FRGC} & \multicolumn{3}{c}{Overall} \\
    \cmidrule(lr){2-6} \cmidrule(lr){7-11} \cmidrule(lr){12-14}
    Model & \cellcolor{yellow!30}{\textbf{\scalebox{0.825}{FFHQ}}} & \cellcolor{green!30}{\scalebox{0.9}{FRGC}} & \cellcolor{green!30}{\scalebox{0.9}{FRLL}} & \cellcolor{green!30}{\scalebox{0.85}{FERET}} & Avg & \cellcolor{green!30}{\scalebox{0.9}{FFHQ}} & \cellcolor{yellow!30}{\textbf{\scalebox{0.825}{FRGC}}} & \cellcolor{green!30}{\scalebox{0.9}{FRLL}} & \cellcolor{green!30}{\scalebox{0.85}{FERET}} & Avg & \cellcolor{yellow!30}{ID} & \cellcolor{green!30}{CD} & Avg \\
    \midrule
    \multicolumn{14}{c}{Model Zoo} \\
    \midrule
    InternViT-300M~\cite{chen2024internvl2}            & \bestcell{0.0} & 0.4 & \bestcell{0.0} & 24.5 & 6.2 & 31.5 & \bestcell{0.0} & 50.0 & 50.1 & 32.9 & \bestcell{0.0} & 26.1 & 19.6 \\
    BEiT-base~\cite{bao2022beit}                       & 0.5 & 1.4 & 0.5 & 29.5 & 8.0 & 45.7 & \bestcell{0.0} & 37.5 & 40.3 & 30.9 & 0.2 & 25.8 & 19.4 \\
    ViT-MAE-base~\cite{he2022mae}                      & 1.8 & 2.5 & 0.5 & 18.5 & 5.8 & 49.9 & \bestcell{0.0} & 31.4 & 39.6 & 30.2 & 0.9 & 23.7 & 18.0 \\
    Qwen2-VL-72B-ViT~\cite{wang2024qwen2vl}            & 0.6 & 0.2 & 3.4 & 10.3 & 3.6 & 43.8 & \bestcell{0.0} & 38.5 & 45.8 & 32.0 & 0.3 & 23.7 & 17.8 \\
    SigLIP-large~\cite{siglip2023}                     & 0.2 & 1.0 & 7.7 & 10.8 & 4.9 & 40.8 & \bestcell{0.0} & 44.4 & 33.3 & 29.6 & 0.1 & 23.0 & 17.3 \\
    Qwen3-VL-2B-ViT~\cite{bai2025qwen3vl}              & 0.1 & 0.7 & 2.7 & 5.7 & 2.3 & 47.2 & \bestcell{0.0} & 50.0 & 20.3 & 29.4 & 0.1 & 21.1 & 15.8 \\
    ViT-MAE-huge~\cite{he2022mae}                      & \bestcell{0.0} & 3.9 & 0.6 & 18.3 & 5.7 & 49.8 & \bestcell{0.0} & 16.4 & 32.4 & 24.7 & \bestcell{0.0} & 20.2 & 15.2 \\
    Qwen2.5-VL-3B-ViT~\cite{bai2025qwen25vl}           & 1.0 & 1.5 & 3.8 & 4.0 & 2.6 & 43.4 & \bestcell{0.0} & 49.0 & 17.7 & 27.5 & 0.5 & 19.9 & 15.0 \\
    Qwen2.5-VL-7B-ViT~\cite{bai2025qwen25vl}           & 0.6 & 0.3 & 2.7 & 4.7 & 2.1 & 47.4 & \bestcell{0.0} & 50.0 & 12.4 & 27.4 & 0.3 & 19.6 & 14.8 \\
    ViT-MSN-large~\cite{assran2023msn}                 & 1.5 & 1.7 & 1.7 & 13.3 & 4.6 & 47.8 & \bestcell{0.0} & 15.7 & 29.4 & 23.2 & 0.7 & 18.3 & 13.9 \\
    BEiT-large~\cite{bao2022beit}                      & 0.3 & \bestcell{0.0} & 0.7 & 5.9 & 1.7 & 42.6 & \bestcell{0.0} & 18.1 & 41.9 & 25.6 & 0.1 & 18.2 & 13.7 \\
    DINOv3-base~\cite{oquab2024dinov3}                 & 0.2 & 0.1 & 1.3 & 0.7 & 0.6 & 36.9 & \bestcell{0.0} & 33.1 & 30.4 & 25.1 & 0.1 & 17.1 & 12.8 \\
    DINOv2-base~\cite{oquab2023dinov2}                 & 0.1 & 0.9 & 2.8 & 5.2 & 2.2 & 46.9 & \bestcell{0.0} & \bestcell{0.0} & 43.9 & 22.7 & 0.1 & 16.6 & 12.5 \\
    DINOv3-H+~\cite{oquab2024dinov3}                   & \bestcell{0.0} & 0.6 & 0.3 & 1.9 & 0.7 & 26.4 & \bestcell{0.0} & 33.8 & 35.8 & 24.0 & \bestcell{0.0} & 16.5 & 12.4 \\
    InternViT-6B~\cite{chen2024internvl2}              & \bestcell{0.0} & \bestcell{0.0} & \bestcell{0.0} & 2.0 & \bestcell{0.5} & 16.7 & \bestcell{0.0} & 33.6 & 37.0 & 21.8 & \bestcell{0.0} & 14.9 & 11.2 \\
    DINOv2-R-base~\cite{darcet2024registers}           & 0.2 & 0.2 & 0.8 & 2.6 & 0.9 & 41.9 & \bestcell{0.0} & 0.3 & 40.8 & 20.7 & 0.1 & 14.4 & 10.8 \\
    \scalebox{0.875}{LLaVA-OneVision-7B-ViT~\cite{li2024llavaonevision}} & \bestcell{0.0} & \bestcell{0.0} & \bestcell{0.0} & 3.3 & 0.8 & 37.6 & \bestcell{0.0} & \bestcell{0.0} & 45.2 & 20.7 & \bestcell{0.0} & 14.4 & 10.8 \\
    CLIP ViT-B/32~\cite{radford2021learning}           & 0.5 & 2.1 & 12.1 & 8.1 & 5.7 & 35.9 & \bestcell{0.0} & \bestcell{0.0} & 23.8 & 14.9 & 0.2 & 13.7 & 10.3 \\
    DINOv2-giant~\cite{oquab2023dinov2}                & 0.1 & 0.2 & 0.5 & 1.5 & \bestcell{0.5} & 40.6 & \bestcell{0.0} & \bestcell{0.0} & 33.4 & 18.5 & \bestcell{0.0} & 12.7 & 9.5 \\
    ViT-MSN-base~\cite{assran2023msn}                  & 1.4 & 0.4 & 3.5 & 14.5 & 4.9 & 39.9 & \bestcell{0.0} & 0.3 & 15.6 & 13.9 & 0.7 & 12.4 & 9.4 \\
    DINOv2-R-giant~\cite{darcet2024registers}          & 0.1 & 0.4 & 3.9 & 2.1 & 1.6 & 43.8 & \bestcell{0.0} & \bestcell{0.0} & 25.1 & 17.2 & 0.1 & 12.5 & 9.4 \\
    \scalebox{0.875}{DeepSeek-VL2-small-ViT~\cite{wu2024deepseekvl2}}    & \bestcell{0.0} & \bestcell{0.0} & \bestcell{0.0} & 22.9 & 5.7 & 49.4 & \bestcell{0.0} & 0.3 & \bestcell{0.5} & 12.5 & \bestcell{0.0} & 12.2 & 9.1 \\
    LLaVA-NeXT-7B-ViT~\cite{liu2023improvedllava}          & \bestcell{0.0} & \bestcell{0.0} & \bestcell{0.0} & 8.2 & 2.1 & 46.3 & \bestcell{0.0} & \bestcell{0.0} & 14.9 & 15.3 & \bestcell{0.0} & 11.6 & 8.7 \\
    SigLIP-base~\cite{siglip2023}                      & 0.2 & 0.3 & 1.8 & 5.7 & 2.0 & 36.3 & \bestcell{0.0} & 15.4 & 8.0 & 14.9 & 0.1 & 11.2 & 8.5 \\
    Qwen3-VL-32B-ViT~\cite{bai2025qwen3vl}             & 0.7 & 1.2 & 3.2 & 3.5 & 2.2 & 35.4 & \bestcell{0.0} & 6.4 & 14.9 & 14.2 & 0.4 & 10.8 & 8.2 \\
    DINOv3-large~\cite{oquab2024dinov3}                & 0.1 & \bestcell{0.0} & 0.7 & 2.4 & 0.8 & 34.7 & \bestcell{0.0} & 7.8 & 14.7 & 14.3 & \bestcell{0.0} & 10.1 & 7.5 \\
    \scalebox{0.925}{DeepSeek-VL2-7B-ViT~\cite{wu2024deepseekvl2}}       & \bestcell{0.0} & \bestcell{0.0} & \bestcell{0.0} & 16.5 & 4.1 & 26.6 & \bestcell{0.0} & \bestcell{0.0} & 12.2 & 9.7 & \bestcell{0.0} & 9.2 & 6.9 \\
    EVA-CLIP-8B~\cite{sun2023eva}                      & 0.1 & 1.1 & 1.6 & 3.7 & 1.6 & 37.8 & \bestcell{0.0} & \bestcell{0.0} & 6.4 & 11.0 & \bestcell{0.0} & 8.4 & 6.3 \\
    CLIP ViT-H/14~\cite{radford2021learning}           & \bestcell{0.0} & 1.0 & 1.0 & 1.0 & 0.8 & 25.9 & \bestcell{0.0} & \bestcell{0.0} & 13.8 & 9.9 & \bestcell{0.0} & 7.1 & 5.3 \\
    CLIP ViT-L/14~\cite{radford2021learning}           & 0.2 & 3.1 & 0.4 & \bestcell{0.6} & 1.1 & \bestcell{11.5} & \bestcell{0.0} & 5.4 & 8.1 & \bestcell{6.3} & 0.1 & \bestcell{4.9} & \bestcell{3.7} \\
    \bottomrule
  \end{tabular}
  }
\end{table*}